\documentclass{article}
\usepackage{amsmath}
\allowdisplaybreaks

\usepackage[preprint]{neurips_2026}
\usepackage[pdftex]{graphicx}

\usepackage[utf8]{inputenc}
\usepackage[T1]{fontenc}
\usepackage{hyperref}
\usepackage[capitalize,nameinlink]{cleveref}
\usepackage{url}
\usepackage{booktabs}
\usepackage{amsfonts}
\usepackage{nicefrac}
\usepackage{microtype}
\usepackage{xcolor}
\usepackage{multirow}
\usepackage{xspace}
\usepackage{subcaption}
\usepackage{array}
\usepackage{wasysym}
\usepackage{tabularx}
\usepackage{algorithm}
\usepackage{algpseudocode}
\usepackage{colortbl}
\usepackage{fvextra}
\usepackage{tablefootnote}
\usepackage{xurl}

\DeclareUnicodeCharacter{2013}{\textendash}
\DeclareUnicodeCharacter{2014}{\textemdash}
\DeclareUnicodeCharacter{2022}{\textbullet}
\DeclareUnicodeCharacter{2026}{\ldots}
\DeclareUnicodeCharacter{2039}{\ensuremath{\langle}}
\DeclareUnicodeCharacter{203A}{\ensuremath{\rangle}}
\DeclareUnicodeCharacter{2192}{\ensuremath{\rightarrow}}

\definecolor{paperpaletteblue}{HTML}{1F77B4}
\definecolor{paperpalettered}{HTML}{D62728}
\definecolor{paperpaletteorange}{HTML}{FFAF0E}
\colorlet{redtier}{paperpalettered!32}
\colorlet{yelltier}{paperpaletteorange!34}
\colorlet{gaptier}{paperpaletteblue!28}

\definecolor{sbblue}{HTML}{4878d0}
\definecolor{sbred}{HTML}{d65f5f}
\definecolor{sbpurple}{HTML}{926db1}
\definecolor{sbgreen}{HTML}{6acc64}
\definecolor{sbbluedeep}{HTML}{4c72b0}
\definecolor{sbreddeep}{HTML}{c44e52}
\definecolor{sbpurpledeep}{HTML}{8073b0}
\definecolor{sbgreendeep}{HTML}{55a868}
\definecolor{sborange}{HTML}{ee7542}
\definecolor{sborangedeep}{HTML}{dd8452}

\hypersetup{
  colorlinks,
  citecolor=sbgreendeep,
  linkcolor=sbred,
  urlcolor=sbgreendeep}

\title{\textsc{RedactionBench}}
\author{%
  Sean Brynj\'{o}lfsson\thanks{Corresponding author: \texttt{sbrynjolfson@a10networks.com}} \quad
  Shashvat Jayakrishnan \quad
  Esha Sali \quad
  Diptanshu Purwar \\[1.5ex]
  \textbf{Madhav Aggarwal} \\[1.5ex]
  A10 Networks, Inc.\\
}

\begin{document}

\newif\ifsubmit
\definecolor{author_colorA}{rgb}{0,0.5,1}
\definecolor{author_colorB}{rgb}{0.2,.64,0}
\definecolor{author_colorC}{RGB}{33, 145, 140}
\definecolor{author_colorD}{rgb}{0,1,1}
\definecolor{changes_color}{rgb}{0.05,0.5,0.3}
\definecolor{mathbrace_color}{rgb}{0.2,0.5,1.0}

\newcommand{\benchmark}{\textsc{RedactionBench}\xspace}
\newcommand{\metric}{\emph{R-Score}\xspace}

\ifsubmit
    \newcommand{\sean}[1]{}
    \newcommand{\madhav}[1]{}
    \newcommand{\esha}[1]{}
    \newcommand{\shashvat}[1]{}
    \newenvironment{changes}
      {
      }
      {}
\else
    \newcommand{\madhav}[1]{\textbf{\textcolor{author_colorA}{MA: #1}}}
    \newcommand{\sean}[1]{\textbf{\textcolor{author_colorB}{SB: #1}}}
    \newcommand{\shashvat}[1]{\textbf{\textcolor{author_colorC}{SJ: #1}}}
    \newcommand{\esha}[1]{\textbf{\textcolor{author_colorD}{ES: #1}}}
    \newcommand{\todo}[1]{{\textcolor{red}{[TODO: #1]}}}
    
    \newenvironment{changes}
      {
      \textbf{Changes:}
      \color{changes_color}
      }
      {}
\fi

\newcommand{\stoken}[1]{\colorbox{blue!10}{\texttt{\textbf{#1}}}}
\newcommand{\var}[1]{\text{\itshape #1}}
\newcommand{\tier}[2]{\begin{tabular}[t]{@{}c@{}}#1\\[1pt]{\scriptsize\rm #2}\end{tabular}}

\newtheorem{theorem}{Theorem}
\newtheorem{lemma}{Lemma}
\newtheorem{definition}{Definition}
\newtheorem{proposition}{Proposition}


\maketitle
\begin{abstract}
LLMs are increasingly being applied to sensitive domains that require redacting personally-identifiable information (PII) before processing. While redacting PII has become a de facto data-cleaning prerequisite, existing benchmarks conflate the mechanics of extraction with the semantics of privacy. A phone number in a public directory is not equivalent to one in a medical record. Whether a given piece of information constitutes a violation depends heavily on who holds it, why, and in what context---fundamentally differentiating the redaction task from simple entity recognition. Grounded in this principle of contextual integrity, we introduce \benchmark, a manually annotated benchmark comprising 200 diverse documents across 11 domains, with a majority seeded from real-world sources. \benchmark also introduces a novel character-level redaction metric called \metric that treats semantically similar redactions equally and nullifies the impact of shallow formatting choices (e.g., redacting a \texttt{phone\_number} as: "$\texttt{(***) ***-****}$" vs. "$\texttt{**************}$"). Extensive evaluations across Named-Entity Recognition (NER) models, entity-extraction Small Language Models (SLM), and frontier LLMs equipped with agentic tools (Claude Opus, OpenAI GPT) demonstrate that contextual redaction remains an unsolved problem. Results from our human evaluation (85 participants) on \benchmark reveal a stark dichotomy in privacy perception: annotators show consensus with our target labels for mandatory redactions (89.4\%) and safe text preservations (94.1\%), but fail to agree with contextual redactions (47.7\%). This variance demonstrates the subjective nature of contextual privacy and motivates our evaluation metric \metric, which decouples contextual ambiguity from strict redaction precision. We compare 35 models using \benchmark across model families and report their performance for PII redaction. Finally, we release \benchmark publicly to establish a baseline for future privacy-preserving redaction systems. We hope this benchmark inspires a shift towards efficient model design and standardized evaluations for text redaction. 
\end{abstract}

\vspace{-0.3cm}
\section{Introduction}
\begin{figure}[htbp]
\centering
\includegraphics[width=1.0\linewidth]{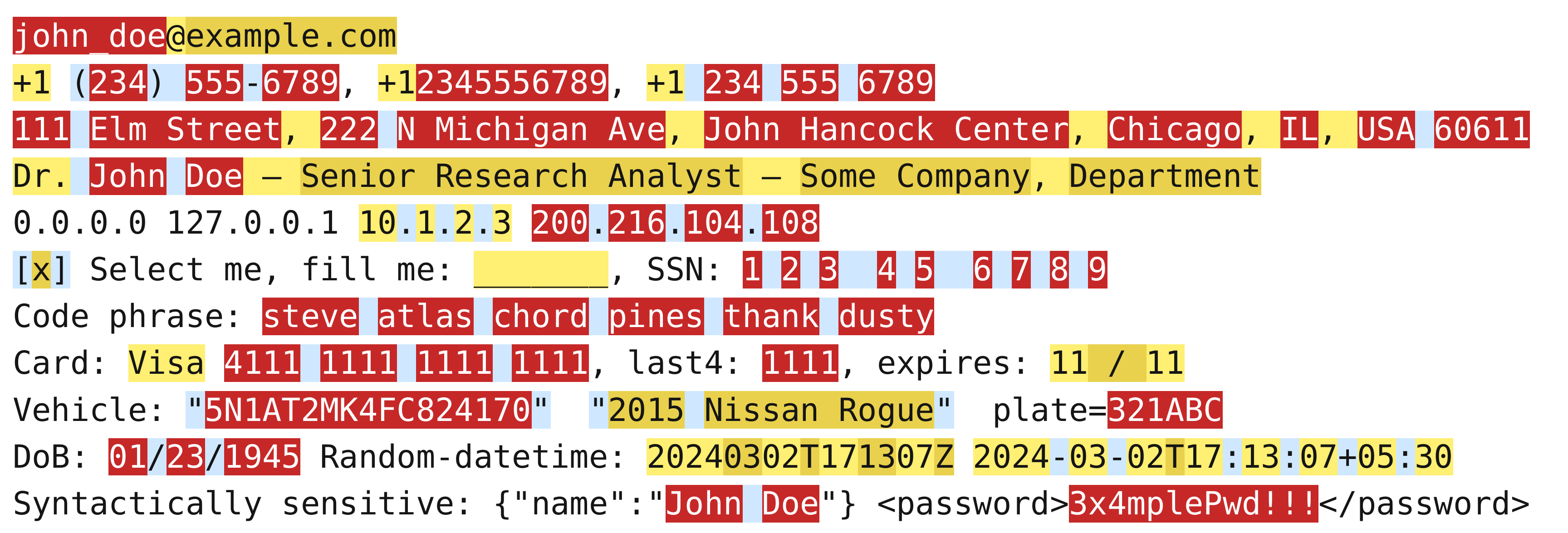}
\caption{\benchmark provides rich segmentations across two tiers: mandatory (red) and contextual (yellow). An alternating shade of yellow is used to disambiguate adjacent contextual spans. Combinators, which connect parts of coherent entities, are represented by light blue.}
\label{fig:labeling}
\end{figure}

Personally Identifiable Information (PII) redaction is the precise, localized masking of targeted entities to remove sensitive information while preserving overall semantic utility. This constraint to operate only on specific characters fundamentally distinguishes redaction from privacy techniques that involve rewriting using structured outputs, typically performed with generative models. By instead framing redaction as a targeted character-level binary classification task, the approach successfully avoids severe computational bottlenecks inherent to unbounded text generation.

This computational efficiency makes redaction uniquely suited to "privacy-on-the-edge," where sensitive data must be sanitized on local devices before transmission over the network. Despite its irrefutable benefits in several high-stakes environments, redaction has remained largely unexamined in the peer-reviewed literature. This has led to a variety of ad hoc datasets, incomplete task definitions, and a lack of standardized evaluations and comparisons. Strikingly, neither the top 30 downloaded PII models nor the three dominant benchmarks on Hugging Face come from peer-reviewed work. This gap has been exacerbated by a shift towards sophisticated privacy applications that utilize LLMs, which remain computationally infeasible in low-resource, low-latency settings.

The widespread utility of Large Language Models (LLMs) is largely driven by their robust instruction-following capabilities, which allow complex constraints—including privacy directives—to be specified natively in natural language. This uniquely positions LLMs to operationalize sophisticated privacy frameworks, such as Contextual Integrity (CI)---a theory of information transfer that considers social norms and circumstances before making privacy decisions. Given enough context and an understanding of intent, contextual integrity demonstrates how the same piece of information can go from being benign in some contexts to redactable PII in others. For example, details on a quarterly financial report can be considered benign entities until "\texttt{DO NOT DISTRIBUTE - INTERNAL MEMO}" appears at the top.

We identify three key challenges with existing works in the field of redaction:
\begin{enumerate}
    \item While LLM-based \textit{privacy} techniques have converged on using contextual integrity as a framework, current \textit{redaction} techniques do not adopt a formal privacy framework.
    \item Existing dataset benchmarks do not emulate real documents, making it unclear if they can generalize to them (\Cref{tab:benchmark-comparison}).
    \item Deployed models are siloed into their individual taxonomies, rendering them incomparable and creating an environment where each model performs best on its own benchmark.
\end{enumerate}

To address the above gaps, we create \benchmark, a real-world privacy benchmark that unifies evaluation across an ecosystem of independently-evolved redaction models through a universal privacy framework. Complementing our benchmark, we introduce \metric, a novel redaction metric with properties inspired by both the conventional F1 score and Intersection-over-Union (IoU), a metric commonly found in semantic image segmentation literature (\Cref{subsec:metric}). 

We find that Small Language Models (SLMs) finetuned for entity extraction or PII redaction are bested by smaller BERT-based and GLiNER models (\Cref{sec:experiments}). The best-performing SLM achieves an \metric of \textbf{0.45} compared to the best BERT-based \metric of \textbf{0.58} and GLiNER at \textbf{0.47}. Notably, the newly released Privacy Filter model \citep{openai2026privacyfilter} achieves an \metric of \textbf{0.58}. Across the full 200-document benchmark, frontier models equipped with agentic tool-calling harnesses achieve the highest mean \metric of \textbf{0.71}. On the 48-document user-study split, the same class of models reaches a mean \metric of \textbf{0.81}, exceeding the aggregated human baseline of \textbf{0.77} (\Cref{tab:human-model-comparison}).

To ensure our target labels and metric reflect real-world privacy expectations, every document in \benchmark has been meticulously hand-annotated (\Cref{subsec:benchmark_construction}). We also run a dedicated user study (\Cref{sec:user-study}), making it the first redaction benchmark with an extrinsically verified, real-world definition of privacy. \benchmark serves as both a formidable target for small, dedicated edge models and a necessary sanity check for complex systems navigating obfuscation or nuanced entity classification. 

\begin{table}[htbp]
\centering
\caption{\benchmark compared to existing PII benchmarks.}
\label{tab:benchmark-comparison}
\renewcommand{\arraystretch}{1.05}
\resizebox{\textwidth}{!}{%
\scriptsize
\begin{tabular}{@{}l c c c c c c c@{}}
    \toprule
    \textbf{Benchmark} & \textbf{Size} & \begin{tabular}{@{}c@{}}\textbf{Source} \\ \textbf{Documents}\end{tabular} & \begin{tabular}{@{}c@{}}\textbf{Median Sample} \\ \textbf{Annotations}\end{tabular} & \begin{tabular}{@{}c@{}}\textbf{Total} \\ \textbf{Annotations}\end{tabular} & \begin{tabular}{@{}c@{}}\textbf{Privacy} \\ \textbf{Definition}\end{tabular} & \textbf{Source} & \begin{tabular}{@{}c@{}}\textbf{User} \\ \textbf{Study}\end{tabular} \\
    \midrule
    \textbf{RedactionBench} (ours) & 200 & 101 & 108 & 53{,}286 & CI & Human & Yes \\
    Ai4Privacy~\citep{ai4privacy2024piimasking} & 43.5K\tablefootnote{English subset only, train split as no explicit test split is provided.} & 0 & 3 & 137{,}093 & Taxonomy & Synthetic & No \\
    Nemotron-PII~\citep{nemotron-pii} & 100K & 0 & 7 & 850{,}340 & Taxonomy & Synthetic & No \\
    Gretel-PII~\citep{gretelai2024synthetic} & 5.6K & 0 & 6 & 37{,}017 & Taxonomy & Synthetic & No \\
    \bottomrule
\end{tabular}%
}
\end{table}

\section{Related Work}
\subsection{Privacy Frameworks and Benchmarks}
\benchmark adopts the contextual integrity (CI) framework of \citet{nissenbaum2004privacy}, which has become the standard theoretical lens for privacy in language models \citep{mireshghallah2024can, 
cheng2024cibenchbenchmarkingcontextualintegrity,
li2025privacychecklistprivacyviolation,
jeon2026redacbench}. These aforementioned works investigate CI, but not specifically redaction. Closer to our investigation are works covering anonymization, rewriting, and query-aware redaction \citep{sun2024depromptdesensitizationevaluationpersonal,shen2025piibenchevaluatingqueryawareprivacy, 
prvl, kim2026privasis, 
ponomarenko-etal-2026-capid}, which do invoke CI, but do not make it a primary evaluation criterion.

The open-source landscape for general PII redaction is anchored by three benchmarking datasets \citep{ai4privacy2024piimasking, gretelai2024synthetic, nemotron-pii}. Despite widespread adoption, none are peer-reviewed because each benchmark defines its own label taxonomy rather than an explicit notion of privacy. Further, it is difficult to determine whether high performance indicates privacy-preserving behavior or simply agreement with dataset-specific labeling conventions.
\benchmark intends to alleviate this gap in future work on PII redaction, guided by the principle of CI.

\subsection{Evaluation Metrics}
Machine privacy practitioners have adopted various techniques from Named Entity Recognition (NER), a subfield of Natural Language Processing (NLP). The dominant convention for NER originates in \cite{ramshaw-marcus-ner}, which introduces BIO labels (Beginning, Inside, Outside) to demarcate entity boundaries (e.g., \texttt{"Hi, I'm John Doe"} maps to [\texttt{"Hi":O, "I'm":O, "John": B-PER, "Doe": I-PER}]). Traditionally, models have been evaluated using Precision, Recall, and F1 score \citep{ng-etal-2013-conll}. Under this \textit{strict} scheme, predicted spans that do not exactly match a gold span---whether off-by-one or partial---count as false positives, while undetected entities count as false negatives. Despite limitations across fuzzy boundaries, strict F1 remains the dominant reported measure across NER research 
\citep{sang2003conll, mitchell2005ace2004, walker2006ace2005, hovy-etal-2006-ontonotes, shen-etal-2021-locate, fries2022bigbio, zhang2019biowordvec, scalable-multilingual-pii} and is reported by a majority of NER and PII models (\Cref{tab:model-performance}).

The inadequacy of strict matching for entities with legitimate boundary variations has motivated a range of alternate formulations. \citet{nejadgholi-etal-2020-extensive} catalogs several named tolerances---\textit{boundary matching, subset redaction, any-fragment inclusion, core-term correction}. \citet{zhu-li-2022-boundary} addresses the same problem from a training perspective, proposing boundary smoothing to reduce over-confident predictions at span edges. At the token level, \citet{prvl} introduces the SPriV score, which measures the proportion of sensitive tokens successfully masked; while intuitive, this technique fails to capture false positives, remains sensitive to tokenization, and is biased toward longer entities. \citet{snomed-ct-entity-linking-challenge} adopts per-class macro IoU (Jaccard index), directly rewarding character-span overlap rather than exact binary matches. 

Our metric subsumes the tolerance cases enumerated by 
\citet{nejadgholi-etal-2020-extensive} and treats them as equivalent, scoring all degrees of partial overlap continuously rather than through a discrete taxonomy. We draw additional inspiration from image semantic segmentation, where \citet{bhowmik2025unionoverintersections} permits multiple coverings of the same region at different granularities---analogous to how a redaction benchmark must simultaneously credit detecting \textit{``John,''} \textit{``Doe,''} and \textit{``John Doe''} as overlapping but valid predictions of the same sensitive entity. Nested NER \citep{shen-etal-2021-locate, gliner} can achieve similar effects by predicting multiple granularities at once.

\subsection{Diverse Model Architectures}
NER solutions have evolved from feature statistical models to RNN and LSTM architectures, and now to the transformer era ushered in by \citet{devlin-etal-2019-bert}. BERT and its successors \citep{liu2019roberta, debertav3, modernbert} remain standard baselines, against which the Generalist Lightweight NER (GLiNER) family \citep{gliner, gliner2} is benchmarked. GLiNER models are themselves used as baselines when evaluating larger autoregressive models---such as LLaMA \citep{grattafiori2024llama3}, GPT-5 \citep{singh2025openaigpt5}, Claude Opus \citep{anthropic2025claude_opus_4_5} and GLM-5.1 \citep{model_zai_org_glm_5_1}---which can be prompted for extraction with custom objectives and tool calls. A variant of the standard encoder-only architecture was also recently released by \citet{openai2026privacyfilter}.

\section{Method}
\subsection{Benchmark Details}
\label{subsec:benchmark_construction}
\textbf{Document Composition: }We enumerate a set of 200 distinct \textit{genres}\footnotemark\ (\Cref{tab:unstructured-genres}, \Cref{tab:redactionbench-structured-genres}) for use in our benchmark, drawn from publicly available documents from reputable sources, including government websites, academic resources, and company-provided forms. These genres were later organized into 11 taxonomic categories, seven unstructured and four structured (\Cref{sec:data_insights}).
\footnotetext{\textit{genre} refers to a document class recognized by its purpose, structure, and the conventions that govern how it is read}

\begin{figure}[htbp]
\centering
\includegraphics[width=1.0\linewidth]{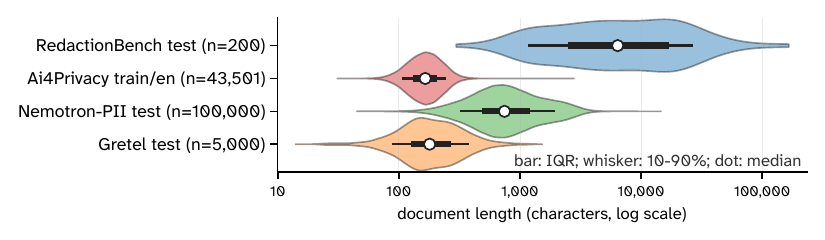}
\caption{Length Distributions across popular general-purpose PII benchmarks.}
\label{fig:document-length-violins}
\end{figure}
\textbf{Document Extraction: }A majority of our source documents are in \texttt{.pdf} format. These documents were extracted using pdfminer as the primary text extraction tool. In extreme cases, we resort to a mix of OCR techniques such as OlmOCR \citep{poznanski2025olmocr}, InfinityParser \citep{wang2025infinityparser}, and PaddleOCR \citep{cui2025paddleocr}. Additional non-PDF formats, such as \texttt{.doc}(\texttt{x}), were either handled manually or via \texttt{soffice}, \texttt{python-docx}, and \texttt{docx2txt}.

While OCR techniques helped us achieve our document targets, using LMs for extraction was especially involved due to their tendency to generate homogeneous Markdown, which compromises structure. We prefer PDFMiner's slightly mangled extractions over OCR LMs for preserving document realism. Gemini, Claude Opus, and GPT-5 helped us repair PDFMiner outputs in a strictly guided manner; edits were specifically prompted, and no generic query could be formulated (e.g., \textit{"repair this flattened table using \texttt{\textbackslash t} characters instead of spaces"}). Some augmentations were achieved using regex filtering; checkboxes, for instance, lack a standardized convention, and were rendered as \texttt{\CheckedBox} in some documents and as \texttt{[X]} in others. Compared to other publicly available benchmarks, we place strong emphasis on document length diversity during curation. The longest documents in \benchmark are limited to 10 pages (\Cref{fig:document-length-violins}).

\subsection{Labeling Procedure}
\label{subsection:labeling-procedure}
\benchmark has three label classes: mandatory \textit{"unsafe"} entities, contextual entities, and implicit gaps between them. As the name suggests, \textit{Mandatory} entities represent information units that are considered unsafe to share across all contexts. \textit{Contextual} entities, on the other hand, are inherently ambiguous and require more context to be strictly categorized as safe or unsafe. Since contextual spans are not necessarily redacted under our benchmark's metric, we overload their definition further by representing "structural" and "combination" characters---rendered in blue (\Cref{fig:labeling}). These special characters are used as aggregators to combine entity units of the same color. This prevents an IPv4 address entity, composed of 4 mandatory sub-spans, from being considered $4\times$ as sensitive as a single mandatory-span password. Multi-line entity boundaries are definitively broken across newline characters while labeling.

Entities in \benchmark are semantically partitioned to permit equivalent interpretations across entity boundaries. This is especially important when dealing with highly composite entities, such as addresses and names with titles and complex affixes. Our target labels decompose composite entities into their constituent parts. For instance, while "John Hancock Center" (mentioned in \Cref{fig:labeling}) contains the name "John Hancock", the entity itself is a venue, not a person. While this behavior is entirely logical, it will be recorded as a deviation from our target labels. Language Models complement our manual annotation and provide a warm-start checkpoint via semantic pattern matching for documents with hundreds of entities.

\textbf{Label Augmentation: }Language models were heavily utilized to generate plausible synthetic content for our documents. Throughout this process, we observed several common failure modes which appeared to be endemic across large models (\Cref{tab:synthetic-pitfalls}). Conventional techniques, such as \textit{structured output mode}, were unable to prevent several LM failure cases. The final recipe that worked for us was to wire together prompt engineering, in-context learning~\citep{brown2020language}, custom logit processors, and agentic tool calls with seed data generators (the Faker library \citep{faker_library}) to construct a partially reproducible workflow.

\subsection{Benchmark Metric}
\label{subsec:metric}
\metric lies between the IoU and the strict F1 score, inheriting desirable properties from both. Drawing parallels with IoU, we assign partial credit for partially correct redactions, but, unlike the F1 score, \metric is invariant to entity length and assigns steep penalties for imprecise boundaries. The key properties of our metric are that (1) model performance is governed by coverage of mandatory entities and (2) performance is independent of contextual semantics as long as contextual span boundaries are clean.
    
Our metric at a high level is:
$$ \textrm{\metric} = \frac{\text{Mandatory Entity Coverages}}{\text{\# Mandatory Entities}+\text{Contextual Residual Penalties} + \text{False Positive Penalties}} $$

Before proceeding, it is important to establish that \emph{spans} comprise \emph{entities}. Let $P$ be the set of characters redacted by a user or model. Let $\mathbf{M}$ be the set of all mandatory spans and $R$ be a partition of $\mathbf{M}$ into entities. Similarly, let $\mathbf{C}$ be the set of all contextual spans and $Y$ its partition into contextual entities. Partitioning behavior is specified by our combination rules along with definitions for False Positives. Each span $s_i$ contributes a local \textit{numerator} and \textit{denominator} score \((n_i,d_i)\), pooled by the parent definition into a single, entity-level \((n,d)\) term. Finally, the overall \metric can be computed over entities as:
\[
\textrm{\metric} =\frac{\sum_{i} n_i}{\sum_{i}d_i}
\]

\textbf{Mandatory Entities} must be covered in all cases and include unambiguously sensitive items such as API keys. For some mandatory entity $r\in R$, we compute its score as the mean coverage of its constituent spans, $s\in r$:
\[
    (n_r,d_r)=\left(\underset{s\in r}{\text{mean}}\frac{|s\cap P|}{|s|},\ 1\right)
\]

\textbf{Contextual Entities} are ambiguous redactions, and therefore only contribute to the final score if attempted (i.e., with a nonzero intersection) or \textit{selected} by combinators and grouping rules. Each constituent contextual span may be attempted or selected independently. For a contextual entity $y\in Y$, we first define the active subset
\[
{
A_y(P)=\{s\in y:\ |s\cap P|>0\ \mathrm{or}\ \mathrm{selected}(s\mid P)\}.
}
\]
If \(A_y(P)\) is empty, the contextual entity emits no scoring term. Otherwise, the mean is taken only over active spans, so inactive contextual subspans are excluded rather than included as zero-valued terms:
\[
{
(n_y,d_y)=
\left(0,\ 1-\underset{s\in A_y(P)}{\mathrm{mean}}\frac{|s\cap P|}{|s|}\right).
}
\]
Correctly labeled contextual entities will not increase the score, yet mislabeled attempts will dock points off the final score. Necessary for completing the definition of our metric, contextual spans soak up variance from ambiguities. In documents with no mandatory labels or all contextual labels, spans behave like active optional mandatory labels and can contribute to a positive score while remaining optional: 
This exception is necessary because otherwise an all-contextual document would have no mandatory terms from which a positive score could be earned:
\[
{
(n_y,d_y)=
\left(\underset{s\in A_y(P)}{\mathrm{mean}}\frac{|s\cap P|}{|s|},\ 1\right),
\qquad A_y(P)\neq\emptyset.
}
\]

\textbf{False Positives }, \(f\in F\), are contiguous sets of redacted characters that do not intersect the support of a contextual or mandatory span. False positives contribute $(n,d)=(0,1)$ by default and $(n,d)=(0,2)$ when an entire gap ($\geq 3$ characters) between entities is mislabeled. Doubling the weight ($d=2$) ensures that users are rewarded for identifying discontinuities between separate entities. This small tweak treats a full over-redaction as a double cover---false positives on both sides of a gap.
\[
    (n_f,d_f)=(0,1+1\{\text{covers\_entire\_gap}\geq 3\})
\]

\textbf{Combining Spans into Entities and Span Selection Rules: }For convenience when labeling by hand, we detect "combinators" and "grouping characters" as single-character contextual labels (\Cref{alg:redactionbench_combinator_construction}) to group connected spans of the same color into larger entities. These heuristics help balance the weight of entities that are represented by multiple disconnected spans (e.g., the digits in an IP address \texttt{127.0.0.1}). For example, in the full name \texttt{"John M. Doe"}, the space and punctuation need not be redacted. Our annotations provide permissible fine-grained segmentations: \texttt{"John", "M.", "Doe"} separately, or \texttt{"John M. Doe"} together, are equally valid. Combining them allows us to avoid weighing this entity three times as much as a joined social security number. The contribution of a combined entity is simply the component-wise mean of the $k$ contained spans.

We also introduce a prediction-dependent selection algorithm in~\Cref{alg:redactionbench_combinator_selection} that forces contextual entities to be selected based on surrounding redaction choices.

\section{Experimental Setup and Results}
\label{sec:experiments}
\begin{figure}[htbp]
\centering
\includegraphics[width=\linewidth]{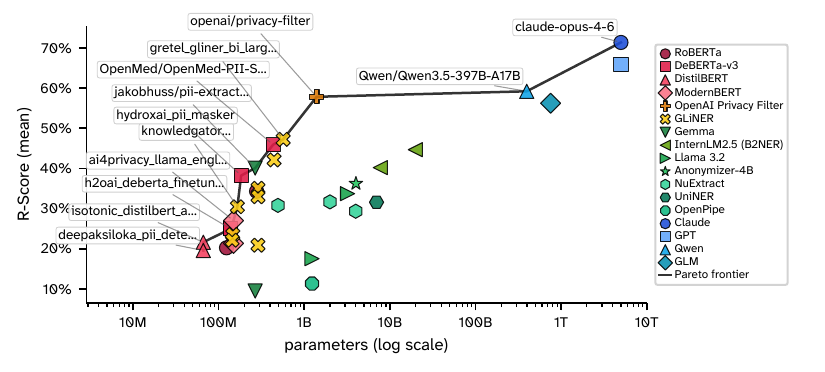}
\caption{
    Cross-architectural comparison of model performance (mean overall \metric) with respect to the number of parameters. All models improving over other models in at least one dimension (size or performance) are identified along the Pareto frontier. Red icons are used for token-prediction models, yellow for span-models, green for generative models, and blue for frontier models. Comprehensive results available in \Cref{tab:model-performance} with per-family results in \Cref{tab:evals-tokenbased,tab:evals-spanbased,tab:evals-generative}.
}
\label{fig:model-performance-over-size}
\end{figure}

We evaluate 35 public model entries, corresponding to 34 unique checkpoints, on \benchmark and report the mean, median ($P_{50}$), and 20th percentile ($P_{20}$) performance across each category using our metric. We also report macro-averages in~\Cref{tab:model-performance} across categories as heuristics for overall performance. Language models for evaluation are chosen based on their popularity, recency, and adaptability to the PII redaction task. For detailed insight into how each evaluated model was adapted for consistency while redacting spans, refer to~\Cref{sec:model_harness}. We mark a Pareto frontier on~\Cref{fig:model-performance-over-size} to represent which models punch above their weight on \benchmark.

\textbf{Token-level: }Refers to models that output per-token classifications and group entities together as a post-processing step. The most common paradigm stems from classical NER, which adopts BIO (Beginning, Inside, Outside) tags for token-level classification. During post-processing, we merge non-\texttt{Outside} predictions that follow an \texttt{Outside} tag, regardless of the target label (Mandatory, Contextual, or Gap). Hence, non-\texttt{Outside} tags are simply mapped to a \texttt{Mandatory} redact decision.

\textbf{Span-level: }Only the Generalist Lightweight NER family (GLiNER) \citep{gliner, gliner2} qualifies under this category. Multiple variants of this family exist, differentiated by their BERT backbones and whether they are finetuned beyond their base versions. While GLiNER models claim to work with dynamic inputs (open-vocabulary class labels as inference-time arguments), we notice top performance with the subset of labels used to train their finetuned extensions. \benchmark leverages a union of label corpora across PII-related GLiNER works as its extended evaluation set. This ensures uniformity in label formatting across sister models. For each GLiNER model, we use a single global score threshold across all samples and select it based on aggregate evaluation performance, maximizing open-vocabulary span model performance.

\textbf{Generative: } Models from this family apply privacy techniques by rewriting or generating structured outputs, rather than directly annotating or masking input sequences at the character level. We evaluate two types of generative models: Generative NER finetunes and general-purpose Language Models, which may or may not be supplemented with agentic tool-calling abilities. Generative NER models produce entity mappings from a set of target labels, provided alongside the query text, as structured outputs. We evaluate B2NER \citep{b2ner} as a representative model from this family. B2NER variants are evaluated with fixed-size sliding windows sized to their model context limits; predictions from overlapping windows are merged before scoring.

For general-purpose Language Models, we first evaluate Frontier LLMs defined by their large parameter counts ($>100$B). We compare models with native agentic capabilities and leverage them for the PII redaction task. We observe that disguising the task as a file-editing tool call yields stronger results than free-form querying. Thus, we prompt all frontier models to edit \benchmark documents through their preferred file-editing tools and intercept their invocations, converting them into spans. Our tool-calling strategies and harnesses for each model are defined under~\Cref{sec:model_harness}.

\section{User Study}
\label{sec:user-study}
\begin{table}[b]
\centering
\caption{Mode-pooled human redactions evaluated on part of \benchmark.}
\label{tab:per-user-scores}
\resizebox{\textwidth}{!}{%
\begin{tabular}{ *{24}{c} }
    \toprule
    \multicolumn{3}{c}{Acad.} & \multicolumn{3}{c}{Email} & \multicolumn{3}{c}{Fin.} & \multicolumn{3}{c}{Gov.} & \multicolumn{3}{c}{Legal} & \multicolumn{3}{c}{Med.} & \multicolumn{3}{c}{Ops.} & \multicolumn{3}{c}{Unstructured} \\
    $P_{20}$ & $P_{50}$ & Mean & $P_{20}$ & $P_{50}$ & Mean & $P_{20}$ & $P_{50}$ & Mean & $P_{20}$ & $P_{50}$ & Mean & $P_{20}$ & $P_{50}$ & Mean & $P_{20}$ & $P_{50}$ & Mean & $P_{20}$ & $P_{50}$ & Mean & $P_{20}$ & $P_{50}$ & Mean \\
    \midrule
    0.69 & 1.00 & 0.85 & 0.68 & 0.79 & 0.82 & 0.67 & 0.81 & 0.78 & 0.51 & 0.68 & 0.70 & 0.70 & 0.83 & 0.82 & 0.50 & 0.59 & 0.64 & 0.67 & 0.74 & 0.75 & 0.61 & 0.78 & 0.77 \\
    \bottomrule
\end{tabular}%
}
\end{table}
We analyze diversity in human privacy strategies and cases in which contextual integrity generates ambiguity with our user study. Users see individual windows from up to 11 documents across 7 unstructured document categories, with 8 documents randomly sampled from each category. We exclude the 4 structured categories because documents in these categories are typically handled only by domain experts. Our windows are roughly 25-line selections inside documents to avoid labeling fatigue from the full text. The surrounding text remains visible to the annotator as persistent context. Remaining details about our setup, such as the labeling tool (\Cref{fig:user-study}) and user personas, have been deferred to~\Cref{sec:study-details}.

\subsection{Redaction Units}
\label{subsubsection:redaction-units}
Direct statistical analysis on the exponential space of possible redactions is hard to interpret (one-bit binary decision per character). By framing each document as a series of \textit{units}, rather than individual characters, we can significantly compress the action space. In particular, a redaction unit corresponds to a single entity or a single false-positive term, as laid out in \Cref{subsec:metric}. We provide worked-out calculations depicting these units in \Cref{sec:worked-example}:
\begin{enumerate}
    \item Entity units: we define one unit for each mandatory entity $r\in R$ and one for each contextual entity $y\in Y$. These units are selected if any redacted characters intersect the entity.
    \item Gap units: for each gap between entities, we define one unit, $g_a$, with a second unit $g_b$ when the gap is three or more characters. $g_a$ is considered selected when any character in the gap has been falsely classified, while $g_b$ is selected when all gap characters are covered.
\end{enumerate}


\subsection{Results}



\begin{table}[b]
\centering
\small
\caption{Disagreement on \benchmark. \(N_t\): qualifying units (\(m_u \geq 2\)). \(\bar{p}_t\): mean redaction rate. \(D_t\): mean pairwise disagreement with bootstrap 95\% CI (\(B = 1000\)). \(\alpha_t\): within-subset Krippendorff's \(\alpha\). Contextual entities show \(1.80\times\) the disagreement of mandatory entities and \(3.22\times\) that of gaps.}
\label{tab:krippendorffs}
\begin{tabular}{lrrlr}
\toprule
\textbf{Unit Type} & \(N_t\) & \(\bar{p}_t\) & \(D_t\) [95\% CI] & \(\alpha_t\) \\
\midrule
Global              & 7{,}627 & 0.237 & ---                       & 0.540 \\
\hline
\addlinespace
Mandatory entities (r)  &   590   & 0.894 & 0.183 [0.162, 0.205]      & 0.069 \\
Contextual entities (y) & 1{,}928 & 0.477 & 0.329 [0.315, 0.343]      & 0.303 \\
Gaps ($g_a$,$g_b$)                & 5{,}109 & 0.059 & 0.102 [0.096, 0.109]      & 0.148 \\
\bottomrule
\end{tabular}
\end{table}

\textbf{Human Performance: }To estimate human performance on \benchmark, we leverage user evaluations with zero starting windows to avoid any confounding effects related to varying \metric starting points across windows. For each character in each document, we take the most common user decision (to redact or not), with ties defaulting to no redaction. These sets of characters are then scored directly against the target labels for an overall mean \metric of $0.77$. A category-level breakdown can be found above in \Cref{tab:per-user-scores}. For comparison to models on the same user-study split, see the appendix \Cref{tab:human-model-comparison}. We observe that humans primarily lose points from redacting imprecisely and more than necessary, which we will establish in the next sections.

\textbf{Diverse Redaction Strategies: }Under our benchmark's operating assumptions, contextual spans are undecidable under the definitions of contextual integrity alone, barring the existence of additional information. To understand how our users resolve ambiguous entities, we measured their agreement using Krippendorff's alpha \citep{krippendorffalpha2013}. This metric ensures accuracy by filtering out instances where independent labelers agreed purely by chance. We compute Krippendorff's alpha across all redaction units with two or more ratings, using all windows with nonzero \metric values.

We assess inter-annotator disagreement at two levels: a global Krippendorff's $\alpha$ capturing overall reliability, and a per unit-type mean pairwise disagreement $(D_t)$ localizing where disagreement concentrates in the label space. The global $\alpha$ is $0.540$ across all 7627 units with two or more ratings. This moderate aggregate value for $\alpha$ indicates that redaction decisions are not unimodal across our population, but it leaves open the localization question that our per-unit-type decomposition addresses.

\begin{figure}[htbp]
    \centering
    \begin{subfigure}[b]{0.4\linewidth}
        \includegraphics[width=\linewidth]{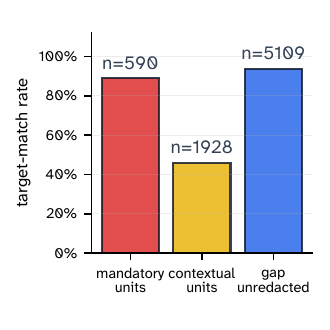}
        \caption{target-match rates}
        \label{subfig:agreement-rates}
    \end{subfigure}
    \hspace{0.05\linewidth}
    \begin{subfigure}[b]{0.4\linewidth}
        \includegraphics[width=\linewidth]{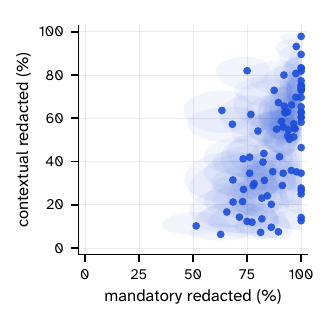}
        \caption{per-user redaction rates}
        \label{subfig:rates-per-user}
    \end{subfigure}
    \caption{
        (a) Mean target-match rate by unit type over all study windows, restricted to units exposed to at least two distinct users. For each qualifying unit, we compute the fraction of exposed user-windows matching the target action---redact mandatory/contextual units and leave gap units visible---then average those fractions within type; bar labels give qualifying-unit counts.
        (b) Per-user redaction rates under the same qualifying-unit filter. Each point's coordinates are the user's mandatory and contextual redaction rates; shaded ellipses show Wilson 95\% confidence intervals for those two rates.
    }
    \label{fig:agreement}
\end{figure}

To isolate this disagreement, we start by comparing rates across unit-types (redacted mandatory, contextual and unredacted gaps). As shown in~\Cref{subfig:agreement-rates}, across all redaction units, users redacted mandatory units at a rate of $89.4\%$, contextual units at a rate of $47.7\%$, and left gaps $94.1\%$ of the time. We attribute disagreement to context-sensitive units because humans have fundamentally different perspectives on privacy. While analyzing the mandatory-to-contextual redaction rate ratios for each user in~\Cref{subfig:rates-per-user}, we observe stark differences in redaction strategies that emerge only across contextual labels: some redact all contextual entities, while others redact hardly any. While the rates define average user behavior, they do not measure user disagreement. A unit-type with a 50\% rate would mean that every user redacted half the units of that type (a coin flip), or that half the users always redacted and half never did (a sharp split). To distinguish these, we report per-unit-type mean pairwise disagreement $D_t$. $D_t$ equals the per-item disagreement quantity in~\citet{fleiss1971} and the per-unit observed-disagreement contribution in~\citet{krippendorffalpha2013}; reporting raw pairwise disagreement per stratum is established practice in NLP work on annotator disagreement and human label variation~\citep{plank2022,uma2021,pavlick2019}. We present the complete formulation for these statistical measures in~\Cref{sec:alpha} and ~\Cref{sec:dt} along with a worked example in~\Cref{sec:worked-example}.

\Cref{tab:krippendorffs} shows that contextual entities have $1.80\times$ the disagreement of mandatory entities and $3.22\times$ that of gaps. We find that users agree substantially more on mandatory entities and gaps than on contextual entities. The residual disagreement rate on mandatory and gap units is asymmetric and predictable---on mandatory entities, it reflects under-redaction ($\sim 10\%$), and on gaps, it reflects over-redaction ($\sim 6\%$). Both are systematic in opposite directions and consistent with users applying a coherent, yet conservative, policy. Disagreement on contextual entities, by contrast, reflects significantly different privacy strategies rather than execution noise---consistent with the bimodal per-user redaction patterns in~\Cref{subfig:rates-per-user}.

Per-tier $\alpha$ values in~\Cref{tab:krippendorffs} are suppressed for mandatory and gap entities and explained by the prevalence paradox~\citep{feinstein1990,byrt1993}: when within-tier marginals are extreme, the chance baseline collapses and $\alpha$ understates rater consistency. We discuss this failure mode in~\Cref{subsec:per-tier-alpha} and report $\alpha_t$ as a negative control to make the failure mode apparent. Our analysis above verifies that (1) no unimodal distribution captures our users' intrinsic redaction preferences, and (2) \benchmark captures diversity in user redaction preferences precisely in its contextually sensitive entities.

\section{Limitations and Future Work}
\label{sec:limitations-and-future-work}
Currently, \benchmark evaluates documents in isolation, without considering surrounding context, such as user queries, conversation history, or system prompts---benign information in one context can become sensitive in another. Query-aware redaction is an exciting direction to pursue for the next version of \benchmark, providing a basis for more realistic context-conditioned evaluation. Acknowledging the possibility of co-referenced and re-identifiable entities is an important evaluation axis for future versions of \benchmark. Re-introducing structured documents to the user study with domain experts can provide additional insights into entities relevant to agentic heavy-use cases and growing fields such as coding and software automation. Finally, \benchmark lacks a formal analysis of the fidelity and irregularity rate of our synthetic entities, which would further strengthen its robustness.

In conclusion, the success of our "mandatory-contextual" labeling scheme captures the correct notions of human privacy and demonstrates that CI is a robust principle. Notable model-relevant takeaways from this work include frontier LLMs outperforming the average human and encoder-only models serving as reliable evaluators for privacy redaction tasks. We hope that \benchmark provides a standardized basis for modeling privacy with Contextual Integrity for future redaction work and encourages efficiency research in model design to drive breakthroughs beyond the currently established Pareto-optimal frontier. We acknowledge that recent work, such as \citet{openai2026privacyfilter}, is a meaningful step in that direction.

\clearpage
\bibliographystyle{plainnat}
\bibliography{ref}

\newpage
\appendix
{\LARGE\bfseries Appendix}
\vspace{1em}

\section{Contextual Integrity for Language Model Interactions}
\label{section:contextual-integrity}
Contextual Integrity (CI) is a theory of information transfer that considers social norms and circumstances before making privacy decisions \citep{nissenbaum2004privacy}. Of particular interest to our investigation is CI for AI assistants. Messages with AI assistants are typically sent over the network, which introduces significant risks in heavily regulated domains.

Many privacy concerns are not addressable solely with standard redaction techniques, but in certain circumstances, we can outline their use cases. Using CI, we can narrow our investigation of redaction privacy in the context of the following assumptions:
\begin{enumerate}
    \item \textit{Feasibility}: There exists a redacted version of the document that does not violate any policies while still maintaining positive utility.
    \item \textit{Permission}: Users can access all the information an assistant has access to, but the opposite does not hold.
    \item \textit{Snoop-free and Stateless}: The service or any downstream adversary is stateless and cannot compile additional information using external sources.
\end{enumerate}


\section{\benchmark Dataset Insights}
\label{sec:data_insights}
\begin{table}[h]
\centering
\caption{Composition of \benchmark}
\begin{tabular}{llll}
    \hline
    \textbf{Category} & \textbf{Count} & \textbf{Source} & \textbf{Structure} \\
    \hline
    Academic    & 16 & Real       & Unstructured \\
    Emails      & 16 & Synthetic  & Unstructured \\
    Financial   & 18 & Real       & Unstructured \\
    Government  & 16 & Real       & Unstructured \\
    Legal       & 18 & Real       & Unstructured \\
    Medical     & 16 & Real       & Unstructured \\
    Operations  & 15 & Real       & Unstructured \\
    Code        & 16 & Synthetic  & Structured   \\
    Files       & 36 & Mixed      & Structured   \\
    Logs        & 17 & Synthetic  & Structured   \\
    Terminal    & 16 & Synthetic  & Structured   \\
    \hline
\end{tabular}
\end{table}
Our definition of contextual integrity in the previous section has implications for coding assistants in \benchmark. For documents with code, concerns regarding intellectual property are assumed to have already been handled by the administrators. Thus, if context is strictly code-related, the goal of redaction is limited to removing secrets and preventing doxxing, since actions taken locally can pose risks of revealing contributor identities.

Structured formats in \benchmark follow a well-defined syntax that allows them to be parsed programmatically (code, logs, terminal output, files). Alternatively, unstructured formats convey information in natural language, either as free-form prose or as standardized documents and web forms. We rely on real documents wherever possible and leverage synthetic samples when suitable real-world examples are hard to find  (\Cref{tab:synthetic-justifications}), resulting in a final balanced corpus of 101 real and 99 synthetic documents.

\begin{table*}[!ht]
\centering
\scriptsize
\setlength{\tabcolsep}{4pt}
\renewcommand{\arraystretch}{1.08}
\caption{\benchmark's \emph{unstructured} categories and the precise constituent genre names. Italicized text represents synthetically generated genres, meaning they were created without a reference}
\label{tab:unstructured-genres}
\begin{tabular}{@{}lllll@{}}
\toprule
\textbf{Category} & \multicolumn{4}{c}{\textbf{Genres}} \\
\midrule
\textbf{academic} & academic\_probation\_appeal & course\_syllabus & diploma & disciplinary\_record \\
 & enrollment & fafsa & financial\_aid\_award\_letter & grade\_appeal \\
 & graded\_assignment & recommendation\_letter & report\_card & research\_agreement \\
 & student\_grievance & title\_ix & transcript & transcript\_request \\
\addlinespace[2pt]
\textbf{emails} & \textit{bank\_alert} & \textit{hr\_onboarding} & \textit{internal\_creds} & \textit{job\_application} \\
 & \textit{legal\_thread} & \textit{medical\_referral} & \textit{meeting\_invite} & \textit{newsletter\_announce} \\
 & \textit{order\_receipt} & \textit{password\_reset} & \textit{personal\_social} & \textit{phishing\_lure} \\
 & \textit{robotarium\_newsletter} & \textit{server\_alert} & \textit{shipping\_confirm} & \textit{travel\_itinerary} \\
\addlinespace[2pt]
\textbf{financial} & 1040 & account\_application & acord\_claim\_form & bank\_statement \\
 & bitcoin\_wallet\_export & card\_statement & chargeback\_rebuttal\_letter & deposit\_slip \\
 & direct\_deposit\_authorization\_form & flight & invoice & k1 \\
 & receipt & remittance\_advice & signature\_card & sim\_registration \\
 & w9 & wire\_transfer\_request &  &  \\
\addlinespace[2pt]
\textbf{government} & birth\_certificate & building\_permit & census\_form & customs\_declaration \\
 & drivers\_license & form\_i\_589 & jury\_summons & police\_report \\
 & public\_records\_request & social\_security & tax\_lien & vaccination \\
 & vehicle\_title\_registration & visa & voter\_registration\_card & workers\_compensation\_claim \\
\addlinespace[2pt]
\textbf{legal} & bail\_bond & bylaws & cease\_desist & certificate\_of\_incorporation \\
 & declarations & ds\_160 & ds\_2019 & employment \\
 & i765\_ead & i9 & lease\_agreement & marital\_settlement \\
 & offer\_letter & pip & prenuptial & promissory\_note \\
 & qdro & sofa &  &  \\
\addlinespace[2pt]
\textbf{medical} & behavioral\_health\_safety\_plan & break\_the\_glass & cms\_1500 & confinement\_form \\
 & consultation\_note & discharge\_summary & ecg\_tracing & explanation\_of\_benefits \\
 & mds\_3 & operative\_report & prescription & prior\_authorization\_request\_form \\
 & psychotherapy\_notes & radiology\_report & sterilization\_consent & triage\_note \\
\addlinespace[2pt]
\textbf{operations} & benefits\_enrollment & bill\_of\_lading & boarding\_pass & hotel\_reservation \\
 & interview\_schedule & offer\_letter & onboarding\_packet & pay\_stub \\
 & proof\_of\_delivery & property\_tax & resignation\_letter & session\_transcript \\
 & shipping\_label & transfer\_notice & uber\_receipt &  \\
\bottomrule
\end{tabular}
\end{table*}

\begin{table*}[!ht]
\centering
\caption{\benchmark's \emph{structured} categories and the precise constituent genre names. Italicized text represents synthetically generated genres, meaning they were created without a reference}
\resizebox{\textwidth}{!}{%
\label{tab:redactionbench-structured-genres}
\begin{tabular}{@{} l l l l l @{}}
\toprule
\textbf{Category} & \multicolumn{4}{c}{\textbf{Genres}} \\
\midrule
\textbf{code} & \textit{bashrc} & \textit{c\_program} & \textit{css\_file} & \textit{dockerfile} \\
 & \textit{github\_actions\_ci} & \textit{go\_service} & \textit{html\_page} & \textit{java\_properties} \\
 & \textit{javascript\_config} & \textit{makefile} & \textit{php\_config} & \textit{python\_db\_script} \\
 & \textit{ruby\_config} & \textit{rust\_main} & \textit{sql\_seed} & \textit{terraform\_main} \\

\textbf{files} & \textit{browser\_storage} & \textit{crypto\_wallet\_export} & \textit{csv\_api\_keys\_export} & \textit{csv\_financial} \\
 & \textit{csv\_identity\_export} & \textit{csv\_server\_inventory} & \textit{dotenv} & \textit{exchange\_api\_config} \\
 & \textit{gpg\_export} & \textit{hl7\_adt\_message} & \textit{htpasswd} & \textit{ics\_calendar} \\
 & \textit{json\_api\_response} & \textit{json\_aws\_config} & \textit{json\_package\_lock} & \textit{jwt\_tokens} \\
 & \textit{ldap\_dump} & \textit{mrz\_passport\_batch} & \textit{ndjson\_app\_errors} & \textit{ndjson\_audit\_log} \\
 & \textit{ndjson\_cloudtrail} & \textit{password\_manager\_csv} & \textit{recovery\_keys} & \textit{ssh\_config} \\
 & \textit{terraform\_tfvars} & \textit{toml\_config} & \textit{toml\_pyproject} & \textit{xml\_incident\_report} \\
 & \textit{xml\_junit\_results} & \textit{xml\_maven\_pom} & \textit{xml\_rss\_feed} & \textit{yaml\_ansible\_playbook} \\
 & \textit{yaml\_docker\_compose} & \textit{yaml\_k8s\_secret} & loan\_agreement & vcf \\

\textbf{logs} & \textit{apache\_combined\_log} & \textit{app\_structured\_json} & \textit{auth\_syslog} & \textit{blockchain\_node\_log} \\
 & \textit{ci\_runner\_log} & \textit{cloudtrail\_json} & \textit{docker\_daemon\_log} & \textit{firewall\_log} \\
 & \textit{gcp\_audit\_log} & \textit{kubernetes\_events} & \textit{mysql\_slow\_query} & \textit{nginx\_access\_log} \\
 & \textit{postgresql\_audit\_log} & \textit{redis\_log} & \textit{smtp\_log} & \textit{vpn\_log} \\
 & \textit{windows\_event\_log} &  &  &  \\

\textbf{terminal} & \textit{ansible\_deploy\_session} & \textit{aws\_cli\_session} & \textit{ci\_cd\_runner\_session} & \textit{core\_dump\_session} \\
 & \textit{database\_admin\_session} & \textit{docker\_compose\_session} & \textit{env\_leak\_session} & \textit{git\_history\_session} \\
 & \textit{incident\_response\_session} & \textit{kubernetes\_session} & \textit{nvtop\_monitoring\_session} & \textit{package\_install\_session} \\
 & \textit{python\_traceback\_session} & \textit{reverse\_shell\_forensics} & \textit{ssh\_tunneling\_session} & \textit{sysadmin\_recon\_session} \\
\bottomrule
\end{tabular}%
}
\end{table*}

\newpage
\begin{table}[!ht]
    \centering
    \caption{Justifications for synthetic samples per domain}
    \label{tab:synthetic-justifications}
    \begin{tabularx}{\textwidth}{@{} l X @{}}
        \toprule
        \textbf{Domain} & \textbf{Justification} \\
        \midrule
        \textbf{Emails} & We considered various classic email datasets such as the Enron emails and even controversial contemporary piles, but these human-authored emails tend to be short, terse, low in complexity, and seldom contain sensitive content. Furthermore, a large number are auto-generated, no-reply emails. \\ \addlinespace
        
        \textbf{Code} & While many examples of code can be compiled readily, they rarely contain sensitive information, and most cannot be trivially augmented into plausible samples. To achieve a threshold of realism, we generated samples in which sensitive information, such as API keys, was more likely. With the popularity of coding assistants, synthetically generated code is also an increasingly large portion of real code. \\ \addlinespace
        
        \textbf{Files} & We used synthetic file data to avoid exposing or reproducing real PII while still capturing the structure of common machine-readable formats such as JSON, XML, YAML, etc. \\ \addlinespace
        
        \textbf{Logs} & All public log datasets we found contained only benign content. Therefore, similar to code, we generate our own log data to increase the density of sensitive information. \\ \addlinespace
        
        \textbf{Terminal} & Similar to code. \\
        \bottomrule
    \end{tabularx}
\end{table}

\newpage
\begin{table*}[!ht]
\centering
\scriptsize
\setlength{\tabcolsep}{4pt}
\renewcommand{\arraystretch}{1.08}
\caption{Observed "Giveaway" Patterns in Augmented Entities: A tabulation of the faked entity patterns which we systematically corrected within \benchmark's synthetically augmented entities. These are patterns we noticed in many other benchmarks, especially when freeform generation is involved.}
\label{tab:synthetic-pitfalls}
\begin{tabular}{@{}p{0.24\linewidth}>{\raggedright\arraybackslash}p{0.72\linewidth}@{}}
\toprule
\textbf{Pattern} & \textbf{Description and example matches} \\
\midrule
\texttt{overrepresented\_entities} & Natural entities reused across too many documents, including names, surnames, organizations, domains, handles, localparts, and address or place names (e.g., \mbox{\texttt{Priya}}; \mbox{\texttt{Dmitri}}; \mbox{\texttt{Fatima}}; \mbox{\texttt{Okafor}}; companies with \mbox{\texttt{Harbor*}} or \mbox{\texttt{Shore*}} in their names). \\
\texttt{low\_entropy\_literal} & Explicit placeholder or toy fragments, including cases where the fragment appeared inside a longer identifier, URL, secret, or file path (e.g., \mbox{\texttt{abc123}}; \mbox{\texttt{1a2b3c}}; \mbox{\texttt{a1b2c3}}; \mbox{\texttt{qwerty}}; \mbox{\texttt{asdf}}; \mbox{\texttt{zxcv}}; \mbox{\texttt{deadbeef}}). \\
\texttt{alphabetic\_sequence\_run} & Ascending, descending, wrapped, or mixed-case alphabetic runs, including ordered material embedded in longer tokens (e.g., \mbox{\texttt{ABCDEFGHIJKLMNOPQRSTUVWXYZ}}; \mbox{\texttt{abcdefghijklmnopqrstuvwxyz}}; \mbox{\texttt{AbCdEfGh}}; \mbox{\texttt{ZyXwVuTsRqPoNmLk}}). \\
\texttt{sequential\_numeric\_run} & Ascending or descending digit runs, including those embedded in IDs, account numbers, URLs, or secrets (e.g., \mbox{\texttt{123456}}; \mbox{\texttt{0987654321}}; \mbox{\texttt{DE75500700100987654321}}). \\
\texttt{digit\_permutation\_run} & Length-six-or-longer digit strings where all digits are unique, suggesting sampled-without-replacement fake identifiers (e.g., \mbox{\texttt{487321}}; \mbox{\texttt{1048576}}; \mbox{\texttt{81520467}}; \mbox{\texttt{3902718456}}). \\
\texttt{interleaved\_numeric} & Long digit strings where alternating positions form arithmetic ladders (e.g., \mbox{\texttt{13572468}}; \mbox{\texttt{24681357}}). \\
\texttt{alternating\_alnum\_run} & Repeated letter-digit or digit-letter alternation, case-insensitive. Prefix-bearing keys, hex-like strings, and structured business identifiers were handled here when the payload had this structure (e.g., \mbox{\texttt{1a2b3c4d}}; \mbox{\texttt{0a6b7c8d9e0f1a2b}}; \mbox{\texttt{0x9y8z7w6v5u4t3s}}; \mbox{\texttt{TXN-20260211-9F4K2M7Q1R}}). \\
\texttt{period3\_alnum\_run} & Period-three alphanumeric cycles such as lower/upper/digit or digit/lower/upper (e.g., \mbox{\texttt{aB1cD2eF3}}; \mbox{\texttt{1aB2cD3eF}}). \\
\texttt{repeated\_char\_run} & Repeated-character padding or placeholder material, including repeated runs embedded in longer tokens (e.g., \mbox{\texttt{00000}}; \mbox{\texttt{aaaaa}}; \mbox{\texttt{XXXXX}}). \\
\texttt{fake\_555\_triplet} & Phone numbers or phone-like spans containing the fake 555 exchange (e.g., \mbox{\texttt{555}}; \mbox{\texttt{(312) 555-9481}}). \\
\texttt{top\_of\_hour\_time} & Templated clock times with zero minutes and seconds, typically betraying themselves through repeated top-of-hour entries within the same log (e.g., \mbox{\texttt{14:00:00}}; \mbox{\texttt{15:00:00}}; \mbox{\texttt{16:00:00}}). \\
\texttt{top\_of\_hour\_compact\_utc} & Compact UTC timestamps ending in top-of-hour 0000Z, flagged when several such values appear together in the same record (e.g., \mbox{\texttt{20260314140000Z}}; \mbox{\texttt{20260314150000Z}}). \\
\texttt{fake\_mac\_staircase} & MAC addresses whose octets form the synthetic 00:1A:2B:3C base or an arithmetic staircase with three-or-more consecutive 0x11 or 0xEF steps (e.g., \mbox{\texttt{00:1A:2B:3C:4D:5E}}; \mbox{\texttt{11:22:33:44:55:66}}). \\
\texttt{masked\_ssn\_last4\_pair} & Masked SSNs whose visible last four have repeated-pair structure, especially when several such values co-occur in one document (e.g., \mbox{\texttt{XXX-XX-1122}} together with \mbox{\texttt{XXX-XX-3344}}; \mbox{\texttt{XXX-XX-7788}} together with \mbox{\texttt{XXX-XX-9900}}). \\
\texttt{pattern\_match} & Tokens identified by their structural shape alone—a syntactic prefix, delimiter, or recognizable surface form. The matched payload is re-randomized unconditionally as a safety net for any low-entropy fragment the content-based detectors above might have missed inside. Covers UUIDs; debug-hex literals starting with 0xDEAD, 0xBEEF, 0xC0DE, or 0xBADF00D; req\_ and evt\_ identifiers; bcrypt and SHA256 fingerprints; txid= and addr= assignments; bech32 Bitcoin addresses; WIF private keys; and ASCII-armored key or certificate blocks (e.g., \mbox{\texttt{a1b2c3d4-e5f6-7890-abcd-ef1234567890}}; \mbox{\texttt{0xDEAD9F3A}}; \mbox{\texttt{req\_a1b2c3d4e5f6}}; \mbox{\texttt{evt\_AbCdEfGhIjKlMnOp}}; \mbox{\texttt{\$2b\$10\$Hcr5LxVz1rQwT0NaTbX2bO9pUaC8PwR3H6oG2QmK1FjV5sZeRfYqW}}; \mbox{\texttt{SHA256:AbCdEfGhIjKlMnOpQrStUvWxYz0123}}; \mbox{\texttt{txid=a1b2c3d4e5f6a7b8c9d0e1f2a3b4c5d6e7f8a9b0}}; \mbox{\texttt{bc1qxy2kgdygjrsqtzq2n0yrf2493p83kkfjhx0wlh}}; \mbox{\texttt{addr=1A2B3C4D5E6F7G8H9I0J1K2L3M}}; \mbox{\texttt{privkey WIF: KxFqZmNq3bR4sLkM7pW8nQ2vZyX6cT3aFqRzS5wYbVdLh}}; \mbox{\texttt{BEGIN PGP PUBLIC KEY BLOCK}}). \\
\bottomrule
\end{tabular}
\end{table*}

\clearpage
\section{User Study Details}
\label{sec:study-details}
We conduct the user study on Prolific \citep{prolific}, with 85 participants residing in the United States, aged 25-65, and holding at least a Bachelor's degree. Users were compensated at a mean rate of $\$13.08$/hr with a median completion time of 00:27:32. Users completed a brief tutorial to get comfortable with the mechanics of our redaction tool and were instructed to think critically about what units of text were sensitive, given the document type. Users were permitted to search the web to alleviate doubts regarding unfamiliar terms, fields, documents, or entities. 

We collected user consent for participation and timestamped edits. Although our target labels in \benchmark are ternary, users were only given the option to redact or not, without knowledge of the target labels. Each participant's edits were manually reviewed; four participants were rejected and resampled due to bad-faith attempts.

\subsection{Study Setup}
\label{subsec:study-setup}
Our user study was limited to \textit{unstructured} documents since \textit{structured} documents require domain expertise. From each of the seven \textit{unstructured} categories, we randomly selected eight representative documents, for a total of 56. Each document was viewed by 16 users, and each user was assigned 10-11 documents, ensuring broad coverage of each document in the user study. The assignment aims to minimize overlap so that no two users view the same set of documents. We further randomized the viewing window for each user along a separate axis: partial/incorrect/noisy initial redactions in the starting window \metric (see~\Cref{subfig:starting-iou-histogram}).

\begin{figure}[htbp]
\centering
\includegraphics[width=1.0\linewidth]{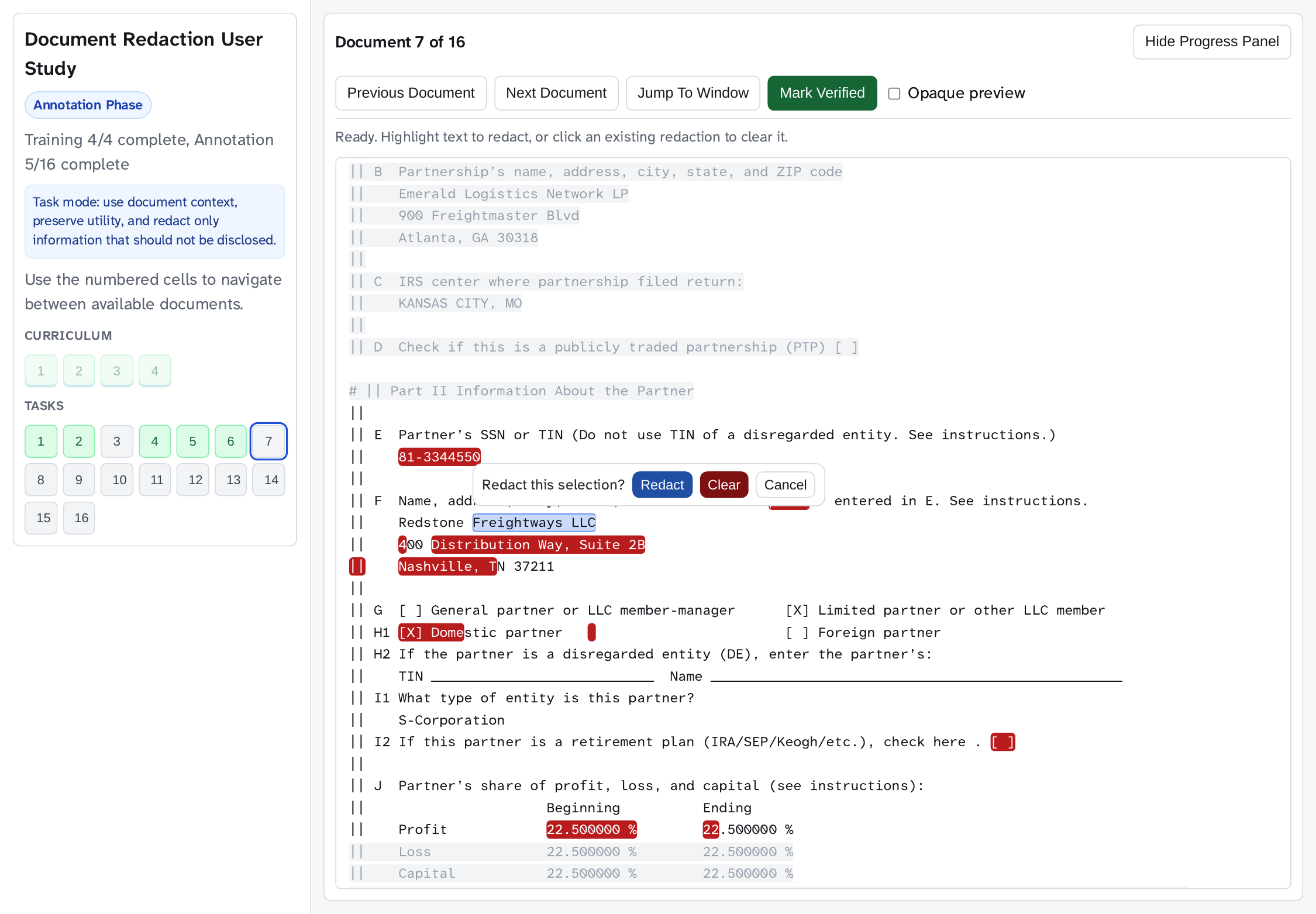}
\caption{Redaction Study Tool. User's view of a labeling window from our study tool. Grayed-out regions are not editable but provide global context---users can scroll within the edit window to view the rest of the document. Users can toggle an opaque preview to help focus on the state of the final document.}
\label{fig:user-study}
\end{figure}

To expedite the labeling process and minimize labeling fatigue, randomized windows were limited to 25 or fewer lines. The remaining context was greyed out but was available to the user for additional context (\Cref{fig:user-study}). Each user saw approximately the same distribution of starting window scores. We aggregated all windows with a starting score of 0.0---the leftmost, skinny column in \Cref{subfig:starting-iou-histogram}---to estimate human performance for an apples-to-apples comparison with the models (\Cref{tab:human-model-comparison}).

\begin{table}
\centering
\caption{Evaluations on the user study subset with 48 total documents for comparison. A significant gap remains between frontier models and specially-finetuned models. Frontier models match or exceed human performance on the benchmark.}
\label{tab:human-model-comparison}
\setlength{\tabcolsep}{3.5pt}
\renewcommand{\arraystretch}{0.9}
\providecommand{\pMetric}[1]{}
\renewcommand{\pMetric}[1]{{\fontsize{8pt}{9pt}\selectfont #1}}
\resizebox{\textwidth}{!}{%
\begin{tabular}{@{}l l r r r@{}}
    \toprule
    \textbf{Family} & \textbf{Variant} & \multicolumn{3}{c}{\textbf{Overall 48}} \\
    \midrule
     &  & \textbf{\pMetric{$P_{20}$}} & \textbf{\pMetric{$P_{50}$}} & \textbf{Mean} \\
    \midrule
    Frontier & gpt-5.4 & \pMetric{0.69} & \pMetric{0.90} & 0.81 \\
    Frontier & \nolinkurl{Qwen/Qwen3.5-397B-A17B} & \pMetric{0.66} & \pMetric{0.88} & 0.80 \\
    Frontier & claude-opus-4-6 & \pMetric{0.62} & \pMetric{0.89} & 0.79 \\
    Frontier & zai-org/GLM-5.1 & \pMetric{0.61} & \pMetric{0.92} & 0.77 \\
    \midrule
    Human & Aggregated performance (mode decision per label) on windows with starting \metric 0 & \pMetric{0.61} & \pMetric{0.78} & 0.77 \\
    \midrule
    OpenAI Privacy Filter & \nolinkurl{openai/privacy-filter} & \pMetric{0.36} & \pMetric{0.72} & 0.64 \\
    B2NER & \nolinkurl{internlm/internlm2-20b} & \pMetric{0.39} & \pMetric{0.63} & 0.60 \\
    DeBERTa & \nolinkurl{OpenMed/OpenMed-PII-SuperClinical-Large-434M-v1} & \pMetric{0.35} & \pMetric{0.65} & 0.59 \\
    GLiNER & \nolinkurl{gretelai/gretel-gliner-bi-large-v1.0} & \pMetric{0.39} & \pMetric{0.60} & 0.56 \\
    B2NER & \nolinkurl{internlm/internlm2_5-7b} & \pMetric{0.31} & \pMetric{0.56} & 0.54 \\
    GLiNER & \nolinkurl{nvidia/gliner-PII} & \pMetric{0.28} & \pMetric{0.58} & 0.52 \\
    SLM / extractor & \nolinkurl{eternisai/Anonymizer-4B} & \pMetric{0.04} & \pMetric{0.58} & 0.51 \\
    SLM / extractor & \nolinkurl{jakobhuss/pii-extractor-gemma-3-270m-it} & \pMetric{0.23} & \pMetric{0.56} & 0.51 \\
    GLiNER & \nolinkurl{E3-JSI/gliner-multi-pii-domains-v1} & \pMetric{0.20} & \pMetric{0.53} & 0.48 \\
    RoBERTa & \nolinkurl{iiiorg/piiranha-v1-detect-personal-information} & \pMetric{0.24} & \pMetric{0.48} & 0.47 \\
    DeBERTa & \nolinkurl{hydroxai/pii_model_weight} & \pMetric{0.32} & \pMetric{0.48} & 0.46 \\
    SLM / extractor & \nolinkurl{numind/NuExtract-1.5-tiny} & \pMetric{0.14} & \pMetric{0.47} & 0.44 \\
    GLiNER & \nolinkurl{urchade/gliner_multi_pii-v1} & \pMetric{0.19} & \pMetric{0.45} & 0.44 \\
    ModernBERT  & \nolinkurl{ai4privacy/llama-ai4privacy-english-anonymiser-openpii} & \pMetric{0.24} & \pMetric{0.45} & 0.43 \\
    SLM / extractor & \nolinkurl{numind/NuExtract-2.0-2B} & \pMetric{0.21} & \pMetric{0.43} & 0.43 \\
    SLM / extractor & \nolinkurl{Universal-NER/UniNER-7B-all} & \pMetric{0.22} & \pMetric{0.43} & 0.42 \\
    SLM / extractor & \nolinkurl{distil-labs/Distil-PII-Llama-3.2-3B-Instruct} & \pMetric{0.00} & \pMetric{0.40} & 0.42 \\
    GLiNER & \nolinkurl{knowledgator/gliner-pii-base-v1.0} & \pMetric{0.11} & \pMetric{0.42} & 0.41 \\
    SLM / extractor & \nolinkurl{numind/NuExtract-2.0-4B} & \pMetric{0.18} & \pMetric{0.32} & 0.36 \\
    DeBERTa & \nolinkurl{h2oai/deberta_finetuned_pii}\(^\dagger\) & \pMetric{0.04} & \pMetric{0.37} & 0.34 \\
    DeBERTa & \nolinkurl{lakshyakh93/deberta_finetuned_pii}\(^\dagger\) & \pMetric{0.04} & \pMetric{0.37} & 0.34 \\
    ModernBERT  & \nolinkurl{ai4privacy/llama-ai4privacy-multilingual-categorical-anonymiser-openpii} & \pMetric{0.15} & \pMetric{0.30} & 0.32 \\
    DistilBERT & \nolinkurl{Isotonic/distilbert_finetuned_ai4privacy_v2} & \pMetric{0.08} & \pMetric{0.28} & 0.32 \\
    RoBERTa & \nolinkurl{tanaos/tanaos-text-anonymizer-v1} & \pMetric{0.00} & \pMetric{0.30} & 0.29 \\
    GLiNER & \nolinkurl{hivetrace/gliner-guard-uniencoder} & \pMetric{0.11} & \pMetric{0.29} & 0.28 \\
    GLiNER & \nolinkurl{hivetrace/gliner-guard-biencoder} & \pMetric{0.08} & \pMetric{0.27} & 0.27 \\
    DistilBERT & \nolinkurl{deepaksiloka/PII-Detection-V2.1} & \pMetric{0.04} & \pMetric{0.20} & 0.25 \\
    GLiNER & \nolinkurl{urchade/gliner_multi-v2.1} & \pMetric{0.06} & \pMetric{0.22} & 0.23 \\
    SLM / extractor & \nolinkurl{distil-labs/Distil-PII-Llama-3.2-1B-Instruct} & \pMetric{0.00} & \pMetric{0.00} & 0.19 \\
    SLM / extractor & \nolinkurl{OpenPipe/PII-Redact-General} & \pMetric{0.00} & \pMetric{0.12} & 0.14 \\
    SLM / extractor & \nolinkurl{distil-labs/Distil-PII-gemma-3-270m-it} & \pMetric{0.00} & \pMetric{0.00} & 0.06 \\
    \bottomrule
\end{tabular}%
}
\footnotesize\(^\dagger\) These two public uploads share byte-identical Hugging Face artifacts and are counted as one unique checkpoint in model totals.
\end{table}

\subsection{Non-expert Annotators}
\label{subsec:humans-not-experts}
\begin{figure}[htbp]
    \centering
    \begin{subfigure}[b]{0.49\linewidth}
        \includegraphics[width=\linewidth]{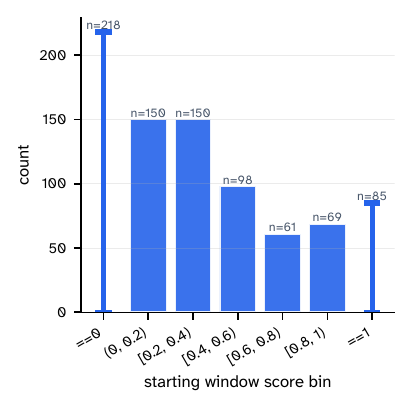}
        \caption{}
        \label{subfig:starting-iou-histogram}
    \end{subfigure}
    \begin{subfigure}[b]{0.49\linewidth}
        \includegraphics[width=\linewidth]{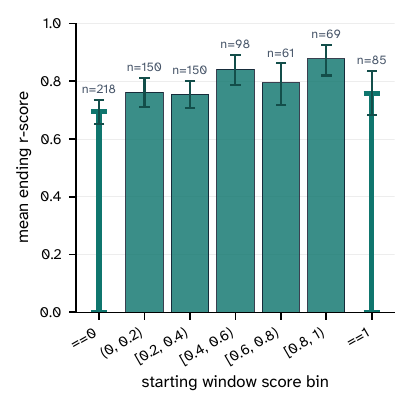}
        \caption{}
        \label{subfig:mean-end-score}
    \end{subfigure}
    \caption{
        (a) The distribution of initial conditions, R-Score, for each redaction window with endpoints excluded. (b) An estimation of the mean \metric after user edits with bins based on starting score. Confidence intervals are estimated at 95\% using bootstrapping with $n=2000$. We note a significant positive correlation as more target label information is injected---this is a coarse view of the underlying continuous distribution for which we computed Spearman's rho in \Cref{subsec:humans-not-experts}.
    }
    \label{fig:p-improvement-and-mean-end-score}
\end{figure}

We define a requisite property of an expert to be invariance to starting conditions. Since a redaction configuration can be achieved from a random initial state, we assume that an expert redactor will always be able to optimize our metric with the right redactions. To test the objectivity of our users, we randomize the initial state of each window with varying levels of initial redactions (\Cref{subfig:starting-iou-histogram}) and observe the resulting scores in each condition (\Cref{subfig:mean-end-score}).

We tested whether partially revealed target labels bias users towards the set of solutions favored by our metric. We used a one-sided Spearman rank correlation test across all 831 windows and measured a significant positive relationship ($\rho=0.158$, $p=2.4\times10^{-6}$) between starting and ending scores (\Cref{subfig:mean-end-score}). For a detailed explanation about the Spearman rank correlation test refer~\Cref{sec:spearman}. In conclusion, our users are not experts, as they are influenced by initial configurations.

We therefore limit our estimate of human performance to windows with an initial \metric of 0.0 to avoid our partial redactions from biasing final statistics. We report a per-category breakdown of human scores across the unstructured categories in \Cref{tab:per-user-scores}. Furthermore, we evaluated models on the same split of unstructured documents presented to our users, as shown in \Cref{tab:human-model-comparison}.

Windows with non-zero starting conditions reveal additional information about human behavior and the performance threshold of our benchmark, which we cover in the next section.

\subsection{Performance Threshold on \benchmark}
\label{sec:properties-of-rscore}
\begin{figure}[htbp]
    \centering
    \includegraphics[width=1.0\linewidth]{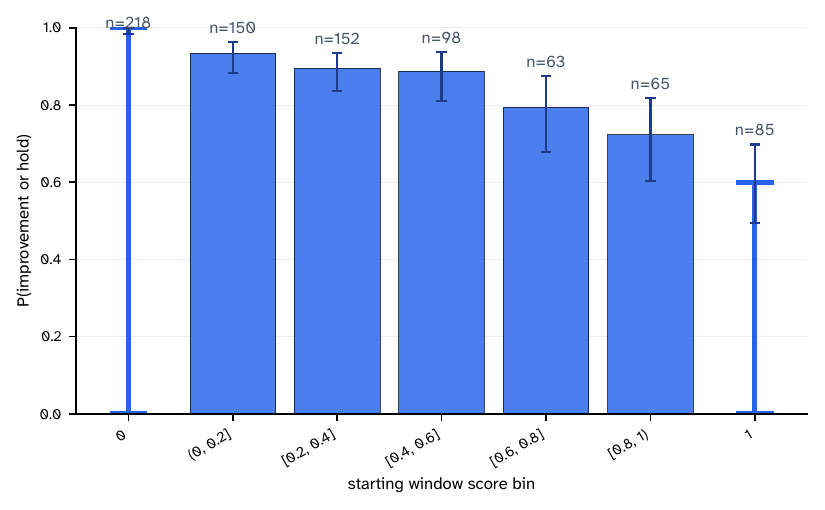}
    \caption{Probability of non-decrease in \metric after user edits as a function of the starting annotation window's \metric. Separate bins at 0.0 and 1.0 performance are shown as thin bars. To estimate the confidence in proportion, we used Wilson's 95\% binomial confidence intervals.}
    \label{fig:p-improvement}
\end{figure}

\label{subsec:smoothness-directional}
With our samples from diverse starting conditions, we can estimate the threshold at which participants may no longer agree with the \metric. This is a useful quantity to estimate the maximum score a model is expected to achieve. We observe a reversal in \metric values driven by our gold labeler's decisions beyond the saturation threshold.

In~\Cref{fig:p-improvement}, we estimate the probability that a user improved or maintained their window's \metric and the expected final score after all edits, including special bins for  0.0 and 1.0. Because the space of possible redactions is large, maintaining a high probability that \metric does not decrease as user annotations approach the gold labels can serve as a proxy for the quality of our labels. 

This confirms that there is \textit{no threshold} at which the median participant disagrees with the \metric, thus median performance should be near 1.0 on the user study subset.

\subsection{Smoothness of \metric and Edit-Trajectories}
\label{subsec:attractive-metric}

\begin{figure}[htbp]
\centering
\includegraphics[width=1.0\linewidth]{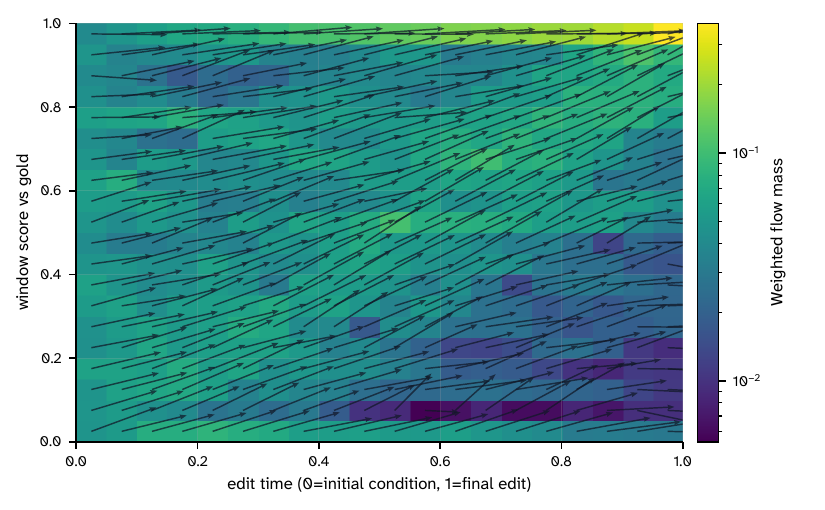}
\caption{Evolution of the window score distribution as a fraction of user edits made.}
\label{fig:privacy-flow}
\end{figure}

We derive additional dynamics from the sequence of edits for each user across all initial conditions. We can assume that each user was optimizing an unknown internal privacy function, with each local edit performing a "gradient ascent" on the hypothetical function, since the participants received no information about the metric. We plot the evolution of the score distribution over edit-time (proportion of edits made) in~\Cref{fig:privacy-flow}.

We see evidence corroborating the analysis in~\Cref{subfig:mean-end-score}: windows that start with higher \metric are more likely to fully maximize the score with user edits. Along with the property that our score does not saturate, this shows that performance exceeding our human baseline or the frontier model baseline \Cref{tab:human-model-comparison} may both be achievable and desirable.

\subsubsection{Diagram Construction} 
The x-axis is \textit{normalized edit time}: 0 is the initial condition, each edit is placed at \texttt{edit\_step} / \texttt{total\_edit\_count}, and 1 is the final state. The y-axis is the window score. \textit{Only trajectories with at least three annotation edits and at least two valid score observations are included.} Trajectories are weighted inversely by starting-score bin density~\Cref{subfig:starting-iou-histogram}. For each point, the flow direction is computed toward the best score reachable within the next two edit observations, then normalized to a unit vector. Color indicates where edit trajectories are concentrated, while arrows indicate the typical near-term direction of score movement from those regions.

\section{Spearman's Rank Correlation Coefficient}
\label{sec:spearman}

We use Spearman's Rank Correlation Coefficient to predict whether a window's starting \metric predicts its ending \metric (\Cref{subsec:humans-not-experts}). Spearman's coefficient, which tests whether one ordinal or continuous variable predicts another, makes no assumptions about linearity or normality and is robust to outliers.

\subsection{Formal Definition}
\label{subsec:spearman-def}

For \(n\) paired observations \((x_i, y_i)\), let \(R(x_i)\) and \(R(y_i)\) denote the ranks of \(x_i\) and \(y_i\) within their respective samples. Spearman's \(\rho\) is the Pearson correlation of the ranks:
\begin{align*}
\rho = \frac{\sum_i \left(R(x_i) - \bar{R}_x\right)\left(R(y_i) - \bar{R}_y\right)}{\sqrt{\sum_i \left(R(x_i) - \bar{R}_x\right)^2 \, \sum_i \left(R(y_i) - \bar{R}_y\right)^2}}.
\end{align*}

When there are no ties in either variable, this simplifies to:
\begin{align*}
\rho = 1 - \frac{6 \sum_i d_i^2}{n(n^2 - 1)},
\end{align*}
where \(d_i = R(x_i) - R(y_i)\) is the rank difference for pair \(i\). \(\rho\) is bounded in \([-1, 1]\): \(\rho = 1\) and \(\rho = -1\) indicate perfect monotonically increasing and decreasing relationships respectively, and \(\rho = 0\) indicates no monotonic relationship.

\subsection{Significance Test}
\label{subsec:spearman-pvalue}

Given two variables that are independently random (\(\rho = 0\)), the test statistic
\begin{align*}
t = \rho \sqrt{\frac{n-2}{1-\rho^2}}
\end{align*}
is approximately \(t\)-distributed with \(n-2\) degrees of freedom for sufficiently large \(n\) (typically \(n \geq 10\)). The two-sided \(p\)-value is \(p = 2 \cdot \Pr(T_{n-2} \geq |t|)\); the one-sided \(p\)-value---appropriate when the alternative hypothesis specifies the direction of the relationship, as in~\Cref{subsec:humans-not-experts} where impressionability predicts a \emph{positive} association between starting and ending \metric---halves this. For small \(n\), exact \(p\)-values via permutation tests over all \(n!\) rank arrangements are tractable; \texttt{scipy.stats.spearmanr} \citep{2020-scipy}, which we use, automatically selects the appropriate method.

\section{Wilson Score Intervals for Binomial Proportions}
\label{sec:wilson-ci}

When estimating a binomial proportion---such as the probability of improvement---we report Wilson's 95\% confidence intervals~\citep{wilson1927}. This interval remains well-behaved for proportions near 0 or 1, rather than overshooting as would occur with a standard normal estimate.

For \(k\) successes in \(n\) trials, let \(\hat{p}=k/n\) and let \(z=1.96\) denote the standard-normal 97.5th percentile (2.5\% probability tails on \textit{each} side). The Wilson interval is:
\begin{align*}
\frac{\hat{p}+\frac{z^2}{2n}}{1+\frac{z^2}{n}}
\;\pm\;
\frac{z}{1+\frac{z^2}{n}}
\sqrt{\frac{\hat{p}(1-\hat{p})}{n}+\frac{z^2}{4n^2}}.
\end{align*}

\section{Krippendorff's Alpha}
\label{sec:alpha}

Because the vast majority of tokens in a document are unredacted, a random labeler could achieve an artificially high agreement rate simply by predicting ``safe text.'' Krippendorff's \(\alpha\) corrects for this by establishing a ratio of observed disagreement to expected chance disagreement~\citep{krippendorffalpha2013}. The resulting \(\alpha\) coefficient ranges from \(1.0\) (perfect reliability) to \(0.0\) (no reliability, equivalent to random noise), with negative values indicating systematic disagreement.

\subsection{Formal Definition}
\label{subsec:alpha-def}

For a coding task with \(n\) ratings spread across qualifying units (units with \(\geq 2\) raters):
\begin{align*}
\alpha &= 1 - \frac{D_o}{D_e}, \\
D_o &= \frac{1}{n} \sum_u \frac{2 \, n_{u0} \, n_{u1}}{m_u - 1}, \qquad
D_e = \frac{2 \, n_0 \, n_1}{n(n-1)},
\end{align*}
where \(m_u\) is the number of raters on unit \(u\), \(n_{uc}\) is the count of class \(c\) on unit \(u\), and \(n_c\) is the global marginal across all qualifying units. This formulation accommodates variable \(m_u\) (units have different rater counts due to windowing).

\subsection{Per-Unit-Type alpha and the Prevalence Paradox}
\label{subsec:per-tier-alpha}

A natural question is whether the global \(\alpha\) can be stratified by target unit-type to localize where disagreement concentrates. The within-subset variant restricts both \(D_o\) and \(D_e\) to a unit-type \(t\):
\begin{align*}
\alpha_t &= 1 - \frac{D_o(t)}{D_e(t)}, \\
D_o(t) &= \frac{1}{n_t} \sum_{u\,:\,t(u)=t} \frac{2\,n_{u0}\,n_{u1}}{m_u-1}, \qquad
D_e(t) = \frac{2\,n_0(t)\,n_1(t)}{n_t(n_t-1)}.
\end{align*}

This stratification fails under the \emph{prevalence paradox}~\citep{feinstein1990,cicchetti1990,byrt1993}: when within-unit-type marginals are extreme, the chance baseline \(D_e(t)\) collapses, suppressing \(\alpha_t\) even when raw agreement is high. Mandatory entities (Red) skew toward redaction; benign gaps (Gap) skew toward non-redaction. Both classes, therefore, fail under \(\alpha_t\) for structural rather than empirical reasons. We report \(\alpha_t\) as a negative control to make the failure mode visible and motivate the alternative metric below.

\section{Per-Unit-Type Mean Pairwise Disagreement}
\label{sec:dt}

\subsection{Formulation}

We report the \emph{mean pairwise disagreement} per unit-type, to compute the stratified disagreement across each of the three unit-types. For unit \(u\) with \(m_u\) raters and \(k_u\) redactions:
\begin{align*}
D_u = \frac{k_u(m_u - k_u)}{\binom{m_u}{2}}, \qquad
D_t = \frac{1}{N_t} \sum_{u\,:\,t(u)=t} D_u,
\end{align*}
where \(N_t\) is the number of qualifying units in unit-type \(t\). \(D_u\) has two equivalent interpretations. \emph{Combinatorially,} it is the fraction of distinct rater pairs on unit \(u\) that disagree: of the \(\binom{m_u}{2}\) pairs, exactly \(k_u(m_u - k_u)\) consist of one redactor and one non-redactor. \emph{Probabilistically,} if raters vote independently with redaction probability \(p_u\), then \(D_u\) is the unbiased estimator of \(2p_u(1-p_u)\) --- the chance a random pair disagrees, maximized at \(p_u = 0.5\). High \(D_u\) thus identifies units where the rater pool splits near 50/50; low \(D_u\) identifies near-consensus in either direction.

This quantity is well-studied under different names: it equals the per-item disagreement \(1 - P_i\) in~\citet{fleiss1971}, the per-unit observed-disagreement contribution in~\citet{krippendorffalpha2013}, and the Gini impurity of the unit's decision distribution. Reporting raw pairwise disagreement per stratum follows established practice in NLP work on annotator disagreement and human label variation~\citep{plank2022,uma2021,pavlick2019,nie2020}.

\(D_t\) is bounded in \([0, R/(2(R-1))]\), where \(R = \min_u m_u\). \(D_t\) is directly comparable across unit-types because no marginal-derived chance term is involved.

\subsection{Computation}

\begin{enumerate}
\item Filter to units with \(m_u \geq 2\) raters.
\item For each unit, compute \(k_u\) and \(m_u\), then \(D_u = k_u(m_u-k_u)/\binom{m_u}{2}\).
\item For each unit-type \(t\), compute \(D_t\) as the mean of \(D_u\) over unit-type-\(t\) units.
\item Compute global \(\alpha\) on all qualifying units pooled.
\item Compute \(\alpha_t\) per unit-type (\Cref{subsec:per-tier-alpha}) as the negative control.
\item Bootstrap 95\% CIs on \(D_t\) by resampling units within unit-type (\(B=1000\);~\citealp{efron1993}).
\end{enumerate}

\section{R-Score Worked Example and Unit Construction}
\label{sec:rscore-worked-example}
\label{sec:worked-example}

We illustrate an example and the calculation of metrics for a 10-unit window, drawn from a \benchmark-style document fragment, as shown in~\Cref{fig:labeling}.

\begin{quote}
\ttfamily\small
\setlength{\tabcolsep}{2pt}
\tier{\colorbox{gaptier}{\strut Vehicle:~}}{\(g_a\,g_b\)}\,\tier{\colorbox{yelltier}{\strut"}}{\(y\)}\,\tier{\colorbox{redtier}{\strut 5N1AT2MK4FC824170}}{\(r\)}\,\tier{\colorbox{yelltier}{\strut"}}{\(y\)}\,\tier{\colorbox{gaptier}{\strut~}}{\(g_a\)}\,\tier{\colorbox{yelltier}{\strut"2015 Nissan Rogue"}}{\(y\)}\,\tier{\colorbox{gaptier}{\strut~plate=}}{\(g_a\,g_b\)}\,\tier{\colorbox{redtier}{\strut 321ABC}}{\(r\)}
\end{quote}

Color encodes target label unit-type: \colorbox{redtier}{Red} (mandatory), \colorbox{yelltier}{Yellow} (contextual), \colorbox{gaptier}{Gap} (benign). \noindent After unit construction (combinators between same unit-type reds absorbed into \(r\); gaps longer than three characters split into two units), this yields the 10-unit target sequence \(g_a\,g_b\,y\,r\,y\,g_a\,y\,g_a\,g_b\,r\); the unit-by-unit mapping is given in~\Cref{tab:unit-mapping}. Four labelers (User 1--User 4) annotate each unit, shown in~\Cref{tab:user-labels}.

\begin{table}[h]
\centering
\small
\caption{Source token to unit mapping for the worked example}
\label{tab:unit-mapping}
\begin{tabular}{cll}
\toprule
\textbf{Unit} & \textbf{GT} & \textbf{Source token} \\
\midrule
1 & \cellcolor{gaptier}\(g_a\) & \texttt{Vehicle:\ } (split a) \\
2 & \cellcolor{gaptier}\(g_b\) & \texttt{Vehicle:\ } (split b) \\
3 & \cellcolor{yelltier}\(y\) & \texttt{"} (open quote) \\
4 & \cellcolor{redtier}\(r\) & \texttt{5N1AT2MK4FC824170} (VIN) \\
5 & \cellcolor{yelltier}\(y\) & \texttt{"} (close quote) \\
6 & \cellcolor{gaptier}\(g_a\) & \texttt{\ } (single space) \\
7 & \cellcolor{yelltier}\(y\) & \texttt{"2015 Nissan Rogue"} (\(y\) merged) \\
8 & \cellcolor{gaptier}\(g_a\) & \texttt{\ plate=} (split a) \\
9 & \cellcolor{gaptier}\(g_b\) & \texttt{\ plate=} (split b) \\
10 & \cellcolor{redtier}\(r\) & \texttt{321ABC} \\
\bottomrule
\end{tabular}
\\[2pt]
{\footnotesize Adjacent whitespace is absorbed into the surrounding \(g\) unit before the \(>\)3 character split rule is applied; \texttt{Vehicle:} (units 1--2) absorbs the trailing space and \texttt{plate=} (units 8--9) absorbs the leading space.}
\end{table}

\begin{table}[h]
\centering
\small
\caption{User labels (1 represents a redacted unit)}
\label{tab:user-labels}
\begin{tabular}{l|llllllllll}
\toprule
\textbf{Unit} & 1 & 2 & 3 & 4 & 5 & 6 & 7 & 8 & 9 & 10 \\
\midrule
\textbf{GT} &
\cellcolor{gaptier}\(g_a\) & \cellcolor{gaptier}\(g_b\) & \cellcolor{yelltier}\(y\) &
\cellcolor{redtier}\(r\) & \cellcolor{yelltier}\(y\) & \cellcolor{gaptier}\(g_a\) &
\cellcolor{yelltier}\(y\) & \cellcolor{gaptier}\(g_a\) & \cellcolor{gaptier}\(g_b\) &
\cellcolor{redtier}\(r\) \\
\midrule
User 1 & 0 & 0 & 1 & 1 & 0 & 0 & 1 & 0 & 1 & 1 \\
User 2 & 0 & 0 & 0 & 1 & 1 & 0 & 0 & 0 & 1 & 1 \\
User 3 & 0 & 0 & 1 & 1 & 0 & 0 & 1 & 0 & 0 & 0 \\
User 4 & 0 & 0 & 0 & 1 & 1 & 0 & 0 & 0 & 0 & 0 \\
\bottomrule
\end{tabular}
\end{table}

With the total number of users \(m_u = 4\) for every unit, \(\binom{m_u}{2} = 6\). Per-unit values for \(D_u\) and Krippendorff's per-unit contribution are shown in~\Cref{tab:du-per-unit} and~\Cref{tab:alpha-per-unit}.

\begin{table}[h]
\centering
\small
\caption{Per-unit pairwise disagreement \(D_u = k_u(m_u{-}k_u)/\binom{m_u}{2}\).}
\label{tab:du-per-unit}
\begin{tabular}{cllcccccc}
\toprule
\textbf{Unit} & \textbf{GT} & \textbf{Labels} & \(k\) & \(m\) & \(k(m{-}k)\) & \(\binom{m}{2}\) & \(D_u\) \\
\midrule
1 & \cellcolor{gaptier}\(g_a\) & [0,0,0,0] & 0 & 4 & 0 & 6 & 0.000 \\
2 & \cellcolor{gaptier}\(g_b\) & [0,0,0,0] & 0 & 4 & 0 & 6 & 0.000 \\
3 & \cellcolor{yelltier}\(y\) & [1,0,1,0] & 2 & 4 & 4 & 6 & 0.667 \\
4 & \cellcolor{redtier}\(r\) & [1,1,1,1] & 4 & 4 & 0 & 6 & 0.000 \\
5 & \cellcolor{yelltier}\(y\) & [0,1,0,1] & 2 & 4 & 4 & 6 & 0.667 \\
6 & \cellcolor{gaptier}\(g_a\) & [0,0,0,0] & 0 & 4 & 0 & 6 & 0.000 \\
7 & \cellcolor{yelltier}\(y\) & [1,0,1,0] & 2 & 4 & 4 & 6 & 0.667 \\
8 & \cellcolor{gaptier}\(g_a\) & [0,0,0,0] & 0 & 4 & 0 & 6 & 0.000 \\
9 & \cellcolor{gaptier}\(g_b\) & [1,1,0,0] & 2 & 4 & 4 & 6 & 0.667 \\
10 & \cellcolor{redtier}\(r\) & [1,1,0,0] & 2 & 4 & 4 & 6 & 0.667 \\
\bottomrule
\end{tabular}
\end{table}

\begin{table}[h]
\centering
\small
\caption{Per-unit Krippendorff disagreement contribution \(2\,n_{u_0}\,n_{u_1}/(m_u-1)\).}
\label{tab:alpha-per-unit}
\begin{tabular}{clccccc}
\toprule
\textbf{Unit} & \textbf{GT} & \textbf{Labels} & \(n_{u_0}\) & \(n_{u_1}\) & Contrib & Decimal \\
\midrule
1 & \cellcolor{gaptier}\(g_a\) & [0,0,0,0] & 4 & 0 & \(0/3\) & 0.000 \\
2 & \cellcolor{gaptier}\(g_b\) & [0,0,0,0] & 4 & 0 & \(0/3\) & 0.000 \\
3 & \cellcolor{yelltier}\(y\) & [1,0,1,0] & 2 & 2 & \(8/3\) & 2.667 \\
4 & \cellcolor{redtier}\(r\) & [1,1,1,1] & 0 & 4 & \(0/3\) & 0.000 \\
5 & \cellcolor{yelltier}\(y\) & [0,1,0,1] & 2 & 2 & \(8/3\) & 2.667 \\
6 & \cellcolor{gaptier}\(g_a\) & [0,0,0,0] & 4 & 0 & \(0/3\) & 0.000 \\
7 & \cellcolor{yelltier}\(y\) & [1,0,1,0] & 2 & 2 & \(8/3\) & 2.667 \\
8 & \cellcolor{gaptier}\(g_a\) & [0,0,0,0] & 4 & 0 & \(0/3\) & 0.000 \\
9 & \cellcolor{gaptier}\(g_b\) & [1,1,0,0] & 2 & 2 & \(8/3\) & 2.667 \\
10 & \cellcolor{redtier}\(r\) & [1,1,0,0] & 2 & 2 & \(8/3\) & 2.667 \\
\bottomrule
\end{tabular}
\end{table}

\paragraph{Per-unit-type \(D_t\).} Averaging \(D_u\) within each unit-type:
\begin{align*}
D_r &= \tfrac{1}{2}(0 + 0.667) = \mathbf{0.333} && \text{(units 4, 10)} \\
D_y &= \tfrac{1}{3}(0.667 + 0.667 + 0.667) = \mathbf{0.667} && \text{(units 3, 5, 7; theoretical max for } m=4\text{)} \\
D_g &= \tfrac{1}{5}(0 + 0 + 0 + 0 + 0.667) = \mathbf{0.133} && \text{(units 1, 2, 6, 8, 9)}
\end{align*}

\paragraph{Global \(\alpha\).}
Pool all 40 ratings (10 units \(\times\) 4 raters).

\emph{Observed disagreement \(D_o\).} Sum the per-unit contributions from~\Cref{tab:alpha-per-unit} across all ten units:
\begin{align*}
\sum_u \frac{2 \, n_{u0} \, n_{u1}}{m_u - 1}
&= 0 + 0 + \tfrac{8}{3} + 0 + \tfrac{8}{3} + 0 + \tfrac{8}{3} + 0 + \tfrac{8}{3} + \tfrac{8}{3}
= \tfrac{40}{3} \approx 13.333.
\end{align*}
The five contributing units (3, 5, 7, 9, 10) each have a 2-2 rater split; the other five (1, 2, 4, 6, 8) have unanimous labels and drop out. Then:
\begin{align*}
D_o = \frac{1}{n} \sum_u \frac{2 \, n_{u0} \, n_{u1}}{m_u - 1} = \frac{40/3}{40} = \tfrac{1}{3} \approx 0.333.
\end{align*}

\emph{Expected disagreement \(D_e\).} Tallied across all 40 ratings in~\Cref{tab:user-labels}: \(n_1 = 14\) ``redact'' votes (\(5{+}4{+}3{+}2\) summed across users 1--4) and \(n_0 = 26\) ``don't redact'' votes. Then:
\begin{align*}
D_e = \frac{2 \, n_0 \, n_1}{n(n-1)} = \frac{2 \cdot 26 \cdot 14}{40 \cdot 39} = \frac{728}{1560} \approx 0.467.
\end{align*}

\emph{Coefficient.}
\begin{align*}
\alpha_{\text{global}} = 1 - \frac{D_o}{D_e} = 1 - \frac{0.333}{0.467} = \mathbf{0.286}.
\end{align*}

\paragraph{Per-unit-type \(\alpha\) (negative control).}
~\Cref{tab:per-tier-alpha} computes \(\alpha_t\) within each unit-type subset. To anchor the reader, take Red (units 4, 10): each unit has \(m_u = 4\) raters, so \(n_r = 8\) ratings. Tallying labels in those rows of~\Cref{tab:user-labels} gives \(n_1(r) = 6\) ``redact'' votes and \(n_0(r) = 2\). Summing per-unit contributions for the same units from~\Cref{tab:alpha-per-unit} gives \(0 + 8/3 = 8/3\), so \(D_o(r) = (8/3)/8 = 0.333\). The chance term is \(D_e(r) = 2 \cdot 2 \cdot 6 / (8 \cdot 7) = 24/56 = 0.429\), and \(\alpha_r = 1 - 0.333/0.429 = 0.222\). The same recipe applies to Yellow and Gap types:

\renewcommand{\arraystretch}{1.4}
\begin{table}[h]
\centering
\small
\caption{Per-unit-type Krippendorff's \(\alpha\) on the worked example.}
\label{tab:per-tier-alpha}
\begin{tabular}{lccccccc}
\toprule
\textbf{Unit-Type} & \textbf{Units} & \(n_t\) & \(n_0(t)\) & \(n_1(t)\) & \(D_o(t)\) & \(D_e(t)\) & \(\alpha_t\) \\
\midrule
\cellcolor{redtier}Red & 4, 10 & 8 & 2 & 6 & \((8/3)/8 = 0.333\) & \(24/56 = 0.429\) & \(\mathbf{0.222}\) \\
\cellcolor{yelltier}Yellow & 3, 5, 7 & 12 & 6 & 6 & \(8/12 = 0.667\) & \(72/132 = 0.545\) & \(\mathbf{-0.222}\) \\
\cellcolor{gaptier}Gap & 1, 2, 6, 8, 9 & 20 & 18 & 2 & \((8/3)/20 = 0.133\) & \(72/380 = 0.189\) & \(\mathbf{0.297}\) \\
\bottomrule
\end{tabular}
\end{table}
\renewcommand{\arraystretch}{1.0}

\paragraph{Reading.}~\Cref{tab:summary} consolidates the per-unit-type metrics. \(D_t\) cleanly separates the unit-types, with Yellow disagreement \(2{-}5\times\) higher than Red or Gap. The variance-concentration claim---annotators converge on Red and Gap, diverge on Yellow---is supported. Per-unit-type \(\alpha\) values are suppressed by the prevalence paradox on Red and Gap (extreme marginals depress \(D_e(t)\)) and by genuinely high disagreement on Yellow (driving \(\alpha_y\) negative).

\begin{table}[h]
\centering
\small
\caption{Summary of disagreement metrics on the worked example. \(\bar{p}_t\) is the within-unit-type mean redaction rate; values near 1 (Red) or 0 (Gap) signal that the prevalence paradox is in play. \(D_y/D_r = 2.00\) and \(D_y/D_g = 5.00\) confirm that disagreement concentrates on Yellow.}
\label{tab:summary}
\begin{tabular}{lcccccl}
\toprule
\textbf{Unit-Type} & \(N_t\) & \(\bar{p}_t\) & \(D_t\) & \(\alpha_t\) & \textbf{Notes} \\
\midrule
\cellcolor{redtier}\textbf{Red} & \cellcolor{redtier}2 & \cellcolor{redtier}0.750 & \cellcolor{redtier}\textbf{0.333} & \cellcolor{redtier}0.222 & \cellcolor{redtier}\textit{paradox: skewed to 1} \\
\cellcolor{yelltier}\textbf{Yellow} & \cellcolor{yelltier}3 & \cellcolor{yelltier}0.500 & \cellcolor{yelltier}\textbf{0.667} & \cellcolor{yelltier}\(-\)0.222 & \cellcolor{yelltier}\textit{at theoretical max} \\
\cellcolor{gaptier}\textbf{Gap} & \cellcolor{gaptier}5 & \cellcolor{gaptier}0.100 & \cellcolor{gaptier}\textbf{0.133} & \cellcolor{gaptier}0.297 & \cellcolor{gaptier}\textit{paradox: skewed to 0} \\
\midrule
\textbf{Global} & 10 & 0.350 & --- & \textbf{0.286} & \textit{global inter-annotator agreement} \\
\bottomrule
\end{tabular}
\end{table}

\clearpage
\begin{algorithm}[!ht]
\caption{Combinator structure construction}
\label{alg:redactionbench_combinator_construction}
\begin{algorithmic}[1]
\Require Document \texttt{text}; disjoint half-open spans partitioned into red \(R\) and yellow \(Y\) (adjacency allowed).
\Ensure Red fusion groups, contextual components, paired-delimiter ranges, effective connector markers.
 
\Statex
\State \textbf{Connector types} (each joins two adjacent labeled spans \(a, b\)):
\State \quad \textsc{Punct}: a single yellow separator char, excluding \texttt{\textbackslash\ / @}, brackets, braces, parentheses, quotes, backticks, and ASCII letters/digits --- e.g.\ the \texttt{.} in \texttt{127.0}, not the \texttt{@} in \texttt{user@g}.
\State \quad \textsc{Slash}: a \texttt{/} between two digit-only spans --- e.g.\ \texttt{03/14}, not \texttt{host/10}.
\State \quad \textsc{Bridge}: a closing delimiter then yellow whitespace --- e.g.\ \texttt{(415) 555}, not unmatched \texttt{(415}.
\State \quad \textsc{Pair}: matched brackets/braces/parentheses/angle brackets/quotes enclosing yellow spans --- e.g.\ \texttt{[ABC]}, not bare \texttt{[ABC}.
\Statex
 
\Function{ConnectorStructure}{\texttt{text}, $R$, $Y$}
    \State Init \texttt{red\_graph}, \texttt{yellow\_graph}: node set \(R \cup Y\), no edges
    \State \texttt{effective\_markers} \(\gets [\,]\); \quad \texttt{pair\_ranges} \(\gets [\,]\)
 
    \ForAll{\textsc{Punct} connectors \(c\) between neighbors \(a, b\)}
        \State Add edge \((a, b)\) to \emph{both} graphs; \quad append \(c\) to \texttt{effective\_markers}
    \EndFor
    \ForAll{\textsc{Slash} connectors \(c\) between digit spans \(a, b\)}
        \State Add edge \((a, b)\) to \texttt{yellow\_graph}; \quad append \(c\) to \texttt{effective\_markers}
    \EndFor
    \ForAll{\textsc{Bridge} connectors between \(a, b\) (delimiter \(d\), whitespace \(w\))}
        \State Add edge \((a, b)\) to \texttt{yellow\_graph}; \quad append \(d, w\) to \texttt{effective\_markers}
    \EndFor
    \ForAll{\textsc{Pair} ranges with delimiters \(o, \ell\) enclosing yellow spans}
        \State Append inclusive range \(o \ldots \ell\) to \texttt{pair\_ranges}; \quad append \(o, \ell\) to \texttt{effective\_markers}
    \EndFor
 
    \State \texttt{red\_fusion\_groups} \(\gets\) red spans grouped by connected component of \texttt{red\_graph}
    \State \texttt{context\_components} \(\gets\) yellow spans grouped by connected component of \texttt{yellow\_graph}
    \State \Return \texttt{red\_fusion\_groups}, \texttt{context\_components}, \texttt{pair\_ranges}, \texttt{effective\_markers}
\EndFunction
\end{algorithmic}
\end{algorithm}

\newpage
\begin{algorithm}[t]
\caption{Prediction-dependent contextual selection and entity fusion}
\label{alg:redactionbench_combinator_selection}
\begin{algorithmic}[1]
\Require Target red spans \texttt{target\_red}, target yellow spans \texttt{target\_yellow}, merged predictions \texttt{pred\_merged}; together one touch-connected target over \(R \cup Y\).
\Ensure Fused red entities, selected contextual spans, fused contextual entities.
 
\Statex
\State \textbf{From Alg.~\ref{alg:redactionbench_combinator_construction}:} \texttt{red\_fusion\_groups}, \texttt{context\_components}, \texttt{pair\_ranges}, \texttt{effective\_markers}.
\State \textbf{Note:} red fusion groups are unconditional; contextual selection is prediction-dependent. Red spans may bridge contextual components, but only yellow spans are returned as context.
\Statex
 
\Function{SelectedContextualSpans}{\texttt{target\_yellow}, \texttt{pred\_merged}}
    \State \texttt{selected} \(\gets\) yellow spans directly intersected by any prediction
    \Repeat
        \State Add to \texttt{selected} any contextual component that \texttt{selected} touches
            \Comment{\texttt{Y}\(_1\) in \texttt{Y}\(_1\)\texttt{-Red-Y}\(_2\) also pulls in \texttt{Y}\(_2\)}
        \State Add to \texttt{selected} any pair range with a delimiter in \texttt{selected}
            \Comment{\texttt{[} in \texttt{[ABC]} pulls in the whole range}
    \Until{\texttt{selected} stops growing}
    \State \Return \texttt{selected}
\EndFunction
 
\Statex
\Function{FusedEntityGroups}{\texttt{target\_red}, \texttt{target\_yellow}}
    \State \texttt{red\_entities} \(\gets\) \texttt{red\_fusion\_groups}
    \State Init contextual graph: one node per span in \texttt{target\_yellow}, no edges
    \State Link all yellow spans within each contextual component
        \Comment{groups e.g.\ \texttt{555-5678}}
    \State Link all yellow spans within each pair range
        \Comment{groups e.g.\ \texttt{[ABC]}}
    \State \texttt{contextual\_entities} \(\gets\) yellow spans grouped by connected component of the contextual graph
    \State Remove singletons whose only span is an effective connector marker
    \State \Return \texttt{red\_entities}, \texttt{contextual\_entities}
\EndFunction
\end{algorithmic}
\end{algorithm}

\clearpage
\section{Model Evaluations}
\label{sec:model_harness}

Frontier models with unavailable open weights are all run using an API platform. For \citet{model_claude_opus_4_6, singh2025openaigpt5}, we run inference from their respective API platforms. For \citet{model_zai_org_glm_5_1, model_qwen_qwen3_5_397b_a17b} we run inference with \citet{togetherai2026}. For all local models, including our largest \citet{b2ner}, we use a single NVIDIA H100 Hopper GPU (80GB VRAM). 

Detailed per-model configurations, prompt templates, tool definitions, label sources, and backend identifiers are provided in the supplemental model configuration section below (\Cref{sec:model-configurations}).

\begin{table}[htbp]
\centering
\caption{Overall comparison of all models, grouped by model family.}
\label{tab:model-performance}
\scriptsize
\setlength{\tabcolsep}{2.5pt}
\renewcommand{\arraystretch}{0.78}
\providecommand{\pMetric}[1]{}
\renewcommand{\pMetric}[1]{{\fontsize{6.2pt}{6.8pt}\selectfont #1}}
\resizebox{\linewidth}{!}{%
\begin{tabular}{@{}l l l r r r r r r r@{}}
    \toprule
    \textbf{Family} & \textbf{Base model} & \textbf{Model [Citation]} & \textbf{Params} & \multicolumn{3}{c}{\textbf{Unstructured}} & \multicolumn{3}{c}{\textbf{Overall}} \\
    \cmidrule(lr){5-7} \cmidrule(lr){8-10}
     &  &  &  & \textbf{\pMetric{$P_{20}$}} & \textbf{\pMetric{$P_{50}$}} & \textbf{Mean} & \textbf{\pMetric{$P_{20}$}} & \textbf{\pMetric{$P_{50}$}} & \textbf{Mean} \\
    \midrule
    \multirow{16}{*}{Gen.} & -- & claude-opus-4-6 \citep{model_claude_opus_4_6} & $\sim$ 5.0T & \pMetric{0.62} & \pMetric{0.89} & 0.81 & \pMetric{0.45} & \pMetric{0.78} & 0.71 \\
     & -- & gpt-5.4 \citep{model_gpt_5_4} & $\sim$ 5.0T & \pMetric{0.67} & \pMetric{0.90} & 0.80 & \pMetric{0.36} & \pMetric{0.73} & 0.66 \\
     & -- & Qwen/Qwen3.5-397B-A17B \citep{model_qwen_qwen3_5_397b_a17b} & 397.0B & \pMetric{0.52} & \pMetric{0.83} & 0.74 & \pMetric{0.25} & \pMetric{0.65} & 0.59 \\
     & -- & zai-org/GLM-5.1 \citep{model_zai_org_glm_5_1} & 754.0B & \pMetric{0.66} & \pMetric{0.92} & 0.78 & \pMetric{0.00} & \pMetric{0.72} & 0.56 \\
     & internlm/internlm2-20b & internlm/internlm2-20b \citep{model_b2ner_internlm2_5} & 19.9B & \pMetric{0.32} & \pMetric{0.65} & 0.60 & \pMetric{0.15} & \pMetric{0.43} & 0.45 \\
     & internlm/internlm2\_5-7b & internlm/internlm2\_5-7b \citep{model_b2ner_internlm2_5_7b} & 7.7B & \pMetric{0.27} & \pMetric{0.61} & 0.54 & \pMetric{0.10} & \pMetric{0.37} & 0.40 \\
     & unsloth/gemma-3-270m-it & jakobhuss/pii-extractor-gemma-3-270m-it \citep{model_jakobhuss_pii_extractor_gemma_3_270m_it} & 268M & \pMetric{0.25} & \pMetric{0.60} & 0.53 & \pMetric{0.04} & \pMetric{0.39} & 0.40 \\
     & Qwen/Qwen3-4B & eternisai/Anonymizer-4B \citep{model_eternisai_anonymizer_4b} & 4.0B & \pMetric{0.05} & \pMetric{0.64} & 0.52 & \pMetric{0.00} & \pMetric{0.25} & 0.36 \\
     & meta-llama/Llama-3.2-3B-Instruct & distil-labs/Distil-PII-Llama-3.2-3B-Instruct \citep{model_distil_labs_distil_pii_llama_3_2_3b_instruct} & 3.2B & \pMetric{0.02} & \pMetric{0.50} & 0.47 & \pMetric{0.00} & \pMetric{0.26} & 0.34 \\
     & Qwen/Qwen2-VL-2B-Instruct & numind/NuExtract-2.0-2B \citep{model_numind_nuextract_2_0_2b} & 2.0B & \pMetric{0.22} & \pMetric{0.44} & 0.46 & \pMetric{0.00} & \pMetric{0.25} & 0.32 \\
     & huggyllama/llama-7b & Universal-NER/UniNER-7B-all \citep{model_universal_ner_uniner_7b_all} & 7.0B & \pMetric{0.19} & \pMetric{0.42} & 0.44 & \pMetric{0.05} & \pMetric{0.25} & 0.32 \\
     & Qwen/Qwen2.5-0.5B & numind/NuExtract-1.5-tiny \citep{model_numind_nuextract_1_5_tiny} & 494M & \pMetric{0.23} & \pMetric{0.48} & 0.46 & \pMetric{0.00} & \pMetric{0.26} & 0.31 \\
     & Qwen/Qwen2.5-VL-3B-Instruct & numind/NuExtract-2.0-4B \citep{model_numind_nuextract_2_0_4b} & 4.0B & \pMetric{0.17} & \pMetric{0.34} & 0.40 & \pMetric{0.00} & \pMetric{0.23} & 0.29 \\
     & meta-llama/Llama-3.2-1B-Instruct & distil-labs/Distil-PII-Llama-3.2-1B-Instruct \citep{model_distil_labs_distil_pii_llama_3_2_1b_instruct} & 1.2B & \pMetric{0.00} & \pMetric{0.12} & 0.28 & \pMetric{0.00} & \pMetric{0.00} & 0.18 \\
     & meta-llama/Llama-3.2-1B-Instruct & OpenPipe/PII-Redact-General \citep{model_openpipe_pii_redact_general} & 1.2B & \pMetric{0.00} & \pMetric{0.12} & 0.17 & \pMetric{0.00} & \pMetric{0.05} & 0.11 \\
     & google/gemma-3-270m & distil-labs/Distil-PII-gemma-3-270m-it \citep{model_distil_labs_distil_pii_gemma_3_270m_it} & 268M & \pMetric{0.00} & \pMetric{0.00} & 0.15 & \pMetric{0.00} & \pMetric{0.00} & 0.09 \\
    \midrule
    \multirow{11}{*}{Token} & custom gpt-oss-like classifier & openai/privacy-filter \citep{model_openai_privacy_filter} & 1.4B/179M A\(^*\) & \pMetric{0.43} & \pMetric{0.75} & 0.68 & \pMetric{0.31} & \pMetric{0.59} & 0.58 \\
     & microsoft/deberta-v3-large & OpenMed/OpenMed-PII-SuperClinical-Large-434M-v1 \citep{model_openmed_openmed_pii_superclinical_large_434m_v1} & 434M & \pMetric{0.35} & \pMetric{0.63} & 0.58 & \pMetric{0.20} & \pMetric{0.45} & 0.46 \\
     & microsoft/deberta-v3-base & hydroxai/pii\_model\_weight \citep{model_hydroxai_pii_masker} & 184M & \pMetric{0.32} & \pMetric{0.50} & 0.51 & \pMetric{0.13} & \pMetric{0.36} & 0.38 \\
     & microsoft/mdeberta-v3-base & iiiorg/piiranha-v1-detect-personal-information \citep{model_iiiorg_piiranha_v1_detect_personal_information} & 278M & \pMetric{0.23} & \pMetric{0.45} & 0.45 & \pMetric{0.07} & \pMetric{0.34} & 0.34 \\
     & answerdotai/ModernBERT-base & ai4privacy/llama-ai4privacy-english-anonymiser-openpii \citep{model_ai4privacy_llama_english_anonymiser_openpii} & 150M & \pMetric{0.19} & \pMetric{0.42} & 0.40 & \pMetric{0.03} & \pMetric{0.23} & 0.27 \\
     & microsoft/deberta-base & h2oai/deberta\_finetuned\_pii \citep{model_h2oai_deberta_finetuned_pii}\(^\dagger\) & 139M & \pMetric{0.15} & \pMetric{0.37} & 0.38 & \pMetric{0.01} & \pMetric{0.17} & 0.25 \\
     & microsoft/deberta-base & lakshyakh93/deberta\_finetuned\_pii \citep{model_lakshyakh93_deberta_finetuned_pii}\(^\dagger\) & 139M & \pMetric{0.15} & \pMetric{0.37} & 0.38 & \pMetric{0.01} & \pMetric{0.17} & 0.25 \\
     & distilbert-base-uncased & Isotonic/distilbert\_finetuned\_ai4privacy\_v2 \citep{model_isotonic_distilbert_ai4privacy_v2} & 66M & \pMetric{0.12} & \pMetric{0.28} & 0.35 & \pMetric{0.00} & \pMetric{0.11} & 0.22 \\
     & answerdotai/ModernBERT-base & ai4privacy/llama-ai4privacy-multilingual-categorical-anonymiser-openpii \citep{model_ai4privacy_llama_ai4privacy_multilingual_categorical_anonymiser_openpii} & 150M & \pMetric{0.14} & \pMetric{0.29} & 0.31 & \pMetric{0.02} & \pMetric{0.18} & 0.21 \\
     & tanaos/tanaos-NER-v1 & tanaos/tanaos-text-anonymizer-v1 \citep{model_tanaos_tanaos_text_anonymizer_v1} & 124M & \pMetric{0.10} & \pMetric{0.33} & 0.33 & \pMetric{0.00} & \pMetric{0.10} & 0.20 \\
     & distilbert-base-uncased & deepaksiloka/PII-Detection-V2.1 \citep{model_deepaksiloka_pii_detection_v21} & 66M & \pMetric{0.10} & \pMetric{0.29} & 0.32 & \pMetric{0.00} & \pMetric{0.09} & 0.20 \\
    \midrule
    \multirow{8}{*}{Span} & microsoft/deberta-v3-large & gretelai/gretel-gliner-bi-large-v1.0 \citep{model_gretel_gliner_bi_large_v1_0} & 569M & \pMetric{0.33} & \pMetric{0.56} & 0.54 & \pMetric{0.29} & \pMetric{0.47} & 0.47 \\
     & microsoft/deberta-v3-large & nvidia/gliner-PII \citep{model_nvidia_gliner_pii} & 445M & \pMetric{0.30} & \pMetric{0.58} & 0.54 & \pMetric{0.17} & \pMetric{0.40} & 0.42 \\
     & urchade/gliner\_multi\_pii-v1 & E3-JSI/gliner-multi-pii-domains-v1 \citep{model_e3_jsi_gliner_multi_pii_domains_v1} & 289M & \pMetric{0.20} & \pMetric{0.50} & 0.48 & \pMetric{0.05} & \pMetric{0.32} & 0.35 \\
     & microsoft/mdeberta-v3-base & urchade/gliner\_multi\_pii-v1 \citep{model_urchade_gliner_multi_pii_v1} & 289M & \pMetric{0.19} & \pMetric{0.44} & 0.43 & \pMetric{0.09} & \pMetric{0.28} & 0.33 \\
     & microsoft/deberta-v3-small & knowledgator/gliner-pii-base-v1.0 \citep{model_knowledgator_gliner_pii_base_v1_0} & 166M & \pMetric{0.13} & \pMetric{0.40} & 0.40 & \pMetric{0.06} & \pMetric{0.25} & 0.30 \\
     & jhu-clsp/mmBERT-small & hivetrace/gliner-guard-uniencoder \citep{model_hivetrace_gliner_guard_uniencoder} & 147M & \pMetric{0.10} & \pMetric{0.26} & 0.28 & \pMetric{0.07} & \pMetric{0.20} & 0.24 \\
     & jhu-clsp/mmBERT-small & hivetrace/gliner-guard-biencoder \citep{model_hivetrace_gliner_guard_biencoder} & 144M & \pMetric{0.08} & \pMetric{0.23} & 0.27 & \pMetric{0.05} & \pMetric{0.18} & 0.22 \\
     & microsoft/mdeberta-v3-base & urchade/gliner\_multi-v2.1 \citep{model_gliner_multi_v21} & 289M & \pMetric{0.11} & \pMetric{0.22} & 0.27 & \pMetric{0.00} & \pMetric{0.17} & 0.21 \\
    \bottomrule
\end{tabular}
}%
\\
\footnotesize\(^*\) The 'A' in this parameter count (1.4B/179M A) indicates that the model has 179M active parameters from a pool of 1.4B.
\\
\footnotesize\(^\dagger\) The h2oai and lakshyakh93 uploads point to byte-identical Hugging Face artifacts for the shared \texttt{pytorch\_model.bin}, configuration, tokenizer, and training arguments. 
\end{table}

\begin{table*}[t]
\centering
\caption{Detailed evaluations of all models within the token-based family.}
\label{tab:evals-tokenbased}
\small
\setlength{\tabcolsep}{3.2pt}
\renewcommand{\arraystretch}{1.05}
\providecommand{\pMetric}[1]{#1}
\renewcommand{\pMetric}[1]{\textcolor{black!65}{#1}}
\resizebox{\linewidth}{!}{%
\begin{tabular}{@{}lccccccccccccc@{}}
    \toprule
    \textbf{Model} & \textbf{Params} & \multicolumn{3}{c}{\textbf{Code}} & \multicolumn{3}{c}{\textbf{Files}} & \multicolumn{3}{c}{\textbf{Logs}} & \multicolumn{3}{c}{\textbf{Term.}} \\
    \cmidrule(lr){3-5} \cmidrule(lr){6-8} \cmidrule(lr){9-11} \cmidrule(lr){12-14}
     &  & \pMetric{$P_{20}$} & \pMetric{$P_{50}$} & Mean & \pMetric{$P_{20}$} & \pMetric{$P_{50}$} & Mean & \pMetric{$P_{20}$} & \pMetric{$P_{50}$} & Mean & \pMetric{$P_{20}$} & \pMetric{$P_{50}$} & Mean \\
    \midrule
    \nolinkurl{openai/privacy-filter} & 1.4B/179M A & \pMetric{\textbf{0.44}} & \pMetric{\textbf{0.60}} & \textbf{0.62} & \pMetric{\textbf{0.29}} & \pMetric{\textbf{0.47}} & \textbf{0.47} & \pMetric{\textbf{0.15}} & \pMetric{\textbf{0.26}} & \textbf{0.23} & \pMetric{\textbf{0.31}} & \pMetric{\textbf{0.42}} & \textbf{0.40} \\
    \nolinkurl{OpenMed/OpenMed-PII-SuperClinical-Large-434M-v1} & 434M & \pMetric{\underline{0.11}} & \pMetric{\underline{0.28}} & \underline{0.26} & \pMetric{\underline{0.17}} & \pMetric{\underline{0.36}} & \underline{0.37} & \pMetric{\underline{0.09}} & \pMetric{0.16} & 0.17 & \pMetric{\underline{0.14}} & \pMetric{\underline{0.28}} & \underline{0.28} \\
    \nolinkurl{hydroxai/pii_model_weight} & 184M & \pMetric{\textit{0.01}} & \pMetric{\textit{0.06}} & \textit{0.12} & \pMetric{\textit{0.07}} & \pMetric{\textit{0.24}} & \textit{0.27} & \pMetric{\textit{0.06}} & \pMetric{\textit{0.17}} & \textit{0.21} & \pMetric{\textit{0.08}} & \pMetric{0.09} & 0.15 \\
    \nolinkurl{iiiorg/piiranha-v1-detect-personal-information} & 278M & \pMetric{0.00} & \pMetric{0.00} & 0.08 & \pMetric{0.06} & \pMetric{0.16} & 0.22 & \pMetric{\textit{0.06}} & \pMetric{\underline{0.21}} & \underline{0.22} & \pMetric{\textit{0.08}} & \pMetric{\textit{0.21}} & \textit{0.23} \\
    \nolinkurl{ai4privacy/llama-ai4privacy-english-anonymiser-openpii} & 150M & \pMetric{0.00} & \pMetric{0.04} & 0.05 & \pMetric{0.02} & \pMetric{0.08} & 0.14 & \pMetric{0.02} & \pMetric{0.09} & 0.10 & \pMetric{0.00} & \pMetric{0.02} & 0.06 \\
    \nolinkurl{ai4privacy/llama-ai4privacy-multilingual-categorical-anonymiser-openpii} & 150M & \pMetric{0.00} & \pMetric{0.03} & 0.04 & \pMetric{0.01} & \pMetric{0.05} & 0.12 & \pMetric{0.01} & \pMetric{0.07} & 0.09 & \pMetric{0.00} & \pMetric{0.01} & 0.05 \\
    \nolinkurl{h2oai/deberta_finetuned_pii} & 139M & \pMetric{0.00} & \pMetric{0.00} & 0.04 & \pMetric{0.03} & \pMetric{0.07} & 0.11 & \pMetric{0.00} & \pMetric{0.01} & 0.01 & \pMetric{0.00} & \pMetric{0.05} & 0.07 \\
    \nolinkurl{lakshyakh93/deberta_finetuned_pii} & 139M & \pMetric{0.00} & \pMetric{0.00} & 0.04 & \pMetric{0.03} & \pMetric{0.07} & 0.11 & \pMetric{0.00} & \pMetric{0.01} & 0.01 & \pMetric{0.00} & \pMetric{0.05} & 0.07 \\
    \nolinkurl{Isotonic/distilbert_finetuned_ai4privacy_v2} & 66M & \pMetric{0.00} & \pMetric{0.00} & 0.02 & \pMetric{0.01} & \pMetric{0.03} & 0.06 & \pMetric{0.00} & \pMetric{0.01} & 0.01 & \pMetric{0.00} & \pMetric{0.02} & 0.03 \\
    \nolinkurl{tanaos/tanaos-text-anonymizer-v1} & 124M & \pMetric{0.00} & \pMetric{0.00} & 0.01 & \pMetric{0.00} & \pMetric{0.02} & 0.06 & \pMetric{0.00} & \pMetric{0.00} & 0.00 & \pMetric{0.00} & \pMetric{0.00} & 0.01 \\
    \nolinkurl{deepaksiloka/PII-Detection-V2.1} & 66M & \pMetric{0.00} & \pMetric{0.00} & 0.06 & \pMetric{0.00} & \pMetric{0.01} & 0.05 & \pMetric{0.00} & \pMetric{0.00} & 0.01 & \pMetric{0.00} & \pMetric{0.00} & 0.00 \\
    \midrule
    \addlinespace[2pt]
    \textbf{Model} & \textbf{Params} & \multicolumn{3}{c}{\textbf{Acad.}} & \multicolumn{3}{c}{\textbf{Email}} & \multicolumn{3}{c}{\textbf{Fin.}} & \multicolumn{3}{c}{\textbf{Gov.}} \\
    \cmidrule(lr){3-5} \cmidrule(lr){6-8} \cmidrule(lr){9-11} \cmidrule(lr){12-14}
     &  & \pMetric{$P_{20}$} & \pMetric{$P_{50}$} & Mean & \pMetric{$P_{20}$} & \pMetric{$P_{50}$} & Mean & \pMetric{$P_{20}$} & \pMetric{$P_{50}$} & Mean & \pMetric{$P_{20}$} & \pMetric{$P_{50}$} & Mean \\
    \midrule
    \nolinkurl{openai/privacy-filter} & 1.4B/179M A & \pMetric{\textbf{0.57}} & \pMetric{\textbf{0.85}} & \textbf{0.78} & \pMetric{\textbf{0.46}} & \pMetric{\textbf{0.70}} & \textbf{0.64} & \pMetric{\underline{0.31}} & \pMetric{\textbf{0.73}} & \textbf{0.60} & \pMetric{\textit{0.36}} & \pMetric{\underline{0.58}} & \underline{0.59} \\
    \nolinkurl{OpenMed/OpenMed-PII-SuperClinical-Large-434M-v1} & 434M & \pMetric{\textit{0.40}} & \pMetric{\underline{0.68}} & \textit{0.65} & \pMetric{0.27} & \pMetric{\underline{0.57}} & \underline{0.53} & \pMetric{\underline{0.31}} & \pMetric{\textit{0.55}} & \underline{0.52} & \pMetric{\textbf{0.42}} & \pMetric{\textbf{0.68}} & \textbf{0.64} \\
    \nolinkurl{hydroxai/pii_model_weight} & 184M & \pMetric{\underline{0.51}} & \pMetric{\underline{0.68}} & \underline{0.70} & \pMetric{\underline{0.39}} & \pMetric{\textit{0.52}} & \textit{0.51} & \pMetric{\textbf{0.32}} & \pMetric{\underline{0.56}} & \textit{0.51} & \pMetric{0.29} & \pMetric{0.33} & 0.38 \\
    \nolinkurl{iiiorg/piiranha-v1-detect-personal-information} & 278M & \pMetric{0.25} & \pMetric{\textit{0.59}} & 0.51 & \pMetric{\textit{0.37}} & \pMetric{0.43} & 0.45 & \pMetric{0.14} & \pMetric{0.34} & 0.35 & \pMetric{0.34} & \pMetric{0.48} & \textit{0.50} \\
    \nolinkurl{ai4privacy/llama-ai4privacy-english-anonymiser-openpii} & 150M & \pMetric{0.27} & \pMetric{0.42} & 0.43 & \pMetric{0.19} & \pMetric{0.48} & 0.39 & \pMetric{0.14} & \pMetric{0.42} & 0.36 & \pMetric{\underline{0.37}} & \pMetric{\textit{0.51}} & 0.49 \\
    \nolinkurl{ai4privacy/llama-ai4privacy-multilingual-categorical-anonymiser-openpii} & 150M & \pMetric{0.20} & \pMetric{0.31} & 0.34 & \pMetric{0.14} & \pMetric{0.27} & 0.26 & \pMetric{0.13} & \pMetric{0.32} & 0.29 & \pMetric{0.18} & \pMetric{0.40} & 0.36 \\
    \nolinkurl{h2oai/deberta_finetuned_pii} & 139M & \pMetric{0.20} & \pMetric{0.41} & 0.43 & \pMetric{0.04} & \pMetric{0.28} & 0.31 & \pMetric{0.22} & \pMetric{0.36} & 0.35 & \pMetric{0.24} & \pMetric{0.38} & 0.43 \\
    \nolinkurl{lakshyakh93/deberta_finetuned_pii} & 139M & \pMetric{0.20} & \pMetric{0.41} & 0.43 & \pMetric{0.04} & \pMetric{0.28} & 0.31 & \pMetric{0.22} & \pMetric{0.36} & 0.35 & \pMetric{0.24} & \pMetric{0.38} & 0.43 \\
    \nolinkurl{Isotonic/distilbert_finetuned_ai4privacy_v2} & 66M & \pMetric{0.19} & \pMetric{0.39} & 0.39 & \pMetric{0.08} & \pMetric{0.19} & 0.25 & \pMetric{\textit{0.24}} & \pMetric{0.30} & 0.35 & \pMetric{0.19} & \pMetric{0.40} & 0.44 \\
    \nolinkurl{tanaos/tanaos-text-anonymizer-v1} & 124M & \pMetric{0.24} & \pMetric{0.38} & 0.40 & \pMetric{0.00} & \pMetric{0.11} & 0.18 & \pMetric{0.20} & \pMetric{0.35} & 0.37 & \pMetric{0.15} & \pMetric{0.40} & 0.42 \\
    \nolinkurl{deepaksiloka/PII-Detection-V2.1} & 66M & \pMetric{0.20} & \pMetric{0.32} & 0.39 & \pMetric{0.01} & \pMetric{0.13} & 0.19 & \pMetric{0.13} & \pMetric{0.18} & 0.25 & \pMetric{0.11} & \pMetric{0.39} & 0.33 \\
    \midrule
    \addlinespace[2pt]
    \textbf{Model} & \textbf{Params} & \multicolumn{3}{c}{\textbf{Legal}} & \multicolumn{3}{c}{\textbf{Med.}} & \multicolumn{3}{c}{\textbf{Ops.}} & \multicolumn{3}{c}{\textbf{All}} \\
    \cmidrule(lr){3-5} \cmidrule(lr){6-8} \cmidrule(lr){9-11} \cmidrule(lr){12-14}
     &  & \pMetric{$P_{20}$} & \pMetric{$P_{50}$} & Mean & \pMetric{$P_{20}$} & \pMetric{$P_{50}$} & Mean & \pMetric{$P_{20}$} & \pMetric{$P_{50}$} & Mean & \pMetric{$P_{20}$} & \pMetric{$P_{50}$} & Mean \\
    \midrule
    \nolinkurl{openai/privacy-filter} & 1.4B/179M A & \pMetric{\textbf{0.67}} & \pMetric{\textbf{0.80}} & \textbf{0.78} & \pMetric{\textbf{0.60}} & \pMetric{\textbf{0.72}} & \textbf{0.74} & \pMetric{\textit{0.31}} & \pMetric{\textbf{0.73}} & \underline{0.64} & \pMetric{\textbf{0.31}} & \pMetric{\textbf{0.59}} & \textbf{0.58} \\
    \nolinkurl{OpenMed/OpenMed-PII-SuperClinical-Large-434M-v1} & 434M & \pMetric{\underline{0.45}} & \pMetric{\underline{0.57}} & \underline{0.54} & \pMetric{\textit{0.25}} & \pMetric{\underline{0.62}} & \textit{0.53} & \pMetric{\textbf{0.58}} & \pMetric{\textbf{0.73}} & \textbf{0.68} & \pMetric{\underline{0.20}} & \pMetric{\underline{0.45}} & \underline{0.46} \\
    \nolinkurl{hydroxai/pii_model_weight} & 184M & \pMetric{\textit{0.40}} & \pMetric{\textit{0.50}} & \textit{0.47} & \pMetric{\underline{0.33}} & \pMetric{0.45} & \underline{0.54} & \pMetric{\underline{0.33}} & \pMetric{\textit{0.50}} & 0.47 & \pMetric{\textit{0.13}} & \pMetric{\textit{0.36}} & \textit{0.38} \\
    \nolinkurl{iiiorg/piiranha-v1-detect-personal-information} & 278M & \pMetric{0.24} & \pMetric{0.45} & \textit{0.47} & \pMetric{0.20} & \pMetric{\textit{0.46}} & 0.43 & \pMetric{0.15} & \pMetric{\underline{0.56}} & 0.46 & \pMetric{0.07} & \pMetric{0.34} & 0.34 \\
    \nolinkurl{ai4privacy/llama-ai4privacy-english-anonymiser-openpii} & 150M & \pMetric{0.10} & \pMetric{0.26} & 0.29 & \pMetric{0.20} & \pMetric{0.39} & 0.39 & \pMetric{0.23} & \pMetric{0.43} & 0.44 & \pMetric{0.03} & \pMetric{0.23} & 0.27 \\
    \nolinkurl{ai4privacy/llama-ai4privacy-multilingual-categorical-anonymiser-openpii} & 150M & \pMetric{0.11} & \pMetric{0.23} & 0.24 & \pMetric{0.14} & \pMetric{0.21} & 0.29 & \pMetric{0.27} & \pMetric{0.39} & 0.38 & \pMetric{0.02} & \pMetric{0.18} & 0.21 \\
    \nolinkurl{h2oai/deberta_finetuned_pii} & 139M & \pMetric{0.18} & \pMetric{0.24} & 0.29 & \pMetric{0.14} & \pMetric{0.37} & 0.37 & \pMetric{0.22} & \pMetric{\textit{0.50}} & \textit{0.51} & \pMetric{0.01} & \pMetric{0.17} & 0.25 \\
    \nolinkurl{lakshyakh93/deberta_finetuned_pii} & 139M & \pMetric{0.18} & \pMetric{0.24} & 0.29 & \pMetric{0.14} & \pMetric{0.37} & 0.37 & \pMetric{0.22} & \pMetric{\textit{0.50}} & \textit{0.51} & \pMetric{0.01} & \pMetric{0.17} & 0.25 \\
    \nolinkurl{Isotonic/distilbert_finetuned_ai4privacy_v2} & 66M & \pMetric{0.12} & \pMetric{0.20} & 0.22 & \pMetric{0.09} & \pMetric{0.39} & 0.34 & \pMetric{0.23} & \pMetric{0.38} & 0.47 & \pMetric{0.00} & \pMetric{0.11} & 0.22 \\
    \nolinkurl{tanaos/tanaos-text-anonymizer-v1} & 124M & \pMetric{0.14} & \pMetric{0.20} & 0.24 & \pMetric{0.13} & \pMetric{0.31} & 0.31 & \pMetric{0.15} & \pMetric{0.47} & 0.41 & \pMetric{0.00} & \pMetric{0.10} & 0.20 \\
    \nolinkurl{deepaksiloka/PII-Detection-V2.1} & 66M & \pMetric{0.12} & \pMetric{0.31} & 0.30 & \pMetric{0.09} & \pMetric{0.43} & 0.39 & \pMetric{0.10} & \pMetric{0.35} & 0.38 & \pMetric{0.00} & \pMetric{0.09} & 0.20 \\
    \bottomrule
\end{tabular}
}%
\end{table*}

\begin{table*}[t]
\centering
\caption{Detailed evaluations of all models within the span-based family.}
\label{tab:evals-spanbased}
\small
\setlength{\tabcolsep}{3.2pt}
\renewcommand{\arraystretch}{1.05}
\providecommand{\pMetric}[1]{#1}
\renewcommand{\pMetric}[1]{\textcolor{black!65}{#1}}
\resizebox{\linewidth}{!}{%
\begin{tabular}{@{}lccccccccccccc@{}}
    \toprule
    \textbf{Model} & \textbf{Params} & \multicolumn{3}{c}{\textbf{Code}} & \multicolumn{3}{c}{\textbf{Files}} & \multicolumn{3}{c}{\textbf{Logs}} & \multicolumn{3}{c}{\textbf{Term.}} \\
    \cmidrule(lr){3-5} \cmidrule(lr){6-8} \cmidrule(lr){9-11} \cmidrule(lr){12-14}
     &  & \pMetric{$P_{20}$} & \pMetric{$P_{50}$} & Mean & \pMetric{$P_{20}$} & \pMetric{$P_{50}$} & Mean & \pMetric{$P_{20}$} & \pMetric{$P_{50}$} & Mean & \pMetric{$P_{20}$} & \pMetric{$P_{50}$} & Mean \\
    \midrule
    \nolinkurl{gretelai/gretel-gliner-bi-large-v1.0} & 569M & \pMetric{\textbf{0.30}} & \pMetric{\textbf{0.55}} & \textbf{0.51} & \pMetric{\underline{0.16}} & \pMetric{\underline{0.31}} & \textbf{0.35} & \pMetric{\textbf{0.25}} & \pMetric{\textbf{0.39}} & \textbf{0.36} & \pMetric{\textbf{0.27}} & \pMetric{\textbf{0.35}} & \textbf{0.36} \\
    \nolinkurl{nvidia/gliner-PII} & 445M & \pMetric{\underline{0.07}} & \pMetric{\underline{0.12}} & \underline{0.19} & \pMetric{\textbf{0.18}} & \pMetric{\textbf{0.35}} & \underline{0.34} & \pMetric{\underline{0.10}} & \pMetric{\underline{0.20}} & \underline{0.22} & \pMetric{\underline{0.13}} & \pMetric{\textit{0.18}} & \underline{0.23} \\
    \nolinkurl{E3-JSI/gliner-multi-pii-domains-v1} & 289M & \pMetric{0.00} & \pMetric{0.00} & \textit{0.07} & \pMetric{0.00} & \pMetric{0.15} & 0.24 & \pMetric{0.06} & \pMetric{\textit{0.18}} & \textit{0.20} & \pMetric{0.00} & \pMetric{0.09} & 0.15 \\
    \nolinkurl{urchade/gliner_multi_pii-v1} & 289M & \pMetric{\textit{0.01}} & \pMetric{\textit{0.02}} & 0.04 & \pMetric{0.06} & \pMetric{\textit{0.25}} & \textit{0.27} & \pMetric{\textit{0.08}} & \pMetric{0.15} & 0.16 & \pMetric{0.06} & \pMetric{\underline{0.19}} & \textit{0.19} \\
    \nolinkurl{knowledgator/gliner-pii-base-v1.0} & 166M & \pMetric{0.00} & \pMetric{0.01} & 0.04 & \pMetric{0.06} & \pMetric{\textit{0.25}} & \textit{0.27} & \pMetric{0.05} & \pMetric{0.10} & 0.14 & \pMetric{0.04} & \pMetric{0.13} & 0.14 \\
    \nolinkurl{hivetrace/gliner-guard-uniencoder} & 147M & \pMetric{\textit{0.01}} & \pMetric{\textit{0.02}} & 0.04 & \pMetric{\textit{0.09}} & \pMetric{0.23} & 0.25 & \pMetric{0.06} & \pMetric{0.10} & 0.13 & \pMetric{\textit{0.11}} & \pMetric{0.16} & 0.16 \\
    \nolinkurl{hivetrace/gliner-guard-biencoder} & 144M & \pMetric{\textit{0.01}} & \pMetric{0.01} & 0.03 & \pMetric{0.08} & \pMetric{0.21} & 0.23 & \pMetric{0.05} & \pMetric{0.08} & 0.11 & \pMetric{0.09} & \pMetric{0.13} & 0.13 \\
    \nolinkurl{urchade/gliner_multi-v2.1} & 289M & \pMetric{0.00} & \pMetric{0.00} & 0.04 & \pMetric{0.00} & \pMetric{0.08} & 0.17 & \pMetric{0.00} & \pMetric{0.17} & 0.16 & \pMetric{0.00} & \pMetric{0.02} & 0.10 \\
    \midrule
    \addlinespace[2pt]
    \textbf{Model} & \textbf{Params} & \multicolumn{3}{c}{\textbf{Acad.}} & \multicolumn{3}{c}{\textbf{Email}} & \multicolumn{3}{c}{\textbf{Fin.}} & \multicolumn{3}{c}{\textbf{Gov.}} \\
    \cmidrule(lr){3-5} \cmidrule(lr){6-8} \cmidrule(lr){9-11} \cmidrule(lr){12-14}
     &  & \pMetric{$P_{20}$} & \pMetric{$P_{50}$} & Mean & \pMetric{$P_{20}$} & \pMetric{$P_{50}$} & Mean & \pMetric{$P_{20}$} & \pMetric{$P_{50}$} & Mean & \pMetric{$P_{20}$} & \pMetric{$P_{50}$} & Mean \\
    \midrule
    \nolinkurl{gretelai/gretel-gliner-bi-large-v1.0} & 569M & \pMetric{\textbf{0.40}} & \pMetric{\underline{0.59}} & \textbf{0.62} & \pMetric{0.29} & \pMetric{0.44} & 0.44 & \pMetric{\textbf{0.36}} & \pMetric{\textbf{0.59}} & \textbf{0.57} & \pMetric{\underline{0.35}} & \pMetric{\textit{0.56}} & \textit{0.55} \\
    \nolinkurl{nvidia/gliner-PII} & 445M & \pMetric{\underline{0.30}} & \pMetric{\textit{0.47}} & \textit{0.50} & \pMetric{\underline{0.32}} & \pMetric{\textbf{0.63}} & \textbf{0.56} & \pMetric{\underline{0.33}} & \pMetric{\underline{0.54}} & \textit{0.49} & \pMetric{\textbf{0.39}} & \pMetric{\textbf{0.65}} & \textbf{0.60} \\
    \nolinkurl{E3-JSI/gliner-multi-pii-domains-v1} & 289M & \pMetric{0.19} & \pMetric{\textbf{0.61}} & \underline{0.55} & \pMetric{\textbf{0.34}} & \pMetric{\textit{0.51}} & \textit{0.52} & \pMetric{0.27} & \pMetric{\textbf{0.59}} & \underline{0.51} & \pMetric{\textit{0.30}} & \pMetric{\underline{0.60}} & \underline{0.56} \\
    \nolinkurl{urchade/gliner_multi_pii-v1} & 289M & \pMetric{\textit{0.20}} & \pMetric{0.42} & 0.43 & \pMetric{0.26} & \pMetric{\underline{0.60}} & \underline{0.53} & \pMetric{\textit{0.29}} & \pMetric{\textit{0.53}} & \textit{0.49} & \pMetric{0.28} & \pMetric{\textit{0.56}} & 0.50 \\
    \nolinkurl{knowledgator/gliner-pii-base-v1.0} & 166M & \pMetric{\textit{0.20}} & \pMetric{0.44} & 0.45 & \pMetric{\textit{0.31}} & \pMetric{0.50} & 0.45 & \pMetric{0.18} & \pMetric{0.44} & 0.43 & \pMetric{0.26} & \pMetric{0.48} & 0.45 \\
    \nolinkurl{hivetrace/gliner-guard-uniencoder} & 147M & \pMetric{0.07} & \pMetric{0.18} & 0.22 & \pMetric{0.27} & \pMetric{0.38} & 0.37 & \pMetric{0.13} & \pMetric{0.29} & 0.26 & \pMetric{0.15} & \pMetric{0.37} & 0.38 \\
    \nolinkurl{hivetrace/gliner-guard-biencoder} & 144M & \pMetric{0.08} & \pMetric{0.19} & 0.24 & \pMetric{0.22} & \pMetric{0.39} & 0.36 & \pMetric{0.11} & \pMetric{0.28} & 0.25 & \pMetric{0.13} & \pMetric{0.35} & 0.35 \\
    \nolinkurl{urchade/gliner_multi-v2.1} & 289M & \pMetric{0.12} & \pMetric{0.25} & 0.29 & \pMetric{0.15} & \pMetric{0.31} & 0.28 & \pMetric{0.05} & \pMetric{0.14} & 0.18 & \pMetric{0.21} & \pMetric{0.32} & 0.32 \\
    \midrule
    \addlinespace[2pt]
    \textbf{Model} & \textbf{Params} & \multicolumn{3}{c}{\textbf{Legal}} & \multicolumn{3}{c}{\textbf{Med.}} & \multicolumn{3}{c}{\textbf{Ops.}} & \multicolumn{3}{c}{\textbf{All}} \\
    \cmidrule(lr){3-5} \cmidrule(lr){6-8} \cmidrule(lr){9-11} \cmidrule(lr){12-14}
     &  & \pMetric{$P_{20}$} & \pMetric{$P_{50}$} & Mean & \pMetric{$P_{20}$} & \pMetric{$P_{50}$} & Mean & \pMetric{$P_{20}$} & \pMetric{$P_{50}$} & Mean & \pMetric{$P_{20}$} & \pMetric{$P_{50}$} & Mean \\
    \midrule
    \nolinkurl{gretelai/gretel-gliner-bi-large-v1.0} & 569M & \pMetric{\textbf{0.33}} & \pMetric{\textbf{0.58}} & \textbf{0.50} & \pMetric{\textbf{0.33}} & \pMetric{\underline{0.50}} & \textbf{0.51} & \pMetric{0.36} & \pMetric{0.53} & \textit{0.58} & \pMetric{\textbf{0.29}} & \pMetric{\textbf{0.47}} & \textbf{0.47} \\
    \nolinkurl{nvidia/gliner-PII} & 445M & \pMetric{\underline{0.28}} & \pMetric{\underline{0.50}} & \underline{0.49} & \pMetric{\underline{0.26}} & \pMetric{\textbf{0.51}} & \underline{0.47} & \pMetric{\textbf{0.55}} & \pMetric{\textbf{0.73}} & \textbf{0.65} & \pMetric{\underline{0.17}} & \pMetric{\underline{0.40}} & \underline{0.42} \\
    \nolinkurl{E3-JSI/gliner-multi-pii-domains-v1} & 289M & \pMetric{0.15} & \pMetric{0.40} & 0.36 & \pMetric{\textit{0.13}} & \pMetric{\textit{0.24}} & 0.24 & \pMetric{\underline{0.42}} & \pMetric{\underline{0.62}} & \underline{0.62} & \pMetric{0.05} & \pMetric{\textit{0.32}} & \textit{0.35} \\
    \nolinkurl{urchade/gliner_multi_pii-v1} & 289M & \pMetric{0.11} & \pMetric{0.30} & 0.31 & \pMetric{0.09} & \pMetric{0.20} & 0.22 & \pMetric{\textit{0.41}} & \pMetric{\textit{0.56}} & 0.57 & \pMetric{\textit{0.09}} & \pMetric{0.28} & 0.33 \\
    \nolinkurl{knowledgator/gliner-pii-base-v1.0} & 166M & \pMetric{\textit{0.21}} & \pMetric{\textit{0.41}} & \textit{0.37} & \pMetric{0.06} & \pMetric{0.11} & 0.13 & \pMetric{0.26} & \pMetric{0.49} & 0.52 & \pMetric{0.06} & \pMetric{0.25} & 0.30 \\
    \nolinkurl{hivetrace/gliner-guard-uniencoder} & 147M & \pMetric{0.09} & \pMetric{0.18} & 0.19 & \pMetric{0.05} & \pMetric{0.16} & 0.20 & \pMetric{0.23} & \pMetric{0.40} & 0.40 & \pMetric{0.07} & \pMetric{0.20} & 0.24 \\
    \nolinkurl{hivetrace/gliner-guard-biencoder} & 144M & \pMetric{0.07} & \pMetric{0.14} & 0.16 & \pMetric{0.05} & \pMetric{0.14} & 0.18 & \pMetric{0.22} & \pMetric{0.39} & 0.39 & \pMetric{0.05} & \pMetric{0.18} & 0.22 \\
    \nolinkurl{urchade/gliner_multi-v2.1} & 289M & \pMetric{0.04} & \pMetric{0.18} & 0.24 & \pMetric{0.10} & \pMetric{0.21} & \textit{0.27} & \pMetric{0.10} & \pMetric{0.26} & 0.32 & \pMetric{0.00} & \pMetric{0.17} & 0.21 \\
    \bottomrule
\end{tabular}
}%
\end{table*}

\begin{table*}[t]
\centering
\caption{Detailed evaluations of all models within the generative family.}
\label{tab:evals-generative}
\small
\setlength{\tabcolsep}{3.2pt}
\renewcommand{\arraystretch}{1.05}
\providecommand{\pMetric}[1]{#1}
\renewcommand{\pMetric}[1]{\textcolor{black!65}{#1}}
\resizebox{\linewidth}{!}{%
\begin{tabular}{@{}lccccccccccccc@{}}
    \toprule
    \textbf{Model} & \textbf{Params} & \multicolumn{3}{c}{\textbf{Code}} & \multicolumn{3}{c}{\textbf{Files}} & \multicolumn{3}{c}{\textbf{Logs}} & \multicolumn{3}{c}{\textbf{Term.}} \\
    \cmidrule(lr){3-5} \cmidrule(lr){6-8} \cmidrule(lr){9-11} \cmidrule(lr){12-14}
     &  & \pMetric{$P_{20}$} & \pMetric{$P_{50}$} & Mean & \pMetric{$P_{20}$} & \pMetric{$P_{50}$} & Mean & \pMetric{$P_{20}$} & \pMetric{$P_{50}$} & Mean & \pMetric{$P_{20}$} & \pMetric{$P_{50}$} & Mean \\
    \midrule
    \nolinkurl{claude-opus-4-6} & 5.0T & \pMetric{\textbf{0.35}} & \pMetric{\textbf{0.58}} & \textbf{0.58} & \pMetric{\textbf{0.47}} & \pMetric{\textbf{0.76}} & \textbf{0.65} & \pMetric{\underline{0.21}} & \pMetric{\textbf{0.37}} & \textbf{0.44} & \pMetric{\textbf{0.32}} & \pMetric{\textbf{0.66}} & \textbf{0.60} \\
    \nolinkurl{gpt-5.4} & 5.0T & \pMetric{\underline{0.27}} & \pMetric{\underline{0.45}} & \underline{0.44} & \pMetric{\underline{0.38}} & \pMetric{\underline{0.54}} & \underline{0.56} & \pMetric{\textbf{0.23}} & \pMetric{\underline{0.36}} & \underline{0.38} & \pMetric{\underline{0.27}} & \pMetric{\underline{0.39}} & \textit{0.42} \\
    \nolinkurl{zai-org/GLM-5.1} & 754.0B & \pMetric{0.00} & \pMetric{0.24} & 0.29 & \pMetric{0.00} & \pMetric{0.00} & 0.30 & \pMetric{0.00} & \pMetric{0.00} & 0.08 & \pMetric{0.00} & \pMetric{\textit{0.35}} & 0.35 \\
    \nolinkurl{Qwen/Qwen3.5-397B-A17B} & 397.0B & \pMetric{\textit{0.14}} & \pMetric{\textit{0.40}} & \textit{0.37} & \pMetric{0.00} & \pMetric{\textit{0.48}} & \textit{0.42} & \pMetric{0.00} & \pMetric{\textit{0.25}} & \textit{0.33} & \pMetric{\textit{0.26}} & \pMetric{0.30} & \underline{0.43} \\
    \nolinkurl{B2NER-InternLM2.5} & 19.9B & \pMetric{0.02} & \pMetric{0.13} & 0.17 & \pMetric{0.08} & \pMetric{0.24} & 0.29 & \pMetric{0.09} & \pMetric{\textit{0.25}} & 0.25 & \pMetric{0.12} & \pMetric{0.21} & 0.21 \\
    \nolinkurl{jakobhuss/pii-extractor-gemma-3-270m-it} & 268M & \pMetric{0.00} & \pMetric{0.14} & 0.25 & \pMetric{0.01} & \pMetric{0.15} & 0.23 & \pMetric{0.03} & \pMetric{0.07} & 0.13 & \pMetric{0.00} & \pMetric{0.29} & 0.27 \\
    \nolinkurl{B2NER-InternLM2.5-7B} & 7.7B & \pMetric{0.03} & \pMetric{0.07} & 0.09 & \pMetric{\textit{0.12}} & \pMetric{0.27} & 0.28 & \pMetric{0.09} & \pMetric{0.19} & 0.21 & \pMetric{0.07} & \pMetric{0.12} & 0.18 \\
    \nolinkurl{distil-labs/Distil-PII-Llama-3.2-3B-Instruct} & 3.2B & \pMetric{0.00} & \pMetric{0.00} & 0.09 & \pMetric{0.00} & \pMetric{0.13} & 0.22 & \pMetric{0.01} & \pMetric{0.14} & 0.12 & \pMetric{0.00} & \pMetric{0.00} & 0.14 \\
    \nolinkurl{numind/NuExtract-1.5-tiny} & 494M & \pMetric{0.00} & \pMetric{0.00} & 0.07 & \pMetric{0.00} & \pMetric{0.07} & 0.15 & \pMetric{0.03} & \pMetric{0.07} & 0.07 & \pMetric{0.00} & \pMetric{0.00} & 0.08 \\
    \nolinkurl{eternisai/Anonymizer-4B} & 4.0B & \pMetric{0.00} & \pMetric{0.06} & 0.15 & \pMetric{0.00} & \pMetric{0.00} & 0.15 & \pMetric{0.01} & \pMetric{0.12} & 0.17 & \pMetric{0.00} & \pMetric{0.00} & 0.13 \\
    \nolinkurl{Universal-NER/UniNER-7B-all} & 7.0B & \pMetric{0.00} & \pMetric{0.02} & 0.03 & \pMetric{0.05} & \pMetric{0.15} & 0.20 & \pMetric{\textit{0.10}} & \pMetric{0.13} & 0.19 & \pMetric{0.02} & \pMetric{0.07} & 0.13 \\
    \nolinkurl{numind/NuExtract-2.0-2B} & 2.0B & \pMetric{0.00} & \pMetric{0.00} & 0.06 & \pMetric{0.00} & \pMetric{0.07} & 0.18 & \pMetric{0.00} & \pMetric{0.04} & 0.07 & \pMetric{0.00} & \pMetric{0.08} & 0.12 \\
    \nolinkurl{numind/NuExtract-2.0-4B} & 4.0B & \pMetric{0.00} & \pMetric{0.00} & 0.10 & \pMetric{0.00} & \pMetric{0.05} & 0.17 & \pMetric{0.05} & \pMetric{0.11} & 0.15 & \pMetric{0.00} & \pMetric{0.15} & 0.18 \\
    \nolinkurl{OpenPipe/PII-Redact-General} & 1.2B & \pMetric{0.00} & \pMetric{0.01} & 0.01 & \pMetric{0.00} & \pMetric{0.00} & 0.04 & \pMetric{0.00} & \pMetric{0.02} & 0.05 & \pMetric{0.00} & \pMetric{0.02} & 0.04 \\
    \nolinkurl{distil-labs/Distil-PII-gemma-3-270m-it} & 268M & \pMetric{0.00} & \pMetric{0.00} & 0.00 & \pMetric{0.00} & \pMetric{0.00} & 0.06 & \pMetric{0.00} & \pMetric{0.00} & 0.00 & \pMetric{0.00} & \pMetric{0.00} & 0.00 \\
    \nolinkurl{distil-labs/Distil-PII-Llama-3.2-1B-Instruct} & 1.2B & \pMetric{0.00} & \pMetric{0.00} & 0.06 & \pMetric{0.00} & \pMetric{0.00} & 0.06 & \pMetric{0.00} & \pMetric{0.00} & 0.00 & \pMetric{0.00} & \pMetric{0.00} & 0.00 \\
    \midrule
    \addlinespace[2pt]
    \textbf{Model} & \textbf{Params} & \multicolumn{3}{c}{\textbf{Acad.}} & \multicolumn{3}{c}{\textbf{Email}} & \multicolumn{3}{c}{\textbf{Fin.}} & \multicolumn{3}{c}{\textbf{Gov.}} \\
    \cmidrule(lr){3-5} \cmidrule(lr){6-8} \cmidrule(lr){9-11} \cmidrule(lr){12-14}
     &  & \pMetric{$P_{20}$} & \pMetric{$P_{50}$} & Mean & \pMetric{$P_{20}$} & \pMetric{$P_{50}$} & Mean & \pMetric{$P_{20}$} & \pMetric{$P_{50}$} & Mean & \pMetric{$P_{20}$} & \pMetric{$P_{50}$} & Mean \\
    \midrule
    \nolinkurl{claude-opus-4-6} & 5.0T & \pMetric{\textit{0.61}} & \pMetric{\underline{0.93}} & \textbf{0.85} & \pMetric{\textbf{0.68}} & \pMetric{\textbf{0.87}} & \textbf{0.77} & \pMetric{\textit{0.71}} & \pMetric{\textit{0.93}} & \underline{0.86} & \pMetric{\textit{0.56}} & \pMetric{0.75} & 0.73 \\
    \nolinkurl{gpt-5.4} & 5.0T & \pMetric{\underline{0.73}} & \pMetric{\underline{0.93}} & \underline{0.80} & \pMetric{\underline{0.64}} & \pMetric{\textit{0.81}} & \underline{0.74} & \pMetric{\underline{0.83}} & \pMetric{\textbf{0.98}} & \textbf{0.88} & \pMetric{\textbf{0.62}} & \pMetric{\textbf{0.89}} & \underline{0.79} \\
    \nolinkurl{zai-org/GLM-5.1} & 754.0B & \pMetric{\textbf{0.80}} & \pMetric{\textbf{0.96}} & \textit{0.77} & \pMetric{0.07} & \pMetric{\underline{0.84}} & 0.63 & \pMetric{\textbf{0.89}} & \pMetric{\underline{0.95}} & \textbf{0.88} & \pMetric{\underline{0.60}} & \pMetric{\underline{0.88}} & \textit{0.76} \\
    \nolinkurl{Qwen/Qwen3.5-397B-A17B} & 397.0B & \pMetric{\underline{0.73}} & \pMetric{\textit{0.91}} & \underline{0.80} & \pMetric{\textit{0.49}} & \pMetric{0.73} & \textit{0.67} & \pMetric{0.57} & \pMetric{0.89} & \textit{0.75} & \pMetric{\textbf{0.62}} & \pMetric{\textit{0.86}} & \textbf{0.80} \\
    \nolinkurl{B2NER-InternLM2.5} & 19.9B & \pMetric{0.55} & \pMetric{0.77} & 0.68 & \pMetric{0.37} & \pMetric{0.49} & 0.53 & \pMetric{0.38} & \pMetric{0.53} & 0.54 & \pMetric{0.32} & \pMetric{0.61} & 0.61 \\
    \nolinkurl{jakobhuss/pii-extractor-gemma-3-270m-it} & 268M & \pMetric{0.32} & \pMetric{0.69} & 0.61 & \pMetric{0.13} & \pMetric{0.45} & 0.41 & \pMetric{0.28} & \pMetric{0.56} & 0.51 & \pMetric{0.22} & \pMetric{0.68} & 0.56 \\
    \nolinkurl{B2NER-InternLM2.5-7B} & 7.7B & \pMetric{0.25} & \pMetric{0.56} & 0.53 & \pMetric{0.11} & \pMetric{0.55} & 0.47 & \pMetric{0.27} & \pMetric{0.69} & 0.59 & \pMetric{0.37} & \pMetric{0.48} & 0.54 \\
    \nolinkurl{distil-labs/Distil-PII-Llama-3.2-3B-Instruct} & 3.2B & \pMetric{0.52} & \pMetric{0.64} & 0.69 & \pMetric{0.00} & \pMetric{0.05} & 0.25 & \pMetric{0.21} & \pMetric{0.31} & 0.39 & \pMetric{0.22} & \pMetric{0.52} & 0.47 \\
    \nolinkurl{numind/NuExtract-1.5-tiny} & 494M & \pMetric{0.33} & \pMetric{0.46} & 0.49 & \pMetric{0.17} & \pMetric{0.47} & 0.40 & \pMetric{0.24} & \pMetric{0.42} & 0.39 & \pMetric{0.11} & \pMetric{0.46} & 0.43 \\
    \nolinkurl{eternisai/Anonymizer-4B} & 4.0B & \pMetric{0.20} & \pMetric{0.41} & 0.46 & \pMetric{0.17} & \pMetric{0.51} & 0.50 & \pMetric{0.00} & \pMetric{0.15} & 0.36 & \pMetric{0.01} & \pMetric{0.67} & 0.53 \\
    \nolinkurl{Universal-NER/UniNER-7B-all} & 7.0B & \pMetric{0.16} & \pMetric{0.55} & 0.48 & \pMetric{0.18} & \pMetric{0.36} & 0.38 & \pMetric{0.18} & \pMetric{0.42} & 0.39 & \pMetric{0.34} & \pMetric{0.61} & 0.57 \\
    \nolinkurl{numind/NuExtract-2.0-2B} & 2.0B & \pMetric{0.26} & \pMetric{0.47} & 0.48 & \pMetric{0.25} & \pMetric{0.43} & 0.43 & \pMetric{0.22} & \pMetric{0.32} & 0.40 & \pMetric{0.18} & \pMetric{0.49} & 0.45 \\
    \nolinkurl{numind/NuExtract-2.0-4B} & 4.0B & \pMetric{0.27} & \pMetric{0.37} & 0.44 & \pMetric{0.22} & \pMetric{0.40} & 0.41 & \pMetric{0.11} & \pMetric{0.32} & 0.35 & \pMetric{0.18} & \pMetric{0.40} & 0.40 \\
    \nolinkurl{OpenPipe/PII-Redact-General} & 1.2B & \pMetric{0.02} & \pMetric{0.14} & 0.14 & \pMetric{0.00} & \pMetric{0.09} & 0.14 & \pMetric{0.09} & \pMetric{0.22} & 0.25 & \pMetric{0.04} & \pMetric{0.11} & 0.16 \\
    \nolinkurl{distil-labs/Distil-PII-gemma-3-270m-it} & 268M & \pMetric{0.00} & \pMetric{0.00} & 0.21 & \pMetric{0.00} & \pMetric{0.00} & 0.07 & \pMetric{0.00} & \pMetric{0.00} & 0.02 & \pMetric{0.00} & \pMetric{0.00} & 0.16 \\
    \nolinkurl{distil-labs/Distil-PII-Llama-3.2-1B-Instruct} & 1.2B & \pMetric{0.03} & \pMetric{0.38} & 0.40 & \pMetric{0.00} & \pMetric{0.05} & 0.25 & \pMetric{0.00} & \pMetric{0.15} & 0.22 & \pMetric{0.00} & \pMetric{0.37} & 0.39 \\
    \midrule
    \addlinespace[2pt]
    \textbf{Model} & \textbf{Params} & \multicolumn{3}{c}{\textbf{Legal}} & \multicolumn{3}{c}{\textbf{Med.}} & \multicolumn{3}{c}{\textbf{Ops.}} & \multicolumn{3}{c}{\textbf{All}} \\
    \cmidrule(lr){3-5} \cmidrule(lr){6-8} \cmidrule(lr){9-11} \cmidrule(lr){12-14}
     &  & \pMetric{$P_{20}$} & \pMetric{$P_{50}$} & Mean & \pMetric{$P_{20}$} & \pMetric{$P_{50}$} & Mean & \pMetric{$P_{20}$} & \pMetric{$P_{50}$} & Mean & \pMetric{$P_{20}$} & \pMetric{$P_{50}$} & Mean \\
    \midrule
    \nolinkurl{claude-opus-4-6} & 5.0T & \pMetric{\textit{0.51}} & \pMetric{0.77} & \textit{0.72} & \pMetric{\underline{0.64}} & \pMetric{\textbf{0.97}} & \textbf{0.84} & \pMetric{\textit{0.83}} & \pMetric{0.88} & \underline{0.90} & \pMetric{\textbf{0.45}} & \pMetric{\textbf{0.78}} & \textbf{0.71} \\
    \nolinkurl{gpt-5.4} & 5.0T & \pMetric{\textbf{0.70}} & \pMetric{\textbf{0.87}} & \textbf{0.79} & \pMetric{\textit{0.35}} & \pMetric{0.74} & \textit{0.66} & \pMetric{\underline{0.87}} & \pMetric{\textbf{1.00}} & \textbf{0.91} & \pMetric{\underline{0.36}} & \pMetric{\underline{0.73}} & \underline{0.66} \\
    \nolinkurl{zai-org/GLM-5.1} & 754.0B & \pMetric{\underline{0.68}} & \pMetric{\underline{0.84}} & \underline{0.76} & \pMetric{\textbf{0.67}} & \pMetric{\underline{0.94}} & \underline{0.80} & \pMetric{\textbf{0.89}} & \pMetric{\underline{0.95}} & \textit{0.87} & \pMetric{0.00} & \pMetric{\textit{0.72}} & 0.56 \\
    \nolinkurl{Qwen/Qwen3.5-397B-A17B} & 397.0B & \pMetric{0.28} & \pMetric{\textit{0.83}} & 0.66 & \pMetric{0.32} & \pMetric{\textit{0.75}} & \textit{0.66} & \pMetric{0.75} & \pMetric{\textit{0.90}} & 0.83 & \pMetric{\textit{0.25}} & \pMetric{0.65} & \textit{0.59} \\
    \nolinkurl{B2NER-InternLM2.5} & 19.9B & \pMetric{0.20} & \pMetric{0.68} & 0.55 & \pMetric{0.25} & \pMetric{0.53} & 0.56 & \pMetric{0.64} & \pMetric{0.71} & 0.73 & \pMetric{0.15} & \pMetric{0.43} & 0.45 \\
    \nolinkurl{jakobhuss/pii-extractor-gemma-3-270m-it} & 268M & \pMetric{0.12} & \pMetric{0.51} & 0.44 & \pMetric{0.19} & \pMetric{0.60} & 0.51 & \pMetric{0.51} & \pMetric{0.74} & 0.70 & \pMetric{0.04} & \pMetric{0.39} & 0.40 \\
    \nolinkurl{B2NER-InternLM2.5-7B} & 7.7B & \pMetric{0.22} & \pMetric{0.64} & 0.51 & \pMetric{0.17} & \pMetric{0.60} & 0.52 & \pMetric{0.54} & \pMetric{0.64} & 0.63 & \pMetric{0.10} & \pMetric{0.37} & 0.40 \\
    \nolinkurl{distil-labs/Distil-PII-Llama-3.2-3B-Instruct} & 3.2B & \pMetric{0.00} & \pMetric{0.49} & 0.47 & \pMetric{0.00} & \pMetric{0.54} & 0.51 & \pMetric{0.20} & \pMetric{0.41} & 0.50 & \pMetric{0.00} & \pMetric{0.26} & 0.34 \\
    \nolinkurl{numind/NuExtract-1.5-tiny} & 494M & \pMetric{0.11} & \pMetric{0.39} & 0.42 & \pMetric{0.33} & \pMetric{0.60} & 0.59 & \pMetric{0.37} & \pMetric{0.50} & 0.51 & \pMetric{0.00} & \pMetric{0.26} & 0.31 \\
    \nolinkurl{eternisai/Anonymizer-4B} & 4.0B & \pMetric{0.22} & \pMetric{0.72} & 0.62 & \pMetric{0.26} & \pMetric{0.65} & 0.58 & \pMetric{0.00} & \pMetric{0.80} & 0.58 & \pMetric{0.00} & \pMetric{0.25} & 0.36 \\
    \nolinkurl{Universal-NER/UniNER-7B-all} & 7.0B & \pMetric{0.20} & \pMetric{0.34} & 0.37 & \pMetric{0.18} & \pMetric{0.38} & 0.34 & \pMetric{0.30} & \pMetric{0.55} & 0.54 & \pMetric{0.05} & \pMetric{0.25} & 0.32 \\
    \nolinkurl{numind/NuExtract-2.0-2B} & 2.0B & \pMetric{0.30} & \pMetric{0.55} & 0.54 & \pMetric{0.33} & \pMetric{0.50} & 0.55 & \pMetric{0.00} & \pMetric{0.30} & 0.33 & \pMetric{0.00} & \pMetric{0.25} & 0.32 \\
    \nolinkurl{numind/NuExtract-2.0-4B} & 4.0B & \pMetric{0.02} & \pMetric{0.23} & 0.31 & \pMetric{0.19} & \pMetric{0.37} & 0.48 & \pMetric{0.00} & \pMetric{0.40} & 0.39 & \pMetric{0.00} & \pMetric{0.23} & 0.29 \\
    \nolinkurl{OpenPipe/PII-Redact-General} & 1.2B & \pMetric{0.00} & \pMetric{0.06} & 0.10 & \pMetric{0.01} & \pMetric{0.09} & 0.11 & \pMetric{0.06} & \pMetric{0.18} & 0.29 & \pMetric{0.00} & \pMetric{0.05} & 0.11 \\
    \nolinkurl{distil-labs/Distil-PII-gemma-3-270m-it} & 268M & \pMetric{0.00} & \pMetric{0.00} & 0.19 & \pMetric{0.00} & \pMetric{0.00} & 0.18 & \pMetric{0.00} & \pMetric{0.00} & 0.21 & \pMetric{0.00} & \pMetric{0.00} & 0.09 \\
    \nolinkurl{distil-labs/Distil-PII-Llama-3.2-1B-Instruct} & 1.2B & \pMetric{0.00} & \pMetric{0.00} & 0.06 & \pMetric{0.00} & \pMetric{0.09} & 0.25 & \pMetric{0.00} & \pMetric{0.53} & 0.42 & \pMetric{0.00} & \pMetric{0.00} & 0.18 \\
    \bottomrule
\end{tabular}
}%
\end{table*}

\clearpage
\section{Supplemental Model Configurations}
\label{sec:model-configurations}

\subsection*{Overview}

This supplement records the per-model inference configurations, prompt templates, tool definitions, label sources, and backend identifiers used for the model evaluations in \benchmark.

{\footnotesize
\setlength{\parskip}{0pt}
\subsection{Frontier and Generative Models}

\paragraph*{\texttt{claude-opus-4-6} \hfill\textnormal{\emph{Frontier LLMs}}}
\nopagebreak
\begin{description}\setlength{\itemsep}{0pt}\setlength{\parsep}{0pt}\setlength{\topsep}{0pt}\setlength{\partopsep}{0pt}
\item[Source:] frontier tool overload (MultiEdit)
\item[Model ref:] \texttt{claude-opus-4-6}
\item[Parameter count:] Unknown
\item[Backend:] \texttt{generative.harnessed.anthropic}
\item[Request mode:] tool-call (forced; OpenAI sends two `role=user` messages, Anthropic concatenates them into one)
\item[Tools:] \texttt{MultiEdit} (required args: file\_path, edits) — Apply multiple string replacements to benchmark\_input.txt. List every substring that should be redacted as a separate edit with old\_string set to the substring (verbatim) and new\_string set to "REDACTED". The harness will perform the replacements after intercepting your edit list.
\item[Inference settings:] batch\_size=1
\item[Notes:] Frontier LLM: forced single tool call to `MultiEdit`. The harness intercepts the call, parses the tool arguments, and recovers redaction spans from the substrings the model emits (the system prompt instructs the model to mark them with the marker characters used by `\_parse\_marked\_entity` in `\_harness.py`). The model is never asked to enumerate PII labels.
\end{description}

\vspace{-0.45\baselineskip}
\noindent\textbf{System prompt:}\nopagebreak
\begin{Verbatim}[fontsize=\tiny,frame=single,framesep=0.3mm,baselinestretch=0.65,breaklines=true,breakanywhere=true,breaksymbolleft={},breaksymbolright={}]
You are reviewing one document for public release. Your job is to identify every substring that should be redacted (names, addresses, phone numbers, emails, IDs, account numbers, dates of birth, medical or financial details, anything that could identify a person or sensitive entity) and emit them through a single `MultiEdit` tool call against benchmark_input.txt.

Each edit in the `edits` array should set:
- `old_string` to a substring that exists verbatim in the document
- `new_string` to `REDACTED`

Rules:
- Use the narrowest substring possible (just the name, just the phone number — not the surrounding sentence).
- One substring per edit. Add as many edits as needed.
- If a substring appears multiple times, list it once. Setting `replace_all: true` is fine but optional.
- Make exactly one `MultiEdit` tool call. Do not call any other tool. Do not emit text before or after the tool call.
\end{Verbatim}

\vspace{-0.45\baselineskip}
\noindent\textbf{User template (\texttt{\{text\}} = the chunk being processed):}\nopagebreak
\begin{Verbatim}[fontsize=\tiny,frame=single,framesep=0.3mm,baselinestretch=0.65,breaklines=true,breakanywhere=true,breaksymbolleft={},breaksymbolright={}]
Path: benchmark_input.txt
Document (full):
{text}

User request:
Redact this document as best as possible for public release. Make exactly one tool call as instructed in the system prompt.
\end{Verbatim}

\vspace{-0.45\baselineskip}
\noindent\textbf{Tool definition (JSON; sent to the model as the only allowed function call):}\nopagebreak
\begin{Verbatim}[fontsize=\tiny,frame=single,framesep=0.3mm,baselinestretch=0.65,breaklines=true,breakanywhere=true,breaksymbolleft={},breaksymbolright={}]
[
  {
    "type": "function",
    "function": {
      "name": "MultiEdit",
      "description": "Apply multiple string replacements to benchmark_input.txt. List every substring that should be redacted as a separate edit with old_string set to the substring (verbatim) and new_string set to \"REDACTED\". The harness will perform the replacements after intercepting your edit list.",
      "parameters": {
        "type": "object",
        "additionalProperties": false,
        "properties": {
          "file_path": {
            "type": "string",
            "description": "The file to edit. Always benchmark_input.txt."
          },
          "edits": {
            "type": "array",
            "description": "One entry per substring to redact. Each entry has old_string (the substring) and new_string (\"REDACTED\").",
            "items": {
              "type": "object",
              "additionalProperties": false,
              "properties": {
                "old_string": {
                  "type": "string"
                },
                "new_string": {
                  "type": "string"
                },
                "replace_all": {
                  "type": "boolean"
                }
              },
              "required": [
                "old_string",
                "new_string"
              ]
            }
          }
        },
        "required": [
          "file_path",
          "edits"
        ]
      }
    }
  }
]
\end{Verbatim}

\clearpage

\paragraph*{\texttt{zai-org/GLM-5.1} \hfill\textnormal{\emph{Frontier LLMs}}}
\nopagebreak
\begin{description}\setlength{\itemsep}{0pt}\setlength{\parsep}{0pt}\setlength{\topsep}{0pt}\setlength{\partopsep}{0pt}
\item[Source:] frontier tool overload (bash)
\item[Model ref:] \texttt{zai-org/GLM-5.1}
\item[Parameter count:] 754,000,000,000
\item[Backend:] \texttt{generative.harnessed.togetherai}
\item[Request mode:] tool-call (forced; OpenAI sends two `role=user` messages, Anthropic concatenates them into one)
\item[Tools:] \texttt{bash} (required args: command) — Run a single shell command that redacts every PII or sensitive substring from benchmark\_input.txt. Use chained `sed -i` substitutions, one per substring to redact. The harness will execute the command after intercepting it.
\item[Inference settings:] batch\_size=1
\item[Notes:] Frontier LLM: forced single tool call to `bash`. The harness intercepts the call, parses the tool arguments, and recovers redaction spans from the substrings the model emits (the system prompt instructs the model to mark them with the marker characters used by `\_parse\_marked\_entity` in `\_harness.py`). The model is never asked to enumerate PII labels.
\end{description}

\vspace{-0.45\baselineskip}
\noindent\textbf{System prompt:}\nopagebreak
\begin{Verbatim}[fontsize=\tiny,frame=single,framesep=0.3mm,baselinestretch=0.65,breaklines=true,breakanywhere=true,breaksymbolleft={},breaksymbolright={}]
You are reviewing one document for public release. Your job is to identify every substring that should be redacted (names, addresses, phone numbers, emails, IDs, account numbers, dates of birth, medical or financial details, anything that could identify a person or sensitive entity) and emit them through a single `bash` tool call.

The command should be a single `sed` invocation that chains every substitution against benchmark_input.txt. Example shape:

```
sed -i 's/<substring1>/REDACTED/g; s/<substring2>/REDACTED/g; s/<substring3>/REDACTED/g' benchmark_input.txt
```

Rules:
- Each `s/<substring>/REDACTED/g` substitution targets one substring that exists verbatim in the document.
- Use the narrowest substring possible (just the name, just the phone number — not the surrounding sentence).
- Chain every substitution into one `sed` call using `;` separators (or use multiple `-e` flags — both are fine).
- If a substring contains a `/`, switch the delimiter for that substitution to `#` or `|`, e.g. `s#https://example.com#REDACTED#g`.
- Make exactly one `bash` tool call. Do not call any other tool. Do not emit text before or after the tool call.
\end{Verbatim}

\vspace{-0.45\baselineskip}
\noindent\textbf{User template (\texttt{\{text\}} = the chunk being processed):}\nopagebreak
\begin{Verbatim}[fontsize=\tiny,frame=single,framesep=0.3mm,baselinestretch=0.65,breaklines=true,breakanywhere=true,breaksymbolleft={},breaksymbolright={}]
Path: benchmark_input.txt
Document (full):
{text}

User request:
Redact this document as best as possible for public release. Make exactly one tool call as instructed in the system prompt.
\end{Verbatim}

\vspace{-0.45\baselineskip}
\noindent\textbf{Tool definition (JSON; sent to the model as the only allowed function call):}\nopagebreak
\begin{Verbatim}[fontsize=\tiny,frame=single,framesep=0.3mm,baselinestretch=0.65,breaklines=true,breakanywhere=true,breaksymbolleft={},breaksymbolright={}]
[
  {
    "type": "function",
    "function": {
      "name": "bash",
      "description": "Run a single shell command that redacts every PII or sensitive substring from benchmark_input.txt. Use chained `sed -i` substitutions, one per substring to redact. The harness will execute the command after intercepting it.",
      "parameters": {
        "type": "object",
        "additionalProperties": false,
        "properties": {
          "command": {
            "type": "string",
            "description": "A single shell command. Prefer `sed -i 's/<old1>/REDACTED/g; s/<old2>/REDACTED/g; ...' benchmark_input.txt`."
          }
        },
        "required": [
          "command"
        ]
      }
    }
  }
]
\end{Verbatim}

\clearpage

\paragraph*{\texttt{gpt-5.4} \hfill\textnormal{\emph{Frontier LLMs}}}
\nopagebreak
\begin{description}\setlength{\itemsep}{0pt}\setlength{\parsep}{0pt}\setlength{\topsep}{0pt}\setlength{\partopsep}{0pt}
\item[Source:] frontier tool overload (predict\_entities)
\item[Model ref:] \texttt{gpt-5.4}
\item[Parameter count:] Unknown
\item[Backend:] \texttt{generative.harnessed.openai}
\item[Request mode:] tool-call (forced; OpenAI sends two `role=user` messages, Anthropic concatenates them into one)
\item[Tools:] \texttt{predict\_entities} (required args: entities) — Identify every substring in the document that should be redacted for public release. For each PII or sensitive item, include a short verbatim snippet from the document that contains it, and wrap the exact substring to redact in ‹ › markers. Return all items in the `entities` list in a single tool call.
\item[Inference settings:] batch\_size=1
\item[Notes:] Frontier LLM: forced single tool call to `predict\_entities`. The harness intercepts the call, parses the tool arguments, and recovers redaction spans from the substrings the model emits (the system prompt instructs the model to mark them with the marker characters used by `\_parse\_marked\_entity` in `\_harness.py`). The model is never asked to enumerate PII labels.
\end{description}

\vspace{-0.45\baselineskip}
\noindent\textbf{System prompt:}\nopagebreak
\begin{Verbatim}[fontsize=\tiny,frame=single,framesep=0.3mm,baselinestretch=0.65,breaklines=true,breakanywhere=true,breaksymbolleft={},breaksymbolright={}]
You are reviewing one document for public release. Your job is to identify every substring that should be redacted (names, addresses, phone numbers, emails, IDs, account numbers, dates of birth, medical or financial details, anything that could identify a person or sensitive entity) and emit them through a single `predict_entities` tool call.

Marking format:
- There is a single label: redact.
- Mark every substring to redact using ‹ › markers.
- Do not use any other marker format.

Entity rules:
- Each entity must match document text verbatim.
- Use narrow entity-level spans when possible (just the name, just the phone number — not the surrounding sentence).
- You may include multiple ‹ › spans in one entity string when they form a single logical redaction target.
- If nothing should be redacted, return an empty list.
- Make exactly one `predict_entities` tool call. Do not call any other tool. Do not emit text before or after the tool call.
\end{Verbatim}

\vspace{-0.45\baselineskip}
\noindent\textbf{User template (\texttt{\{text\}} = the chunk being processed):}\nopagebreak
\begin{Verbatim}[fontsize=\tiny,frame=single,framesep=0.3mm,baselinestretch=0.65,breaklines=true,breakanywhere=true,breaksymbolleft={},breaksymbolright={}]
Path: benchmark_input.txt
Document (full):
{text}

User request:
Redact this document as best as possible for public release. Make exactly one tool call as instructed in the system prompt.
\end{Verbatim}

\vspace{-0.45\baselineskip}
\noindent\textbf{Tool definition (JSON; sent to the model as the only allowed function call):}\nopagebreak
\begin{Verbatim}[fontsize=\tiny,frame=single,framesep=0.3mm,baselinestretch=0.65,breaklines=true,breakanywhere=true,breaksymbolleft={},breaksymbolright={}]
[
  {
    "type": "function",
    "function": {
      "name": "predict_entities",
      "description": "Identify every substring in the document that should be redacted for public release. For each PII or sensitive item, include a short verbatim snippet from the document that contains it, and wrap the exact substring to redact in ‹ › markers. Return all items in the `entities` list in a single tool call.",
      "parameters": {
        "type": "object",
        "additionalProperties": false,
        "properties": {
          "entities": {
            "type": "array",
            "description": "One entry per redaction target. Each entry is a verbatim snippet from the document with the sensitive substring wrapped in ‹ › markers, e.g. \"Contact ‹John Smith› at\" or simply \"‹john@example.com›\". A single entry may contain multiple ‹ › spans when they form one logical entity.",
            "items": {
              "type": "string"
            }
          }
        },
        "required": [
          "entities"
        ]
      }
    }
  }
]
\end{Verbatim}

\clearpage

\paragraph*{\texttt{Qwen/Qwen3.5-397B-A17B} \hfill\textnormal{\emph{Frontier LLMs}}}
\nopagebreak
\begin{description}\setlength{\itemsep}{0pt}\setlength{\parsep}{0pt}\setlength{\topsep}{0pt}\setlength{\partopsep}{0pt}
\item[Source:] frontier tool overload (bash)
\item[Model ref:] \texttt{Qwen/Qwen3.5-397B-A17B}
\item[Parameter count:] 397,000,000,000
\item[Backend:] \texttt{generative.harnessed.togetherai}
\item[Request mode:] tool-call (forced; OpenAI sends two `role=user` messages, Anthropic concatenates them into one)
\item[Tools:] \texttt{bash} (required args: command) — Run a single shell command that redacts every PII or sensitive substring from benchmark\_input.txt. Use chained `sed -i` substitutions, one per substring to redact. The harness will execute the command after intercepting it.
\item[Inference settings:] batch\_size=1
\item[Notes:] Frontier LLM: forced single tool call to `bash`. The harness intercepts the call, parses the tool arguments, and recovers redaction spans from the substrings the model emits (the system prompt instructs the model to mark them with the marker characters used by `\_parse\_marked\_entity` in `\_harness.py`). The model is never asked to enumerate PII labels.
\end{description}

\vspace{-0.45\baselineskip}
\noindent\textbf{System prompt:}\nopagebreak
\begin{Verbatim}[fontsize=\tiny,frame=single,framesep=0.3mm,baselinestretch=0.65,breaklines=true,breakanywhere=true,breaksymbolleft={},breaksymbolright={}]
You are reviewing one document for public release. Your job is to identify every substring that should be redacted (names, addresses, phone numbers, emails, IDs, account numbers, dates of birth, medical or financial details, anything that could identify a person or sensitive entity) and emit them through a single `bash` tool call.

The command should be a single `sed` invocation that chains every substitution against benchmark_input.txt. Example shape:

```
sed -i 's/<substring1>/REDACTED/g; s/<substring2>/REDACTED/g; s/<substring3>/REDACTED/g' benchmark_input.txt
```

Rules:
- Each `s/<substring>/REDACTED/g` substitution targets one substring that exists verbatim in the document.
- Use the narrowest substring possible (just the name, just the phone number — not the surrounding sentence).
- Chain every substitution into one `sed` call using `;` separators (or use multiple `-e` flags — both are fine).
- If a substring contains a `/`, switch the delimiter for that substitution to `#` or `|`, e.g. `s#https://example.com#REDACTED#g`.
- Make exactly one `bash` tool call. Do not call any other tool. Do not emit text before or after the tool call.
\end{Verbatim}

\vspace{-0.45\baselineskip}
\noindent\textbf{User template (\texttt{\{text\}} = the chunk being processed):}\nopagebreak
\begin{Verbatim}[fontsize=\tiny,frame=single,framesep=0.3mm,baselinestretch=0.65,breaklines=true,breakanywhere=true,breaksymbolleft={},breaksymbolright={}]
Path: benchmark_input.txt
Document (full):
{text}

User request:
Redact this document as best as possible for public release. Make exactly one tool call as instructed in the system prompt.
\end{Verbatim}

\vspace{-0.45\baselineskip}
\noindent\textbf{Tool definition (JSON; sent to the model as the only allowed function call):}\nopagebreak
\begin{Verbatim}[fontsize=\tiny,frame=single,framesep=0.3mm,baselinestretch=0.65,breaklines=true,breakanywhere=true,breaksymbolleft={},breaksymbolright={}]
[
  {
    "type": "function",
    "function": {
      "name": "bash",
      "description": "Run a single shell command that redacts every PII or sensitive substring from benchmark_input.txt. Use chained `sed -i` substitutions, one per substring to redact. The harness will execute the command after intercepting it.",
      "parameters": {
        "type": "object",
        "additionalProperties": false,
        "properties": {
          "command": {
            "type": "string",
            "description": "A single shell command. Prefer `sed -i 's/<old1>/REDACTED/g; s/<old2>/REDACTED/g; ...' benchmark_input.txt`."
          }
        },
        "required": [
          "command"
        ]
      }
    }
  }
]
\end{Verbatim}

\clearpage

\paragraph*{\texttt{B2NER-InternLM2.5-7B} \hfill\textnormal{\emph{B2NER}}}
\nopagebreak
\begin{description}\setlength{\itemsep}{0pt}\setlength{\parsep}{0pt}\setlength{\topsep}{0pt}\setlength{\partopsep}{0pt}
\item[Source:] compact natural-language PII list (10 entries; NER-format prompt with full label list per chunk)
\item[Model ref:] \texttt{internlm/internlm2\_5-7b}
\item[Parameter count:] 7,748,194,304
\item[Backend:] \texttt{generative.b2ner\_internlm2\_5}
\item[Request mode:] b2ner\_instruction (raw causal-LM completion; no chat template)
\item[Labels (10):] \textit{person name, email address, phone number, address, date, credit card number, ssn, ip address, url, username}
\item[LoRA adapter:] \texttt{Umean/B2NER-Internlm2.5-7B-LoRA}
\item[Compatibility stack:] \texttt{transformers==4.42.3 / peft==0.11.1}
\item[Inference settings:] batch\_size=1, chunk\_overlap\_tokens=64, max\_input\_tokens=4096, max\_new\_tokens=512, preferred\_outer\_batch\_size=1, text\_chunk\_tokens=2048
\item[Notes:] We pin transformers/peft to the versions used in the original B2NER release: newer versions emit empty generations under this PEFT-LoRA setup. The base model is loaded from the Model ref above and the LoRA adapter is then attached.
\end{description}

\vspace{-0.45\baselineskip}
\noindent\textbf{Instruction template (\texttt{\{labels\_str\}} = comma-joined label list; \texttt{\{text\}} = the chunk being processed):}\nopagebreak
\begin{Verbatim}[fontsize=\tiny,frame=single,framesep=0.3mm,baselinestretch=0.65,breaklines=true,breakanywhere=true,breaksymbolleft={},breaksymbolright={}]
Given the label set of entities, please recognize all the entities in the text. The answer format should be "entity label: entity; entity label: entity".
Label Set: {labels_str}

Text: {text}
Answer:
\end{Verbatim}

\ \\

\paragraph*{\texttt{B2NER-InternLM2.5-20B} \hfill\textnormal{\emph{B2NER}}}
\nopagebreak
\begin{description}\setlength{\itemsep}{0pt}\setlength{\parsep}{0pt}\setlength{\topsep}{0pt}\setlength{\partopsep}{0pt}
\item[Source:] compact natural-language PII list (10 entries; NER-format prompt with full label list per chunk)
\item[Model ref:] \texttt{internlm/internlm2-20b}
\item[Parameter count:] 19,861,522,432
\item[Backend:] \texttt{generative.b2ner\_internlm2\_5}
\item[Request mode:] b2ner\_instruction (raw causal-LM completion; no chat template)
\item[Labels (10):] \textit{person name, email address, phone number, address, date, credit card number, ssn, ip address, url, username}
\item[LoRA adapter:] \texttt{Umean/B2NER-Internlm2-20B-LoRA}
\item[Compatibility stack:] \texttt{transformers==4.42.3 / peft==0.11.1}
\item[Inference settings:] batch\_size=1, chunk\_overlap\_tokens=64, max\_input\_tokens=4096, max\_new\_tokens=512, preferred\_outer\_batch\_size=1, text\_chunk\_tokens=2048
\item[Notes:] We pin transformers/peft to the versions used in the original B2NER release: newer versions emit empty generations under this PEFT-LoRA setup. The base model is loaded from the Model ref above and the LoRA adapter is then attached.
\end{description}

\vspace{-0.45\baselineskip}
\noindent\textbf{Instruction template (\texttt{\{labels\_str\}} = comma-joined label list; \texttt{\{text\}} = the chunk being processed):}\nopagebreak
\begin{Verbatim}[fontsize=\tiny,frame=single,framesep=0.3mm,baselinestretch=0.65,breaklines=true,breakanywhere=true,breaksymbolleft={},breaksymbolright={}]
Given the label set of entities, please recognize all the entities in the text. The answer format should be "entity label: entity; entity label: entity".
Label Set: {labels_str}

Text: {text}
Answer:
\end{Verbatim}

\clearpage

\paragraph*{\texttt{Universal-NER/UniNER-7B-all} \hfill\textnormal{\emph{SLMs / Extractors}}}
\nopagebreak
\begin{description}\setlength{\itemsep}{0pt}\setlength{\parsep}{0pt}\setlength{\topsep}{0pt}\setlength{\partopsep}{0pt}
\item[Source:] DEFAULT\_SHARED\_OPENPII\_LABELS (per-label inference; one prompt per label per chunk)
\item[Model ref:] \texttt{Universal-NER/UniNER-7B-all}
\item[Parameter count:] 7,000,000,000
\item[Backend:] \texttt{generative.universal\_ner\_uniner\_7b\_all}
\item[Request mode:] uniner\_chat (custom; one prompt per label per chunk)
\item[Labels (91):] \textit{medical\_record\_number, date\_of\_birth, ssn, date, first\_name, email, last\_name, customer\_id, employee\_id, name, street\_address, phone\_number, ipv4, credit\_card\_number, license\_plate, address, user\_name, device\_identifier, bank\_routing\_number, date\_time, company\_name, unique\_identifier, biometric\_identifier, account\_number, city, certificate\_license\_number, time, postcode, vehicle\_identifier, coordinate, country, api\_key, ipv6, password, health\_plan\_beneficiary\_number, national\_id, tax\_id, url, state, swift\_bic, cvv, pin, prefix, imei, gender, job\_area, job\_type, job\_title, street, secondary\_address, county, age, user\_agent, account\_name, currency\_symbol, amount, credit\_card\_issuer, sex, ip\_address, ethereum\_address, bitcoin\_address, middle\_name, iban, vehicle\_registration\_number, currency, currency\_name, currency\_code, building\_number, ordinal\_direction, masked\_number, zip\_code, bic, mac\_address, gps\_coordinates, vin, eye\_color, height, occupation, credit\_debit\_card, education\_level, race\_ethnicity, employment\_status, fax\_number, language, political\_view, http\_cookie, religious\_belief, blood\_type, sexuality, username, unique\_id}
\item[Inference settings:] batch\_size=1, chunk\_overlap\_tokens=48, max\_input\_tokens=8192, max\_new\_tokens=128, max\_prompts\_per\_batch=4, preferred\_outer\_batch\_size=1, torch\_dtype=bfloat16
\end{description}

\vspace{-0.45\baselineskip}
\noindent\textbf{User template (\texttt{\{text\}} = the chunk being processed):}\nopagebreak
\begin{Verbatim}[fontsize=\tiny,frame=single,framesep=0.3mm,baselinestretch=0.65,breaklines=true,breakanywhere=true,breaksymbolleft={},breaksymbolright={}]
A virtual assistant answers questions from a user based on the provided text. USER: Text: {text} ASSISTANT: I've read this text.</s>USER: What describes {label} in the text? ASSISTANT:
\end{Verbatim}

\clearpage

\paragraph*{\texttt{distil-labs/Distil-PII-gemma-3-270m-it} \hfill\textnormal{\emph{SLMs / Extractors}}}
\nopagebreak
\begin{description}\setlength{\itemsep}{0pt}\setlength{\parsep}{0pt}\setlength{\topsep}{0pt}\setlength{\partopsep}{0pt}
\item[Source:] open prompt (strategy=distil)
\item[Model ref:] \texttt{distil-labs/Distil-PII-gemma-3-270m-it}
\item[Backend:] \texttt{generative.distil\_labs\_distil\_pii\_gemma\_3\_270m\_it}
\item[Request mode:] chat
\item[Inference settings:] batch\_size=4, chunk\_overlap\_tokens=64, do\_sample=false, max\_input\_tokens=8192, max\_new\_tokens=4096, torch\_dtype=bfloat16
\end{description}

\vspace{-0.45\baselineskip}
\noindent\textbf{System prompt:}\nopagebreak
\begin{Verbatim}[fontsize=\tiny,frame=single,framesep=0.3mm,baselinestretch=0.65,breaklines=true,breakanywhere=true,breaksymbolleft={},breaksymbolright={}]
You are a problem solving model working on task_description XML block:
<task_description>
Produce a redacted version of texts, removing sensitive personal data while preserving operational signals. The model must return a single json blob with:

* **redacted_text** is the input with minimal, in-place replacements of redacted entities.
* **entities** as an array of objects with exactly three fields {value: original_value, replacement_token: replacement, reason: reasoning}.

## What to redact (→ replacement token)

* **PERSON** — customer/patient/person names (first/last/full; identifying initials) → `[PERSON]`
* **EMAIL** — any email, including obfuscated `name(at)domain(dot)com` → `[EMAIL]`
* **PHONE** — any international/national format (separators/emoji bullets allowed) → `[PHONE]`
* **ADDRESS** — street + number; full postal lines; apartment/unit numbers → `[ADDRESS]`
* **SSN** — US Social Security numbers → `[SSN]`
* **ID** — national IDs (PESEL, NIN, Aadhaar, DNI, etc.) when personal → `[ID]`
* **UUID** — person-scoped system identifiers (e.g., MRN/NHS/patient IDs/customer UUIDs) → `[UUID]`
* **CREDIT_CARD** — 13–19 digits (spaces/hyphens allowed) → `[CARD_LAST4:####]` (keep last-4 only)
* **IBAN** — IBAN/bank account numbers → `[IBAN_LAST4:####]` (keep last-4 only)
* **GENDER** — self-identification (male/female/non-binary/etc.) → `[GENDER]`
* **AGE** — stated ages ("I'm 29", "age: 47", "29 y/o") → `[AGE_YEARS:##]`
* **RACE** — race/ethnicity self-identification → `[RACE]`
* **MARITAL_STATUS** — married/single/divorced/widowed/partnered → `[MARITAL_STATUS]`

## Keep (do not redact)

* Card **last-4** when only last-4 is present (e.g., "ending 9021", "•••• 9021").
* Operational IDs: order/ticket/invoice numbers, shipment tracking, device serials, case IDs.
* Non-personal org info: company names, product names, team names.
* Cities/countries alone (redact full street+number, not plain city/country mentions).

## Output schema (exactly these fields)
* **redacted_text** The original text with all the sensitive information replaced with redacted tokens
* **entities** Array with all the replaced elements, each element represented by following fields
  * **replacement_token**: one of `[PERSON] | [EMAIL] | [PHONE] | [ADDRESS] | [SSN] | [ID] | [UUID] | [CREDIT_CARD] | [IBAN] | [GENDER] | [AGE] | [RACE] | [MARITAL_STATUS]`
  * **value**: original text that was redacted
  * **reason**: brief string explaining the rule/rationale

for example
{
  "redacted_text": "Hi, I'm [PERSON] and my email is [EMAIL].",
  "entities": [
    { "type": "PERSON", "value": "John Smith", "reason": "person name"},
    { "type": "EMAIL", "value": "john.smith@example.com", "reason": "email"}
  ]
}
</task_description>
You will be given a single task with context in the context XML block and the task in the question XML block
Solve the task in question block based on the context in context block.
Generate only the answer, do not generate anything else
\end{Verbatim}

\vspace{-0.45\baselineskip}
\noindent\textbf{User template (\texttt{\{text\}} = the chunk being processed):}\nopagebreak
\begin{Verbatim}[fontsize=\tiny,frame=single,framesep=0.3mm,baselinestretch=0.65,breaklines=true,breakanywhere=true,breaksymbolleft={},breaksymbolright={}]
Now for the real task, solve the task in question block based on the context in context block.
Generate only the solution, do not generate anything else
<context>
{text}
</context>
<question>Redact provided text according to the task description and return redacted elements.</question>
\end{Verbatim}

\clearpage

\paragraph*{\texttt{distil-labs/Distil-PII-Llama-3.2-1B-Instruct} \hfill\textnormal{\emph{SLMs / Extractors}}}
\nopagebreak
\begin{description}\setlength{\itemsep}{0pt}\setlength{\parsep}{0pt}\setlength{\topsep}{0pt}\setlength{\partopsep}{0pt}
\item[Source:] open prompt (strategy=distil)
\item[Model ref:] \texttt{distil-labs/Distil-PII-Llama-3.2-1B-Instruct}
\item[Backend:] \texttt{generative.distil\_labs\_distil\_pii\_llama\_3\_2\_1b\_instruct}
\item[Request mode:] chat
\item[Inference settings:] batch\_size=2, chunk\_overlap\_tokens=64, do\_sample=false, max\_input\_tokens=8192, max\_new\_tokens=4096, torch\_dtype=bfloat16, use\_redacted\_text\_alignment=true
\end{description}

\vspace{-0.45\baselineskip}
\noindent\textbf{System prompt:}\nopagebreak
\begin{Verbatim}[fontsize=\tiny,frame=single,framesep=0.3mm,baselinestretch=0.65,breaklines=true,breakanywhere=true,breaksymbolleft={},breaksymbolright={}]
You are a problem solving model working on task_description XML block:
<task_description>
Produce a redacted version of texts, removing sensitive personal data while preserving operational signals. The model must return a single json blob with:

* **redacted_text** is the input with minimal, in-place replacements of redacted entities.
* **entities** as an array of objects with exactly three fields {value: original_value, replacement_token: replacement, reason: reasoning}.

## What to redact (→ replacement token)

* **PERSON** — customer/patient/person names (first/last/full; identifying initials) → `[PERSON]`
* **EMAIL** — any email, including obfuscated `name(at)domain(dot)com` → `[EMAIL]`
* **PHONE** — any international/national format (separators/emoji bullets allowed) → `[PHONE]`
* **ADDRESS** — street + number; full postal lines; apartment/unit numbers → `[ADDRESS]`
* **SSN** — US Social Security numbers → `[SSN]`
* **ID** — national IDs (PESEL, NIN, Aadhaar, DNI, etc.) when personal → `[ID]`
* **UUID** — person-scoped system identifiers (e.g., MRN/NHS/patient IDs/customer UUIDs) → `[UUID]`
* **CREDIT_CARD** — 13–19 digits (spaces/hyphens allowed) → `[CARD_LAST4:####]` (keep last-4 only)
* **IBAN** — IBAN/bank account numbers → `[IBAN_LAST4:####]` (keep last-4 only)
* **GENDER** — self-identification (male/female/non-binary/etc.) → `[GENDER]`
* **AGE** — stated ages ("I'm 29", "age: 47", "29 y/o") → `[AGE_YEARS:##]`
* **RACE** — race/ethnicity self-identification → `[RACE]`
* **MARITAL_STATUS** — married/single/divorced/widowed/partnered → `[MARITAL_STATUS]`

## Keep (do not redact)

* Card **last-4** when only last-4 is present (e.g., "ending 9021", "•••• 9021").
* Operational IDs: order/ticket/invoice numbers, shipment tracking, device serials, case IDs.
* Non-personal org info: company names, product names, team names.
* Cities/countries alone (redact full street+number, not plain city/country mentions).

## Output schema (exactly these fields)
* **redacted_text** The original text with all the sensitive information replaced with redacted tokens
* **entities** Array with all the replaced elements, each element represented by following fields
  * **replacement_token**: one of `[PERSON] | [EMAIL] | [PHONE] | [ADDRESS] | [SSN] | [ID] | [UUID] | [CREDIT_CARD] | [IBAN] | [GENDER] | [AGE] | [RACE] | [MARITAL_STATUS]`
  * **value**: original text that was redacted
  * **reason**: brief string explaining the rule/rationale

for example
{
  "redacted_text": "Hi, I'm [PERSON] and my email is [EMAIL].",
  "entities": [
    { "type": "PERSON", "value": "John Smith", "reason": "person name"},
    { "type": "EMAIL", "value": "john.smith@example.com", "reason": "email"}
  ]
}
</task_description>
You will be given a single task with context in the context XML block and the task in the question XML block
Solve the task in question block based on the context in context block.
Generate only the answer, do not generate anything else
\end{Verbatim}

\vspace{-0.45\baselineskip}
\noindent\textbf{User template (\texttt{\{text\}} = the chunk being processed):}\nopagebreak
\begin{Verbatim}[fontsize=\tiny,frame=single,framesep=0.3mm,baselinestretch=0.65,breaklines=true,breakanywhere=true,breaksymbolleft={},breaksymbolright={}]
Now for the real task, solve the task in question block based on the context in context block.
Generate only the solution, do not generate anything else
<context>
{text}
</context>
<question>Redact provided text according to the task description and return redacted elements.</question>
\end{Verbatim}

\clearpage

\paragraph*{\texttt{distil-labs/Distil-PII-Llama-3.2-3B-Instruct} \hfill\textnormal{\emph{SLMs / Extractors}}}
\nopagebreak
\begin{description}\setlength{\itemsep}{0pt}\setlength{\parsep}{0pt}\setlength{\topsep}{0pt}\setlength{\partopsep}{0pt}
\item[Source:] open prompt (strategy=distil)
\item[Model ref:] \texttt{distil-labs/Distil-PII-Llama-3.2-3B-Instruct}
\item[Backend:] \texttt{generative.distil\_labs\_distil\_pii\_llama\_3\_2\_3b\_instruct}
\item[Request mode:] chat
\item[Inference settings:] batch\_size=1, chunk\_overlap\_tokens=64, do\_sample=false, max\_input\_tokens=8192, max\_new\_tokens=4096, torch\_dtype=bfloat16
\end{description}

\vspace{-0.45\baselineskip}
\noindent\textbf{System prompt:}\nopagebreak
\begin{Verbatim}[fontsize=\tiny,frame=single,framesep=0.3mm,baselinestretch=0.65,breaklines=true,breakanywhere=true,breaksymbolleft={},breaksymbolright={}]
You are a problem solving model working on task_description XML block:
<task_description>
Produce a redacted version of texts, removing sensitive personal data while preserving operational signals. The model must return a single json blob with:

* **redacted_text** is the input with minimal, in-place replacements of redacted entities.
* **entities** as an array of objects with exactly three fields {value: original_value, replacement_token: replacement, reason: reasoning}.

## What to redact (→ replacement token)

* **PERSON** — customer/patient/person names (first/last/full; identifying initials) → `[PERSON]`
* **EMAIL** — any email, including obfuscated `name(at)domain(dot)com` → `[EMAIL]`
* **PHONE** — any international/national format (separators/emoji bullets allowed) → `[PHONE]`
* **ADDRESS** — street + number; full postal lines; apartment/unit numbers → `[ADDRESS]`
* **SSN** — US Social Security numbers → `[SSN]`
* **ID** — national IDs (PESEL, NIN, Aadhaar, DNI, etc.) when personal → `[ID]`
* **UUID** — person-scoped system identifiers (e.g., MRN/NHS/patient IDs/customer UUIDs) → `[UUID]`
* **CREDIT_CARD** — 13–19 digits (spaces/hyphens allowed) → `[CARD_LAST4:####]` (keep last-4 only)
* **IBAN** — IBAN/bank account numbers → `[IBAN_LAST4:####]` (keep last-4 only)
* **GENDER** — self-identification (male/female/non-binary/etc.) → `[GENDER]`
* **AGE** — stated ages ("I'm 29", "age: 47", "29 y/o") → `[AGE_YEARS:##]`
* **RACE** — race/ethnicity self-identification → `[RACE]`
* **MARITAL_STATUS** — married/single/divorced/widowed/partnered → `[MARITAL_STATUS]`

## Keep (do not redact)

* Card **last-4** when only last-4 is present (e.g., "ending 9021", "•••• 9021").
* Operational IDs: order/ticket/invoice numbers, shipment tracking, device serials, case IDs.
* Non-personal org info: company names, product names, team names.
* Cities/countries alone (redact full street+number, not plain city/country mentions).

## Output schema (exactly these fields)
* **redacted_text** The original text with all the sensitive information replaced with redacted tokens
* **entities** Array with all the replaced elements, each element represented by following fields
  * **replacement_token**: one of `[PERSON] | [EMAIL] | [PHONE] | [ADDRESS] | [SSN] | [ID] | [UUID] | [CREDIT_CARD] | [IBAN] | [GENDER] | [AGE] | [RACE] | [MARITAL_STATUS]`
  * **value**: original text that was redacted
  * **reason**: brief string explaining the rule/rationale

for example
{
  "redacted_text": "Hi, I'm [PERSON] and my email is [EMAIL].",
  "entities": [
    { "type": "PERSON", "value": "John Smith", "reason": "person name"},
    { "type": "EMAIL", "value": "john.smith@example.com", "reason": "email"}
  ]
}
</task_description>
You will be given a single task with context in the context XML block and the task in the question XML block
Solve the task in question block based on the context in context block.
Generate only the answer, do not generate anything else
\end{Verbatim}

\vspace{-0.45\baselineskip}
\noindent\textbf{User template (\texttt{\{text\}} = the chunk being processed):}\nopagebreak
\begin{Verbatim}[fontsize=\tiny,frame=single,framesep=0.3mm,baselinestretch=0.65,breaklines=true,breakanywhere=true,breaksymbolleft={},breaksymbolright={}]
Now for the real task, solve the task in question block based on the context in context block.
Generate only the solution, do not generate anything else
<context>
{text}
</context>
<question>Redact provided text according to the task description and return redacted elements.</question>
\end{Verbatim}

\clearpage

\paragraph*{\texttt{numind/NuExtract-1.5-tiny} \hfill\textnormal{\emph{SLMs / Extractors}}}
\nopagebreak
\begin{description}\setlength{\itemsep}{0pt}\setlength{\parsep}{0pt}\setlength{\topsep}{0pt}\setlength{\partopsep}{0pt}
\item[Source:] NuExtract JSON schema template — documented list-of-strings (17 fields, model-side schema-following)
\item[Model ref:] \texttt{numind/NuExtract-1.5-tiny}
\item[Parameter count:] 494,032,768
\item[Backend:] \texttt{generative.nuextract\_1\_5\_tiny}
\item[Request mode:] nuextract <|input|>/<|output|> raw completion (see prompt skeleton below)
\item[Labels (17):] \textit{name, email, phone\_number, address, ssn, date\_of\_birth, credit\_card\_number, iban, bank\_account\_number, ip\_address, medical\_record\_number, license\_number, account\_number, url, username, password, api\_key}
\item[Inference settings:] batch\_size=1, chunk\_overlap\_tokens=64, do\_sample=false, max\_input\_tokens=8192, max\_new\_tokens=1024, text\_chunk\_tokens=2000
\end{description}

\vspace{-0.45\baselineskip}
\noindent\textbf{Prompt skeleton (\texttt{\{template\_json\}} = the JSON schema below; \texttt{\{text\}} = the chunk being processed):}\nopagebreak
\begin{Verbatim}[fontsize=\tiny,frame=single,framesep=0.3mm,baselinestretch=0.65,breaklines=true,breakanywhere=true,breaksymbolleft={},breaksymbolright={}]
<|input|>
### Template:
{template_json}
### Text:
{text}

<|output|>
\end{Verbatim}

\vspace{-0.45\baselineskip}
\noindent\textbf{NuExtract JSON schema (\texttt{template\_json}):}\nopagebreak
\begin{Verbatim}[fontsize=\tiny,frame=single,framesep=0.3mm,baselinestretch=0.65,breaklines=true,breakanywhere=true,breaksymbolleft={},breaksymbolright={}]
{
  "account_number": [
    ""
  ],
  "address": [
    ""
  ],
  "api_key": [
    ""
  ],
  "bank_account_number": [
    ""
  ],
  "credit_card_number": [
    ""
  ],
  "date_of_birth": [
    ""
  ],
  "email": [
    ""
  ],
  "iban": [
    ""
  ],
  "ip_address": [
    ""
  ],
  "license_number": [
    ""
  ],
  "medical_record_number": [
    ""
  ],
  "name": [
    ""
  ],
  "password": [
    ""
  ],
  "phone_number": [
    ""
  ],
  "ssn": [
    ""
  ],
  "url": [
    ""
  ],
  "username": [
    ""
  ]
}
\end{Verbatim}

\clearpage

\paragraph*{\texttt{numind/NuExtract-2.0-2B} \hfill\textnormal{\emph{SLMs / Extractors}}}
\nopagebreak
\begin{description}\setlength{\itemsep}{0pt}\setlength{\parsep}{0pt}\setlength{\topsep}{0pt}\setlength{\partopsep}{0pt}
\item[Source:] NuExtract JSON schema template — typed verbatim-string (17 fields, model-side schema-following)
\item[Model ref:] \texttt{numind/NuExtract-2.0-2B}
\item[Parameter count:] 2,000,000,000
\item[Backend:] \texttt{generative.nuextract\_2\_0\_2b}
\item[Request mode:] chat with `template` chat-template kwarg (NuExtract-2.0 chat template ingests the JSON schema)
\item[Labels (17):] \textit{account\_number, address, api\_key, bank\_account\_number, credit\_card\_number, date\_of\_birth, email, iban, ip\_address, license\_number, medical\_record\_number, name, password, phone\_number, ssn, url, username}
\item[Inference settings:] batch\_size=1, chunk\_overlap\_tokens=64, do\_sample=false, max\_input\_tokens=8192, max\_new\_tokens=4096, text\_chunk\_tokens=2000, torch\_dtype=bfloat16
\end{description}

\vspace{-0.45\baselineskip}
\noindent\textbf{Prompt skeleton (\texttt{\{template\_json\}} = the JSON schema below; \texttt{\{text\}} = the chunk being processed):}\nopagebreak
\begin{Verbatim}[fontsize=\tiny,frame=single,framesep=0.3mm,baselinestretch=0.65,breaklines=true,breakanywhere=true,breaksymbolleft={},breaksymbolright={}]
{text}
\end{Verbatim}

\vspace{-0.45\baselineskip}
\noindent\textbf{NuExtract JSON schema (\texttt{template\_json}):}\nopagebreak
\begin{Verbatim}[fontsize=\tiny,frame=single,framesep=0.3mm,baselinestretch=0.65,breaklines=true,breakanywhere=true,breaksymbolleft={},breaksymbolright={}]
{
  "account_number": "verbatim-string",
  "address": "verbatim-string",
  "api_key": "verbatim-string",
  "bank_account_number": "verbatim-string",
  "credit_card_number": "verbatim-string",
  "date_of_birth": "verbatim-string",
  "email": "verbatim-string",
  "iban": "verbatim-string",
  "ip_address": "verbatim-string",
  "license_number": "verbatim-string",
  "medical_record_number": "verbatim-string",
  "name": "verbatim-string",
  "password": "verbatim-string",
  "phone_number": "verbatim-string",
  "ssn": "verbatim-string",
  "url": "verbatim-string",
  "username": "verbatim-string"
}
\end{Verbatim}

\paragraph*{\texttt{numind/NuExtract-2.0-4B} \hfill\textnormal{\emph{SLMs / Extractors}}}
\nopagebreak
\begin{description}\setlength{\itemsep}{0pt}\setlength{\parsep}{0pt}\setlength{\topsep}{0pt}\setlength{\partopsep}{0pt}
\item[Source:] NuExtract JSON schema template — documented list-of-strings (17 fields, model-side schema-following)
\item[Model ref:] \texttt{numind/NuExtract-2.0-4B}
\item[Parameter count:] 4,000,000,000
\item[Backend:] \texttt{generative.nuextract\_2\_0\_4b}
\item[Request mode:] chat with `template` chat-template kwarg (NuExtract-2.0 chat template ingests the JSON schema)
\item[Labels (17):] \textit{name, email, phone\_number, address, ssn, date\_of\_birth, credit\_card\_number, iban, bank\_account\_number, ip\_address, medical\_record\_number, license\_number, account\_number, url, username, password, api\_key}
\item[Inference settings:] batch\_size=1, chunk\_overlap\_tokens=64, do\_sample=false, max\_input\_tokens=8192, max\_new\_tokens=4096, text\_chunk\_tokens=2000, torch\_dtype=bfloat16
\end{description}

\vspace{-0.45\baselineskip}
\noindent\textbf{Prompt skeleton (\texttt{\{template\_json\}} = the JSON schema below; \texttt{\{text\}} = the chunk being processed):}\nopagebreak
\begin{Verbatim}[fontsize=\tiny,frame=single,framesep=0.3mm,baselinestretch=0.65,breaklines=true,breakanywhere=true,breaksymbolleft={},breaksymbolright={}]
{text}
\end{Verbatim}

\vspace{-0.45\baselineskip}
\noindent\textbf{NuExtract JSON schema (\texttt{template\_json}):}\nopagebreak
\begin{Verbatim}[fontsize=\tiny,frame=single,framesep=0.3mm,baselinestretch=0.65,breaklines=true,breakanywhere=true,breaksymbolleft={},breaksymbolright={}]
{
  "account_number": [
    ""
  ],
  "address": [
    ""
  ],
  "api_key": [
    ""
  ],
  "bank_account_number": [
    ""
  ],
  "credit_card_number": [
    ""
  ],
  "date_of_birth": [
    ""
  ],
  "email": [
    ""
  ],
  "iban": [
    ""
  ],
  "ip_address": [
    ""
  ],
  "license_number": [
    ""
  ],
  "medical_record_number": [
    ""
  ],
  "name": [
    ""
  ],
  "password": [
    ""
  ],
  "phone_number": [
    ""
  ],
  "ssn": [
    ""
  ],
  "url": [
    ""
  ],
  "username": [
    ""
  ]
}
\end{Verbatim}

\clearpage

\paragraph*{\texttt{OpenPipe/PII-Redact-General} \hfill\textnormal{\emph{SLMs / Extractors}}}
\nopagebreak
\begin{description}\setlength{\itemsep}{0pt}\setlength{\parsep}{0pt}\setlength{\topsep}{0pt}\setlength{\partopsep}{0pt}
\item[Source:] open prompt (strategy=openpipe)
\item[Model ref:] \texttt{OpenPipe/PII-Redact-General}
\item[Parameter count:] 1,235,814,400
\item[Backend:] \texttt{generative.openpipe\_pii\_redact\_general}
\item[Request mode:] chat (with tool-call contract)
\item[Tools:] \texttt{report\_entities} (required args: entities) — Return exact entity substrings to redact.
\item[Inference settings:] batch\_size=4, chunk\_overlap\_tokens=64, do\_sample=false, max\_input\_tokens=8192, max\_new\_tokens=1024, max\_span\_chars=40, text\_chunk\_tokens=1024, torch\_dtype=bfloat16
\end{description}

\vspace{-0.45\baselineskip}
\noindent\textbf{System prompt:}\nopagebreak
\begin{Verbatim}[fontsize=\tiny,frame=single,framesep=0.3mm,baselinestretch=0.65,breaklines=true,breakanywhere=true,breaksymbolleft={},breaksymbolright={}]
Find personally identifying or sensitive entities in the user text and call `report_entities` exactly once. Return exact substrings only, not whole sentences, explanations, punctuation, or common words. Include names, organizations, locations, dates/times, addresses, emails, phone numbers, URLs, account identifiers, credentials, demographic attributes, and person- or organization-linked monetary values.
\end{Verbatim}

\vspace{-0.45\baselineskip}
\noindent\textbf{User template (\texttt{\{text\}} = the chunk being processed):}\nopagebreak
\begin{Verbatim}[fontsize=\tiny,frame=single,framesep=0.3mm,baselinestretch=0.65,breaklines=true,breakanywhere=true,breaksymbolleft={},breaksymbolright={}]
Text to inspect:
{text}
\end{Verbatim}

\vspace{-0.45\baselineskip}
\noindent\textbf{Tool definition (JSON; sent to the model as the only allowed function call):}\nopagebreak
\begin{Verbatim}[fontsize=\tiny,frame=single,framesep=0.3mm,baselinestretch=0.65,breaklines=true,breakanywhere=true,breaksymbolleft={},breaksymbolright={}]
[
  {
    "type": "function",
    "function": {
      "name": "report_entities",
      "description": "Return exact entity substrings to redact.",
      "parameters": {
        "type": "object",
        "additionalProperties": false,
        "properties": {
          "entities": {
            "type": "array",
            "items": {
              "type": "string"
            }
          }
        },
        "required": [
          "entities"
        ]
      }
    }
  }
]
\end{Verbatim}

\ \\

\paragraph*{\texttt{jakobhuss/pii-extractor-gemma-3-270m-it} \hfill\textnormal{\emph{SLMs / Extractors}}}
\nopagebreak
\begin{description}\setlength{\itemsep}{0pt}\setlength{\parsep}{0pt}\setlength{\topsep}{0pt}\setlength{\partopsep}{0pt}
\item[Source:] open prompt (strategy=jakobhuss)
\item[Model ref:] \texttt{jakobhuss/pii-extractor-gemma-3-270m-it}
\item[Backend:] \texttt{generative.jakobhuss\_pii\_extractor\_gemma\_3\_270m\_it}
\item[Request mode:] chat
\item[Inference settings:] batch\_size=4, chunk\_overlap\_tokens=64, do\_sample=false, max\_input\_tokens=8192, max\_new\_tokens=512, torch\_dtype=bfloat16
\end{description}

\vspace{-0.45\baselineskip}
\noindent\textbf{User template (\texttt{\{text\}} = the chunk being processed):}\nopagebreak
\begin{Verbatim}[fontsize=\tiny,frame=single,framesep=0.3mm,baselinestretch=0.65,breaklines=true,breakanywhere=true,breaksymbolleft={},breaksymbolright={}]
List PII or secret substrings that should be redacted. Return either JSON with `entities` or a concise plain list.

{text}
\end{Verbatim}

\clearpage

\paragraph*{\texttt{eternisai/Anonymizer-4B} \hfill\textnormal{\emph{SLMs / Extractors}}}
\nopagebreak
\begin{description}\setlength{\itemsep}{0pt}\setlength{\parsep}{0pt}\setlength{\topsep}{0pt}\setlength{\partopsep}{0pt}
\item[Source:] open prompt (strategy=eternisai)
\item[Model ref:] \texttt{eternisai/Anonymizer-4B}
\item[Backend:] \texttt{generative.eternisai\_anonymizer\_4b}
\item[Request mode:] chat (with tool-call contract)
\item[Tools:] \texttt{replace\_entities} (required args: replacements) — List replacements for PII entities.
\item[Inference settings:] batch\_size=1, chunk\_overlap\_tokens=64, do\_sample=false, max\_input\_tokens=8192, max\_new\_tokens=512, torch\_dtype=bfloat16
\end{description}

\vspace{-0.45\baselineskip}
\noindent\textbf{System prompt:}\nopagebreak
\begin{Verbatim}[fontsize=\tiny,frame=single,framesep=0.3mm,baselinestretch=0.65,breaklines=true,breakanywhere=true,breaksymbolleft={},breaksymbolright={}]
You are an anonymizer. Your task is to identify and replace all personally identifiable information (PII) in the given text.
Replace PII entities with semantically equivalent alternatives that preserve the context needed for a good response.
If no PII is found or replacement is not needed, return an empty replacements list.
\end{Verbatim}

\vspace{-0.45\baselineskip}
\noindent\textbf{User template (\texttt{\{text\}} = the chunk being processed):}\nopagebreak
\begin{Verbatim}[fontsize=\tiny,frame=single,framesep=0.3mm,baselinestretch=0.65,breaklines=true,breakanywhere=true,breaksymbolleft={},breaksymbolright={}]
Text to anonymize:
{text}
/no_think
\end{Verbatim}

\vspace{-0.45\baselineskip}
\noindent\textbf{Tool definition (JSON; sent to the model as the only allowed function call):}\nopagebreak
\begin{Verbatim}[fontsize=\tiny,frame=single,framesep=0.3mm,baselinestretch=0.65,breaklines=true,breakanywhere=true,breaksymbolleft={},breaksymbolright={}]
[
  {
    "type": "function",
    "function": {
      "name": "replace_entities",
      "description": "List replacements for PII entities.",
      "parameters": {
        "type": "object",
        "additionalProperties": false,
        "properties": {
          "replacements": {
            "type": "array",
            "items": {
              "type": "object",
              "additionalProperties": true,
              "properties": {
                "original": {
                  "type": "string"
                },
                "replacement": {
                  "type": "string"
                }
              },
              "required": [
                "original",
                "replacement"
              ]
            }
          }
        },
        "required": [
          "replacements"
        ]
      }
    }
  }
]
\end{Verbatim}

\clearpage
\subsection{Span-Based GLiNER Models}
\ \\

\paragraph*{\texttt{hivetrace/gliner-guard-biencoder} \hfill\textnormal{\emph{GLiNER}}}
\nopagebreak
\begin{description}\setlength{\itemsep}{0pt}\setlength{\parsep}{0pt}\setlength{\topsep}{0pt}\setlength{\partopsep}{0pt}
\item[Source:] per-model PII training labels (label\_profile='model')
\item[Model ref:] \texttt{hivetrace/gliner-guard-biencoder}
\item[Parameter count:] 144,392,726
\item[Backend:] \texttt{spanbased.hivetrace\_gliner\_guard\_biencoder}
\item[Request mode:] GLiNER zero-shot span tagging (model.predict\_entities)
\item[Labels (32):] \textit{person, first\_name, last\_name, alias, title, country, region, city, district, street, building, unit, postal\_code, landmark, address, company, government, education, media, product, email, phone, social\_account, messenger, passport, national\_id, document\_id, date\_of\_birth, event\_date, card\_number, bank\_account, crypto\_wallet}
\item[Inference settings:] batch\_size=16, chunk\_infer\_batch\_size=2, chunk\_tokens=384, preferred\_outer\_batch\_size=4, threshold=0.01, threshold\_source=optimized
\item[Notes:] GLiNER models accept the zero-shot label list at inference; we feed the model its own training labels (the MODEL\_LABELS list exported from each backend module). The decision threshold is optimised on a held-out slice.
\end{description}

\ \\

\paragraph*{\texttt{hivetrace/gliner-guard-uniencoder} \hfill\textnormal{\emph{GLiNER}}}
\nopagebreak
\begin{description}\setlength{\itemsep}{0pt}\setlength{\parsep}{0pt}\setlength{\topsep}{0pt}\setlength{\partopsep}{0pt}
\item[Source:] per-model PII training labels (label\_profile='model')
\item[Model ref:] \texttt{hivetrace/gliner-guard-uniencoder}
\item[Parameter count:] 146,941,205
\item[Backend:] \texttt{spanbased.hivetrace\_gliner\_guard\_uniencoder}
\item[Request mode:] GLiNER zero-shot span tagging (model.predict\_entities)
\item[Labels (32):] \textit{person, first\_name, last\_name, alias, title, country, region, city, district, street, building, unit, postal\_code, landmark, address, company, government, education, media, product, email, phone, social\_account, messenger, passport, national\_id, document\_id, date\_of\_birth, event\_date, card\_number, bank\_account, crypto\_wallet}
\item[Inference settings:] batch\_size=16, chunk\_infer\_batch\_size=2, chunk\_tokens=384, preferred\_outer\_batch\_size=4, threshold=0.01, threshold\_source=optimized
\item[Notes:] GLiNER models accept the zero-shot label list at inference; we feed the model its own training labels (the MODEL\_LABELS list exported from each backend module). The decision threshold is optimised on a held-out slice.
\end{description}

\ \\

\paragraph*{\texttt{nvidia\_gliner\_pii} \hfill\textnormal{\emph{GLiNER}}}
\nopagebreak
\begin{description}\setlength{\itemsep}{0pt}\setlength{\parsep}{0pt}\setlength{\topsep}{0pt}\setlength{\partopsep}{0pt}
\item[Source:] per-model PII training labels (label\_profile='model')
\item[Model ref:] \texttt{nvidia/gliner-PII}
\item[Parameter count:] 445,463,040
\item[Backend:] \texttt{spanbased.nvidia\_gliner\_pii}
\item[Request mode:] GLiNER zero-shot span tagging (model.predict\_entities)
\item[Labels (55):] \textit{account\_number, age, api\_key, bank\_routing\_number, biometric\_identifier, blood\_type, certificate\_license\_number, city, company\_name, coordinate, country, county, credit\_debit\_card, customer\_id, cvv, date, date\_of\_birth, date\_time, device\_identifier, education\_level, email, employee\_id, employment\_status, fax\_number, first\_name, gender, health\_plan\_beneficiary\_number, http\_cookie, ipv4, ipv6, language, last\_name, license\_plate, mac\_address, medical\_record\_number, national\_id, occupation, password, phone\_number, pin, political\_view, postcode, race\_ethnicity, religious\_belief, sexuality, ssn, state, street\_address, swift\_bic, tax\_id, time, unique\_id, url, user\_name, vehicle\_identifier}
\item[Inference settings:] batch\_size=4, chunk\_infer\_batch\_size=2, chunk\_tokens=384, preferred\_outer\_batch\_size=4, threshold=0.6531, threshold\_source=optimized
\item[Notes:] GLiNER models accept the zero-shot label list at inference; we feed the model its own training labels (the MODEL\_LABELS list exported from each backend module). The decision threshold is optimised on a held-out slice.
\end{description}

\clearpage

\paragraph*{\texttt{urchade/gliner\_multi\_pii-v1} \hfill\textnormal{\emph{GLiNER}}}
\nopagebreak
\begin{description}\setlength{\itemsep}{0pt}\setlength{\parsep}{0pt}\setlength{\topsep}{0pt}\setlength{\partopsep}{0pt}
\item[Source:] per-model PII training labels (label\_profile='model')
\item[Model ref:] \texttt{urchade/gliner\_multi\_pii-v1}
\item[Parameter count:] 288,949,504
\item[Backend:] \texttt{spanbased.urchade\_gliner\_multi\_pii\_v1}
\item[Request mode:] GLiNER zero-shot span tagging (model.predict\_entities)
\item[Labels (53):] \textit{person, organization, phone number, address, passport number, email, credit card number, social security number, health insurance id number, date of birth, mobile phone number, bank account number, medication, cpf, driver's license number, tax identification number, medical condition, identity card number, national id number, ip address, email address, iban, credit card expiration date, username, health insurance number, registration number, student id number, insurance number, flight number, landline phone number, blood type, cvv, reservation number, digital signature, social media handle, license plate number, cnpj, postal code, passport\_number, serial number, vehicle registration number, credit card brand, fax number, visa number, insurance company, identity document number, transaction number, national health insurance number, cvc, birth certificate number, train ticket number, passport expiration date, social\_security\_number}
\item[Inference settings:] batch\_size=4, chunk\_infer\_batch\_size=2, chunk\_tokens=384, preferred\_outer\_batch\_size=4, threshold=0.1019, threshold\_source=optimized
\item[Notes:] GLiNER models accept the zero-shot label list at inference; we feed the model its own training labels (the MODEL\_LABELS list exported from each backend module). The decision threshold is optimised on a held-out slice.
\end{description}

\ \\

\paragraph*{\texttt{gretel\_gliner\_bi\_large\_v1\_0} \hfill\textnormal{\emph{GLiNER}}}
\nopagebreak
\begin{description}\setlength{\itemsep}{0pt}\setlength{\parsep}{0pt}\setlength{\topsep}{0pt}\setlength{\partopsep}{0pt}
\item[Source:] per-model PII training labels (label\_profile='model')
\item[Model ref:] \texttt{gretelai/gretel-gliner-bi-large-v1.0}
\item[Parameter count:] 569,075,712
\item[Backend:] \texttt{spanbased.gretel\_gliner\_bi\_large\_v1\_0}
\item[Request mode:] GLiNER zero-shot span tagging (model.predict\_entities)
\item[Labels (42):] \textit{medical\_record\_number, date\_of\_birth, ssn, date, first\_name, email, last\_name, customer\_id, employee\_id, name, street\_address, phone\_number, ipv4, credit\_card\_number, license\_plate, address, user\_name, device\_identifier, bank\_routing\_number, date\_time, company\_name, unique\_identifier, biometric\_identifier, account\_number, city, certificate\_license\_number, time, postcode, vehicle\_identifier, coordinate, country, api\_key, ipv6, password, health\_plan\_beneficiary\_number, national\_id, tax\_id, url, state, swift\_bic, cvv, pin}
\item[Inference settings:] batch\_size=4, chunk\_infer\_batch\_size=2, chunk\_tokens=384, preferred\_outer\_batch\_size=4, threshold=0.1019, threshold\_source=optimized
\item[Notes:] GLiNER models accept the zero-shot label list at inference; we feed the model its own training labels (the MODEL\_LABELS list exported from each backend module). The decision threshold is optimised on a held-out slice.
\end{description}

\ \\

\paragraph*{\texttt{knowledgator/gliner-pii-base-v1.0} \hfill\textnormal{\emph{GLiNER}}}
\nopagebreak
\begin{description}\setlength{\itemsep}{0pt}\setlength{\parsep}{0pt}\setlength{\topsep}{0pt}\setlength{\partopsep}{0pt}
\item[Source:] per-model PII training labels (label\_profile='model')
\item[Model ref:] \texttt{knowledgator/gliner-pii-base-v1.0}
\item[Parameter count:] 166,023,936
\item[Backend:] \texttt{spanbased.knowledgator\_gliner\_pii\_base\_v1\_0}
\item[Request mode:] GLiNER zero-shot span tagging (model.predict\_entities)
\item[Labels (35):] \textit{name, first name, last name, name medical professional, dob, age, gender, marital status, email address, phone number, ip address, url, location address, location street, location city, location state, location country, location zip, account number, bank account, routing number, credit card, credit card expiration, cvv, ssn, money, condition, medical process, drug, dose, blood type, injury, organization medical facility, healthcare number, medical code}
\item[Inference settings:] batch\_size=4, chunk\_infer\_batch\_size=2, chunk\_tokens=384, preferred\_outer\_batch\_size=4, threshold=0.255, threshold\_source=optimized
\item[Notes:] GLiNER models accept the zero-shot label list at inference; we feed the model its own training labels (the MODEL\_LABELS list exported from each backend module). The decision threshold is optimised on a held-out slice.
\end{description}

\clearpage

\paragraph*{\texttt{E3-JSI/gliner-multi-pii-domains-v1} \hfill\textnormal{\emph{GLiNER}}}
\nopagebreak
\begin{description}\setlength{\itemsep}{0pt}\setlength{\parsep}{0pt}\setlength{\topsep}{0pt}\setlength{\partopsep}{0pt}
\item[Source:] per-model PII training labels (label\_profile='model')
\item[Model ref:] \texttt{E3-JSI/gliner-multi-pii-domains-v1}
\item[Parameter count:] 288,949,504
\item[Backend:] \texttt{spanbased.e3\_jsi\_gliner\_multi\_pii\_domains\_v1}
\item[Request mode:] GLiNER zero-shot span tagging (model.predict\_entities)
\item[Labels (51):] \textit{person, organization, phone number, address, passport number, email, credit card number, social security number, health insurance id number, date of birth, mobile phone number, bank account number, medication, cpf, driver's license number, tax identification number, medical condition, identity card number, national id number, ip address, email address, iban, credit card expiration date, username, health insurance number, registration number, student id number, insurance number, flight number, landline phone number, blood type, cvv, reservation number, digital signature, social media handle, license plate number, cnpj, postal code, serial number, vehicle registration number, credit card brand, fax number, visa number, insurance company, identity document number, transaction number, national health insurance number, cvc, birth certificate number, train ticket number, passport expiration date}
\item[Inference settings:] batch\_size=4, chunk\_infer\_batch\_size=2, chunk\_tokens=384, preferred\_outer\_batch\_size=4, threshold=0.01, threshold\_source=optimized
\item[Notes:] GLiNER models accept the zero-shot label list at inference; we feed the model its own training labels (the MODEL\_LABELS list exported from each backend module). The decision threshold is optimised on a held-out slice.
\end{description}

\ \\

\paragraph*{\texttt{gliner\_multi\_v21} \hfill\textnormal{\emph{GLiNER}}}
\nopagebreak
\begin{description}\setlength{\itemsep}{0pt}\setlength{\parsep}{0pt}\setlength{\topsep}{0pt}\setlength{\partopsep}{0pt}
\item[Source:] GLiNER zero-shot label list (label\_profile='model')
\item[Model ref:] \texttt{urchade/gliner\_multi-v2.1}
\item[Parameter count:] 288,949,504
\item[Backend:] \texttt{spanbased.gliner\_multi\_v21}
\item[Request mode:] GLiNER zero-shot span tagging (model.predict\_entities)
\item[Labels (172):] \textit{medical record number, date of birth, ssn, date, first name, email, last name, customer id, employee id, name, street address, phone number, ipv4, credit card number, license plate, address, user name, device identifier, bank routing number, date time, company name, unique identifier, biometric identifier, account number, city, certificate license number, time, postcode, vehicle identifier, coordinate, country, api key, ipv6, password, health plan beneficiary number, national id, tax id, url, state, swift bic, cvv, pin, age, blood type, county, credit debit card, education level, employment status, fax number, gender, http cookie, language, mac address, occupation, political view, race ethnicity, religious belief, sexuality, unique id, prefix, imei, job area, job type, job title, street, secondary address, user agent, account name, currency symbol, amount, credit card issuer, sex, ip address, ethereum address, bitcoin address, middle name, iban, vehicle registration number, currency, currency name, currency code, building number, ordinal direction, masked number, zip code, bic, gps coordinates, vin, eye color, height, username, name medical professional, dob, marital status, email address, location address, location street, location city, location state, location country, location zip, bank account, routing number, credit card, credit card expiration, money, condition, medical process, drug, dose, injury, organization medical facility, healthcare number, medical code, person, organization, passport number, social security number, health insurance id number, mobile phone number, bank account number, medication, cpf, driver's license number, tax identification number, medical condition, identity card number, national id number, credit card expiration date, health insurance number, registration number, student id number, insurance number, flight number, landline phone number, reservation number, digital signature, social media handle, license plate number, cnpj, postal code, serial number, credit card brand, visa number, insurance company, identity document number, transaction number, national health insurance number, cvc, birth certificate number, train ticket number, passport expiration date, alias, title, region, district, building, unit, landmark, company, government, education, media, product, phone, social\_account, messenger, passport, document\_id, event\_date, card\_number, crypto\_wallet}
\item[Inference settings:] batch\_size=4, chunk\_infer\_batch\_size=2, chunk\_tokens=384, preferred\_outer\_batch\_size=4, threshold=0.1019, threshold\_source=optimized
\item[Notes:] GLiNER models accept the zero-shot label list at inference; we feed the model its own training labels (the MODEL\_LABELS list exported from each backend module). The decision threshold is optimised on a held-out slice.
\end{description}

\clearpage
\subsection{Fixed-Label Token Classifiers}
\ \\

\paragraph*{\texttt{openai/privacy-filter} \hfill\textnormal{\emph{OpenAI Privacy Filter}}}
\nopagebreak
\begin{description}\setlength{\itemsep}{0pt}\setlength{\parsep}{0pt}\setlength{\topsep}{0pt}\setlength{\partopsep}{0pt}
\item[Source:] OpenAI Privacy Filter API (closed PII categories)
\item[Model ref:] \texttt{openai/privacy-filter}
\item[Parameter count:] 1,399,486,865
\item[Backend:] \texttt{tokenbased.openai\_privacy\_filter}
\item[Request mode:] token-classification (BIO head)
\item[Inference settings:] attn\_implementation=kernels-community/vllm-flash-attn3, batch\_size=1, preferred\_outer\_batch\_size=1, use\_bf16=true
\item[Notes:] Closed-vocabulary token classifier: the label inventory is fixed by the model's classification head and is not user-configurable. Any non-O tag is treated as a redaction.
\end{description}

\ \\

\paragraph*{\texttt{ai4privacy\_llama\_english\_anonymiser\_openpii} \hfill\textnormal{\emph{ModernBERT (BIO)}}}
\nopagebreak
\begin{description}\setlength{\itemsep}{0pt}\setlength{\parsep}{0pt}\setlength{\topsep}{0pt}\setlength{\partopsep}{0pt}
\item[Source:] model-internal BIO taxonomy (closed; non-O outputs are mapped to redact)
\item[Model ref:] \texttt{ai4privacy/llama-ai4privacy-english-anonymiser-openpii}
\item[Parameter count:] 149,607,171
\item[Backend:] \texttt{tokenbased.ai4privacy\_llama\_english\_anonymiser\_openpii}
\item[Request mode:] token-classification (BIO head)
\item[Inference settings:] batch\_size=16, chunk\_infer\_batch\_size=8, chunk\_overlap\_tokens=64, chunk\_tokens=512, min\_score=0, preferred\_outer\_batch\_size=32
\item[Notes:] Closed-vocabulary token classifier: the label inventory is fixed by the model's classification head and is not user-configurable. Any non-O tag is treated as a redaction.
\end{description}

\ \\

\paragraph*{\texttt{ai4privacy/llama-ai4privacy-multilingual-categorical-anonymiser-openpii} \hfill\textnormal{\emph{ModernBERT (BIO)}}}
\nopagebreak
\begin{description}\setlength{\itemsep}{0pt}\setlength{\parsep}{0pt}\setlength{\topsep}{0pt}\setlength{\partopsep}{0pt}
\item[Source:] model-internal BIO taxonomy (closed; non-O outputs are mapped to redact)
\item[Model ref:] \texttt{ai4privacy/llama-ai4privacy-multilingual-categorical-anonymiser-openpii}
\item[Parameter count:] 149,635,624
\item[Backend:] \texttt{tokenbased.ai4privacy\_llama\_multilingual\_categorical\_anonymiser\_openpii}
\item[Request mode:] token-classification (BIO head)
\item[Inference settings:] batch\_size=16, chunk\_infer\_batch\_size=8, chunk\_overlap\_tokens=64, chunk\_tokens=512, min\_score=0, preferred\_outer\_batch\_size=32
\item[Notes:] Closed-vocabulary token classifier: the label inventory is fixed by the model's classification head and is not user-configurable. Any non-O tag is treated as a redaction.
\end{description}

\ \\

\paragraph*{\texttt{hydroxai\_pii\_masker} \hfill\textnormal{\emph{DeBERTa-v3}}}
\nopagebreak
\begin{description}\setlength{\itemsep}{0pt}\setlength{\parsep}{0pt}\setlength{\topsep}{0pt}\setlength{\partopsep}{0pt}
\item[Source:] model-internal BIO taxonomy (closed; non-O outputs are mapped to redact)
\item[Model ref:] \texttt{hydroxai/pii\_model\_weight}
\item[Parameter count:] 183,841,549
\item[Backend:] \texttt{tokenbased.hydroxai\_pii\_masker}
\item[Request mode:] token-classification (BIO head)
\item[Inference settings:] base\_model=microsoft/deberta-v3-base, batch\_size=8, chunk\_infer\_batch\_size=8, chunk\_overlap\_tokens=64, chunk\_tokens=512, min\_score=0, preferred\_outer\_batch\_size=32
\item[Notes:] Closed-vocabulary token classifier: the label inventory is fixed by the model's classification head and is not user-configurable. Any non-O tag is treated as a redaction.
\end{description}

\ \\

\paragraph*{\texttt{iiiorg/piiranha-v1-detect-personal-information} \hfill\textnormal{\emph{RoBERTa / Other}}}
\nopagebreak
\begin{description}\setlength{\itemsep}{0pt}\setlength{\parsep}{0pt}\setlength{\topsep}{0pt}\setlength{\partopsep}{0pt}
\item[Source:] model-internal BIO taxonomy (closed; non-O outputs are mapped to redact)
\item[Model ref:] \texttt{iiiorg/piiranha-v1-detect-personal-information}
\item[Parameter count:] 278,232,594
\item[Backend:] \texttt{tokenbased.iiiorg\_piiranha\_v1\_detect\_personal\_information}
\item[Request mode:] token-classification (BIO head)
\item[Inference settings:] batch\_size=8, chunk\_infer\_batch\_size=8, chunk\_overlap\_tokens=64, chunk\_tokens=512, min\_score=0, preferred\_outer\_batch\_size=32
\item[Notes:] Closed-vocabulary token classifier: the label inventory is fixed by the model's classification head and is not user-configurable. Any non-O tag is treated as a redaction.
\end{description}

\ \\

\paragraph*{\texttt{isotonic\_distilbert\_ai4privacy\_v2} \hfill\textnormal{\emph{DistilBERT (BIO)}}}
\nopagebreak
\begin{description}\setlength{\itemsep}{0pt}\setlength{\parsep}{0pt}\setlength{\topsep}{0pt}\setlength{\partopsep}{0pt}
\item[Source:] model-internal BIO taxonomy (closed; non-O outputs are mapped to redact)
\item[Model ref:] \texttt{Isotonic/distilbert\_finetuned\_ai4privacy\_v2}
\item[Parameter count:] 66,448,239
\item[Backend:] \texttt{tokenbased.isotonic\_distilbert\_ai4privacy\_v2}
\item[Request mode:] token-classification (BIO head)
\item[Inference settings:] batch\_size=32, min\_score=0
\item[Notes:] Closed-vocabulary token classifier: the label inventory is fixed by the model's classification head and is not user-configurable. Any non-O tag is treated as a redaction.
\end{description}

\ \\

\paragraph*{\texttt{lakshyakh93\_deberta\_finetuned\_pii} \hfill\textnormal{\emph{DeBERTa-v3}}}
\nopagebreak
\begin{description}\setlength{\itemsep}{0pt}\setlength{\parsep}{0pt}\setlength{\topsep}{0pt}\setlength{\partopsep}{0pt}
\item[Source:] model-internal BIO taxonomy (closed; non-O outputs are mapped to redact)
\item[Model ref:] \texttt{lakshyakh93/deberta\_finetuned\_pii}
\item[Parameter count:] 277,381,864
\item[Backend:] \texttt{tokenbased.lakshyakh93\_deberta\_finetuned\_pii}
\item[Request mode:] token-classification (BIO head)
\item[Inference settings:] batch\_size=32, min\_score=0, use\_bf16=false
\item[Notes:] Closed-vocabulary token classifier: the label inventory is fixed by the model's classification head and is not user-configurable. Any non-O tag is treated as a redaction.
\end{description}

\ \\

\paragraph*{\texttt{h2oai\_deberta\_finetuned\_pii} \hfill\textnormal{\emph{DeBERTa-v3}}}
\nopagebreak
\begin{description}\setlength{\itemsep}{0pt}\setlength{\parsep}{0pt}\setlength{\topsep}{0pt}\setlength{\partopsep}{0pt}
\item[Source:] model-internal BIO taxonomy (closed; non-O outputs are mapped to redact)
\item[Model ref:] \texttt{h2oai/deberta\_finetuned\_pii}
\item[Parameter count:] 138,690,932
\item[Backend:] \texttt{tokenbased.h2oai\_deberta\_finetuned\_pii}
\item[Request mode:] token-classification (BIO head)
\item[Inference settings:] batch\_size=32, min\_score=0, use\_bf16=false
\item[Notes:] Closed-vocabulary token classifier: the label inventory is fixed by the model's classification head and is not user-configurable. Any non-O tag is treated as a redaction.
\end{description}

\ \\

\paragraph*{\texttt{tanaos/tanaos-text-anonymizer-v1} \hfill\textnormal{\emph{RoBERTa / Other}}}
\nopagebreak
\begin{description}\setlength{\itemsep}{0pt}\setlength{\parsep}{0pt}\setlength{\topsep}{0pt}\setlength{\partopsep}{0pt}
\item[Source:] model-internal BIO taxonomy (closed; non-O outputs are mapped to redact)
\item[Model ref:] \texttt{tanaos/tanaos-text-anonymizer-v1}
\item[Parameter count:] 124,063,499
\item[Backend:] \texttt{tokenbased.tanaos\_tanaos\_text\_anonymizer\_v1}
\item[Request mode:] token-classification (BIO head)
\item[Inference settings:] batch\_size=16, min\_score=0
\item[Notes:] Closed-vocabulary token classifier: the label inventory is fixed by the model's classification head and is not user-configurable. Any non-O tag is treated as a redaction.
\end{description}

\ \\

\paragraph*{\texttt{deepaksiloka\_pii\_detection\_v21} \hfill\textnormal{\emph{DistilBERT (BIO)}}}
\nopagebreak
\begin{description}\setlength{\itemsep}{0pt}\setlength{\parsep}{0pt}\setlength{\topsep}{0pt}\setlength{\partopsep}{0pt}
\item[Source:] model-internal BIO taxonomy (closed; non-O outputs are mapped to redact)
\item[Model ref:] \texttt{deepaksiloka/PII-Detection-V2.1}
\item[Parameter count:] 66,402,099
\item[Backend:] \texttt{tokenbased.deepaksiloka\_pii\_detection\_v21}
\item[Request mode:] token-classification (BIO head)
\item[Inference settings:] batch\_size=32, min\_score=0
\item[Notes:] Closed-vocabulary token classifier: the label inventory is fixed by the model's classification head and is not user-configurable. Any non-O tag is treated as a redaction.
\end{description}

}

\end{document}